\documentclass[manuscript,screen]{acmart}

\usepackage{amsmath}
\usepackage{algorithm,algorithmic}
\usepackage{array,multirow,cases}
\usepackage{graphicx}
\usepackage{float, comment}
\usepackage[caption=false,font=footnotesize]{subfig}
\usepackage{pgf, tikz}
\usepackage{adjustbox, dashbox, microtype}
\usepackage[10pt]{moresize}
\usepackage{dirtytalk}
\usepackage[switch, modulo]{lineno}
\nolinenumbers
\usepackage{color}

\AtBeginDocument{%
  }

\setcopyright{acmlicensed}
\copyrightyear{2024}
\acmYear{2024}
\acmDOI{XXXXXXX.XXXXXXX}

\acmJournal{JACM}
\acmVolume{37}
\acmNumber{4}
\acmArticle{111}
\acmMonth{8}

\begin{document}

\title{Generation and Editing of Mandrill Faces:
Application to Sex Editing and Assessment}

\author{Nicolas M. Dibot}
\affiliation{%
%  \institution{LIRMM, Université de Montpellier, CNRS}
  \institution{CEFE, Univ. Montpellier, CNRS, EPHE, IRD}
  \city{Montpellier}
  \country{France}
%  \postcode{34000}
}
\email{nicolas.dibot@cefe.cnrs.fr} 

\author{Julien P. Renoult}
\affiliation{%
  \institution{CEFE, Univ. Montpellier, CNRS, EPHE, IRD}
  \city{Montpellier}
  \country{France}
%  \postcode{34000}
}
\email{julien.renoult@cefe.cnrs.fr} 

\author{William Puech}
\orcid{0000-0001-9383-2401}
\affiliation{%
  \institution{LIRMM, Université de Montpellier, CNRS}
  \city{Montpellier}
  \country{France}
%  \postcode{34095}
}
\email{william.puech@lirmm.fr} 
\authornote{J. P. Renoult and W. Puech contributed equally to this work.} 

\renewcommand{\shortauthors}{Dibot et al.}

\begin{abstract}
 Generative AI has seen major developments in recent years, enhancing the realism of synthetic images, also known as computer-generated images. In addition, generative AI has also made it possible to modify specific image characteristics through image editing. Previous work has developed methods based on generative adversarial networks (GAN) for generating realistic images, in particular faces, but also to modify specific features. However, this work has never been applied to specific animal species. Moreover, the assessment of the results has been generally done subjectively, rather than quantitatively. In this paper, we propose an approach based on methods for generating images of faces of male or female mandrills, a non-human primate. %The proposed method is also capable of editing their sex by identifying a sex axis in the latent space of a specific GAN.
 The main novelty of proposed method is the ability to edit their sex by identifying a sex axis in the latent space of a specific GAN.
 %Statistical features extracted from real image distributions have been used to develop the assessments.  
 In addition, we have developed an assessment of the sex levels based on statistical features extracted from real image distributions.
 The experimental results we obtained from a specific database are not only realistic, but also accurate, meeting a need for future work in behavioral experiments with wild mandrills.
\end{abstract}





\maketitle

%%%%%%%%%%%%%%%%%%%%%%%%%%%%%%%%%%%%%%%%%%%%%%%%%%%%%%%
%                 Introduction                        %
%%%%%%%%%%%%%%%%%%%%%%%%%%%%%%%%%%%%%%%%%%%%%%%%%%%%%%%
\section{Introduction}
\label{sec:introduction}

For more than a decade, AI and more specifically machine learning (ML) has revolutionized a wide range of fields, and image processing in particular. With deep learning, particularly with CNNs, which were revived by LeCun~{\it et al.} in 2015 for image classification~\cite{LeCunBH15}, one of the most common ML tasks involves training a model to predict labels for input data. However, ML encompasses many other tasks such as; segmentation~\cite{9356353}, clustering~\cite{8590804}, and, regression and quality assessment~\cite{7915790}. The role of these ML systems is to learn patterns and relationships from training data, using a variety of techniques. They require the creation of mathematical models capable of generalizing from training data such as images to make predictions on new data.

More recently, these ML models have been used to generate synthetic images, also known as computer-generated images, and perform various tasks on them ~\cite{bermano_state---art_2022}. These models, known as generative AI (GenAI), and, in particular, generative adversarial networks (GAN), have demonstrated their performance~\cite{goodfellow_generative_2014}. One of the most popular tasks at the moment is style transfer, which consists of editing an image to transform it into another image with a different style~\cite{9906917, 10021829}. Among the most popular GenAI models, the StyleGAN3 architecture has proved particularly innovative in combining GAN with the style transfer technique~\cite{Karras2021}. Innovations have even made it possible to use StyleGAN to modify real images according to variables of interest. 

In the fields of ecology and evolution, deep learning has already been used for detection ~\cite{9411844,5665772} or classification tasks~\cite{tieo_social_2023}. 
Previous GenAI-based approaches have been proposed in ecology~\cite{talas_camogan_2020, hirn_deep_2022, roy_using_2022}, but not specifically for animal behavior, including human behavior, or for the study of visual signals. However, work on mandrill behavior has been carried out using traditional computer vision metrics~\cite{renoult_evolution_2011} or non-generative deep learning~\cite{tieo_social_2023}. 

In this paper, we propose a method for generating artificial mandrill faces, a non-human primate from central Africa, and then editing them according to a specific characteristic which is the sex level, for mandrill males (masculinity) and females (feminity). Furthermore, in our approach, we propose to assess this sex-level variation as a function of a real distribution. In order to confirm the realism and the visual quality of the images of mandrill faces obtained, we applied our approach to a specific database containing a large number of mandrill faces from Gabon, both male and female, the Mandrillus Face Database (MFD)~\cite{TIEO2023108939}.

The main contributions we propose in this paper are:
\begin{itemize}
\item An application of GAN-based synthetic image generation to face images of a primate species, the mandrill;
\item Editing the sex level of these mandrill faces;
\item A method for quantitative assessment of the variable used for editing, with a perspective of application to experimental choice tests for a behavior analysis. 
\end{itemize}

The rest of this paper is organized as follows. In Section~\ref{sec:related_work}, we first detail current state-of-the-art approaches, related to GAN and image editing, and then present previous studies on animal behavior and ecology studies using deep learning and generative AI. Section~\ref{sec:method} describes the proposed method for editing mandrill faces and assessing their sex level. Section~\ref{sec:Results} presents experimental results obtained with the method we propose, followed by a discussion. Finally, this paper is concluded in Section~\ref{sec:conclusion}.

%%%%%%%%%%%%%%%%%%%%%%%%%%%%%%%%%%%%%%%%%%%%%%%%%%%%%%%
%                 Related work                        %
%%%%%%%%%%%%%%%%%%%%%%%%%%%%%%%%%%%%%%%%%%%%%%%%%%%%%%%
\section{Related work}
\label{sec:related_work}

In this section, we present the previous work on which our proposed method is based. In Section~\ref{sec:GAN}, we first detail some generative AI concepts, specifically GAN. The application of these concepts to image editing is then discussed  in Section~\ref{sec:handmade}. Finally, links with the field of ecology, and more specifically with the study of animal behavior are then detailed in Section~\ref{sec:mandrills}.

%%%%%%%%%%%%%%%%%%%%%%%%%%%%%%%%%%%%%%%%%%%%%%%%%%%%%%%
\subsection{Generative AI, GAN and encoding}
\label{sec:GAN}

Generative adversarial networks (GAN) were first developped 10 years ago~\cite{goodfellow_generative_2014}.  They are based on an adversarial training mechanism where two neural networks, a generator and a discriminator, are trained together. While the generator aims to create realistic synthetic images whose feature distribution matches that of a real training image database, the discriminator must learn to distinguish between images from the same real image database and the generator's synthetic images. Subsequently, major improvements were proposed by the Wasserstein GAN~\cite{arjovsky_wasserstein_2017} which improves the convergence and the realism of these models, using an optimization strategy based on optimal transport~\cite{villani_optimal_2009}, while the Progressive Growing GAN generates more realistic images~\cite{karras_progressive_2018}. Afterwards, StyleGAN significantly improves the realism of the images, particularly portraits~\cite{karras_style-based_2019}. This architecture was enhanced in 2020 with StyleGAN2~\cite{karras_analyzing_2020} and in 2021 with StyleGAN3~\cite{Karras2021}. StyleGAN3 is particularly powerful and realistic, enabling the construction of latent spaces where image features are linear, that is, disentangled~\cite{wu_stylespace_2020,bermano_state---art_2022}.
The specificity of the family of StyleGan models lies in the integration of a third neural network, called the mapping network, in addition to the generator and discriminator. This additional network generates a feature vector, which is injected into the generator at various stages, enabling fine-tuned control of the attributes of the generated image.
In comparison, to date and to our knowledge, there are no similar approaches based on other generative deep learning architectures for editing, encoding and precisely manipulating images at the same time. Results obtained by variational autoencoders (VAE)~\cite{kingma_auto-encoding_2022} are not as realistic, while diffusion models although realistic, do not allow easy manipulation of their latent encoding spaces~\cite{sohl-dickstein_deep_2015}.

With VAE, by construction, it is trivial to find the position of a given image through its latent vector in the latent space of the trained model, since these models consist of an encoder and a decoder~\cite{kingma_auto-encoding_2022}. GAN, on the other hand, only have a decoder, but no encoder. They are therefore able to generate realistic synthetic images, but are unable to find the position of a real image in their latent space. For this purpose, encoder architectures additional to a StyleGAN3 generator have been proposed~\cite{bermano_state---art_2022}.  In particular, the pSp framework enables fairly faithful encoding of real images~\cite{richardson_encoding_2021}. To achieve this, it encodes not in the latent space located at the beginning of the generator, but in a larger space corresponding to all layers of the generator, this enables greater precision. However, the method is still not perfected but this is an area for future improvements.

%%%%%%%%%%%%%%%%%%%%%%%%%%%%%%%%%%%%%%%%%%%%%%%%%%%%%%%
\subsection{Handmade versus GAN-based editing}
\label{sec:handmade}

Image editing consists of modifying specific characteristics of an image. For example, it is possible to edit the sex of a portrait, making it more masculine or feminine, or the age, making it younger or older. Morphing techniques based on Delaunay triangulation~\cite{tiddeman_general_2001} or wavelets~\cite{tiddeman_prototyping_2001,nila_male_2019} are not visually efficient. GAN in general and StyleGAN in particular offer advantages over these previous methods. GAN latent spaces perform well to predict continuous variables independently of their generative capacity~\cite{nitzan_large_2022}. In particular, StyleGAN performs well not only to construct a latent space where a specific variable can be identified, but also to generate images modified according to this specific variable~\cite{alaluf_only_2021}. Several methods have been proposed to identify the axis that describes this specific variable in the GAN latent space. The supervised linear method uses a Support Vector Machine (SVM) to separate two clusters corresponding to the two extremities of the variable, and to isolate the axis orthogonal to the SVM support vector~\cite{alaluf_third_2022}. In the absence of labeled data for the supervised linear method, there are also unsupervised linear approaches without labels~\cite{voynov_unsupervised_2020, harkonen_ganspace_2020}.

%%%%%%%%%%%%%%%%%%%%%%%%%%%%%%%%%%%%%%%%%%%%%%%%%%%%%%%
\subsection{Deep learning applied to mandrill behavior and visual communication} 
\label{sec:mandrills}

Mandrills, \textit{Mandrillus sphinx}, are a species of primate in the cercopithecidae family~\cite{hill_primates_1953}. They are classified as vulnerable by the International Union for Conservation of Nature (IUCN)~\cite{stirling_iucn_2016}. Academic work to better understand their behavior within an evolutionary biology paradigm has been ongoing for many years~\cite{charpentier_distribution_2012}. More recently, a database of mandrill portraits has been created, the Mandrillus Face Database (MFD)~\cite{TIEO2023108939}. The MFD database detailed in Section~\ref{subsec:bdd}, makes it possible to use computer vision approaches based on deep learning to study mandrill behavior~\cite{charpentier_same_2020, charpentier_mandrill_2022, tieo_social_2023}. The use of deep learning in behavioral ecology has increased in recent years to understand the evolution of visual signals and visual perception in many species, beyond the application to mandrills~\cite{hulse_using_2022,hejja-brichard_using_2023, dibot_sparsity_2023}. Generative models such as GAN, have been rarely used in ecology, and usually for quite different questions, such as modeling the evolution of the prey/predator relationships~\cite{talas_camogan_2020}, predicting species coexistence patterns~\cite{hirn_deep_2022} or modeling the trajectory of birds~\cite{roy_using_2022}.

To our knowledge, there are no GAN-based approaches for generating visual stimuli that give rise to behavioral experiments. With this objective in sight, we have developed the method presented in this article to meet a need for ecologists interested in animal behavior using mandrills as a study model.

%%%%%%%%%%%%%%%%%%%%%%%%%%%%%%%%%%%%%%%%%%%%%%%%%%%%%%%
%         method             %
%%%%%%%%%%%%%%%%%%%%%%%%%%%%%%%%%%%%%%%%%%%%%%%%%%%%%%%
\section{The proposed method for editing images of mandrill faces}
\label{sec:method} 

The aim of this work is to edit images of mandrill faces in such a way as to modify their femininity or masculinity. In this paper, we define the concept of sex level as the level of belonging to one of the two sexes on a one-dimensional axis. In this section, we develop in detail the method we propose for editing mandrill faces in order to modify their sex level by editing the corresponding generated images of mandrill faces.

The whole process includes a training phase, and an editing phase. The goal of the training phase consists of generating all the necessary processes and data for the editing phase, in particular a trained StyleGAN3~\cite{Karras2021} specific for mandrill face images, and a trained pSp encoder~\cite{richardson2021encoding} for mandrills to generate image latent vector in latent space. The goal of the training phase is also to obtain a sex distribution of mandrill faces and to calculate the editing vector. The objective of the editing phase is to modify the mandrill’s masculinity (if it is a male) or femininity (if it is a female) by editing a real image of a mandrill face.

First, in Section~\ref{sec:overviewTraining}, we give an overview of the training phase, which requires as an input, a database of real images of mandrill faces. The main parts of the training phase are then detailed to explain, in Section~\ref{sec:GenerationMandrillFaces} how to generate a database of mandrill faces from a specific StyleGAN3, in Section~\ref{sec:pSp} to train a pSp-mandrill encoder, and in Section~\ref{sec:Editing} to determine the editing vector needed for the editing phase. Then, in Section~\ref{sec:overviewEditing}, we give an overview of the editing phase and we detail in Section~\ref{sec:sex level} the computation of the possible edition variation range. Finally, in Section~\ref{sec:editing} we give details to compute  the sex level of an edited generated image of a mandrill face.

%%%%%%%%%%%%%%%%%%%%%%%%%%%%%%%%%%%%%%%%%%%%%%%%%%%%%%%
\subsection{Overview of the training phase}
\label{sec:overviewTraining}

First, in order to generate synthetic images of mandrill faces from a database of real images, we train a StyleGAN3 model~\cite{Karras2021} on a database of real mandrill faces, the MFD Database. We then obtain a GAN called StyleGAN-mandrill. During the editing vector computation phase, then we determine in the GAN latent space, noted $W$, the vector $\boldsymbol{\overrightarrow{v}}$ that would feminize or masculinize an image of a mandrill face based on its vector in latent space. 

\begin{figure}[hbtp!]
    \centering
        \includegraphics[width=14cm]{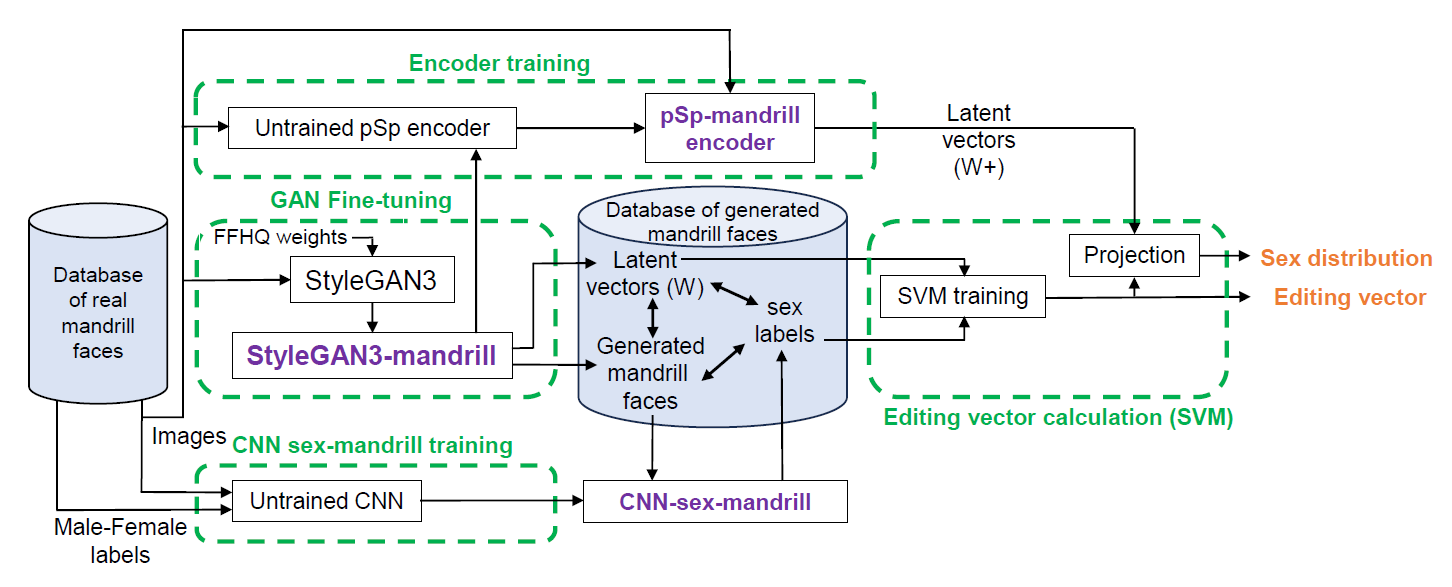}
        \caption{Overview of the training phase.}
    \label{fig:overview}
\end{figure}

For the editing phase presented in Section~\ref{sec:overviewEditing}, we first need to train several machine learning algorithms. Indeed, at the beginning we only have a database of real images of mandrill faces and only untrained or no specific algorithm architectures. An overview of the training phase is illustrated in Fig.~\ref{fig:overview}. 

This vector is calculated in the space $W$. To do this, during the CNN sex-mandrill training phase, we have to also train a classifier neural network to provide the sex label of the images generated by StyleGAN3-mandrill. Therefore we have to also train an encoder to find the latent vectors of the real images in the GAN latent space $W$ in the encoder training phase. This projection is carried out in a space called $W+$. The size of the space $W+$ is 18 times larger than that of the space $W$. According to the following tasks, we frequently need to switch from the space $W$ to the space $W+$, and back again.  Finally, we project the latent vectors of encoded real images in the space $W+$ onto the previously identified vector $\boldsymbol{\overrightarrow{v}}$ to see their distribution for comparison with the femininity or masculinity editing of a single image projected onto the same axis. 

%%%%%%%%%%%%%%%%%%%%%%%%%%%%%%%%%%%%%%%%%%%%%%%%%%%%%%%
\subsection{Generation of mandrill faces from StyleGAN3-mandrill}
\label{sec:GenerationMandrillFaces}

In order to generate random images, with a size of $1024 \times 1024$ pixels of mandrill faces, using our approach, we propose to rely on fine-tuning the StyleGAN3. As illustrated in Fig.~\ref{fig:overview}, to do this, we use a pre-trained version with weights from the FFHQ (Flickr Faces High Quality) dataset~\cite{karras_progressive_2018}, a database of 70,000 human portraits. From these weights, we fine-tune StyleGAN3 in order to obtain a specific trained StyleGAN3, StyleGAN3-mandrill, by defreezing specific layers and by using a database of mandrill faces, the Mandrillus Face Database~\cite{TIEO2023108939}, presented in Section~\ref{subsec:bdd}. We evaluate the training quality using the FID (Frechet Inception Distance) metric~\cite{yu_frechet_2021}. This metric calculates the distance between two distributions of latent vectors extracted from CNN Inception-V3: one for the real images and the other for generated images. The obtained results from the FID metric can vary according to the type of data on which the StyleGAN3 is trained.

To generate images with StyleGAN3-mandrill, the model needs to be used as an image generator. Like a decoder, it transforms small-scale information, \textit{i.e.} a vector of size 512, into large-scale information, \textit{i.e.} a $1024 \times 1024$ pixel image. To generate an image with no other purpose than that it be a realistic mandrill face, we simply pass a random latent vector of a size 512 as input to StyleGAN3-mandrill, which then generates an image of mandrill face. This latent vector, called $\boldsymbol{w}$ is injected between each layer of the generator in place of the other latent vector $\boldsymbol{w}$ generated by the mapping network when training StyleGAN3.

%%%%%%%%%%%%%%%%%%%%%%%%%%%%%%%%%%%%%%%%%%%%%%%%%%%%%%%
\subsection{Training of the pSp-mandrill encoder}
\label{sec:pSp}

Based on the architecture of an untrained encoder pSp~\cite{richardson2021encoding}, and the weights of the trained StyleGAN3-mandrill presented in Section~\ref{sec:GenerationMandrillFaces}, the aim of this step is to obtain a pSp encoder trained to specifically encode mandrill faces and called pSp-mandrill encoder as shown in Fig.~\ref{fig:overview}. Editing an image with a GAN to change a feature of interest like sex or age is an algebraic operation. It involves modifying its latent vector $\boldsymbol{w}$ along a specific direction. For our approach, this means having its vector $\boldsymbol{w}$ in the latent space $W$ of StyleGAN3-mandrill. When an image is generated by the GAN, it is trivial, because the GAN is a decoder that generates an image from a random latent vector $\boldsymbol{w}$ located in the space $W$ and injected between each layer of the generator. With a real image we can not directly obtain its coordinates. As StyleGAN3 is highly stochastic, it cannot be used \say{backwards} as an encoder to find the original latent vector precisely. Several previous work propose to overcome this problem~\cite{wu_stylespace_2020,bermano_state---art_2022}. We use the pSp (pixel2style2pixel) encoder~\cite{richardson_encoding_2021}, which allows us to obtain a latent vector $\boldsymbol{w+}$ corresponding to an input image. This latent vector does not correspond solely to the latent vector $\boldsymbol{w}$ obtained with StyleGAN3-mandrill (located in the space $W$), but to the concatenation of a set of different latent vectors, which are located in a space called $W+$. This set of latent vectors are also injected between each layer of the generator, but whereas in the classical case, the same vector $\boldsymbol{w}$ is injected several times, in this case a different vector is injected between each layer. 
StyleGan3 containing 18 layers, the dimension of latent vector $\boldsymbol{w+}$ is then 18 times larger than the latent latent vector $\boldsymbol{w}$. The proposed approach is the same whether editing in latent space $W$ or in the latent space $W+$. 

Since the encoder pSp is also a neural network, it requires training. Its necessary inputs are the images of real mandrill faces and the StyleGAN3-mandrill weights. Starting from scratch, the obtained pSp-mandrill encoder takes a real image of a mandrill face, and provides the latent vector of a real image of mandrill face in the space $W+$.

%%%%%%%%%%%%%%%%%%%%%%%%%%%%%%%%%%%%%%%%%%%%%%%%%%%%%%%
\subsection{The editing vector calculation}
\label{sec:Editing}
%- SVM
%- Projection sur l'axe

For the editing vector calculation step, from the latent vectors of the generated images, their sex labels and a SVM (Support Vector Machine), we want to calculate the sex vector $\boldsymbol{\overrightarrow{v}}$ giving us the direction along which it is possible to edit femininity or masculinity. In this paper, we postulate that the vector is the same for varying either masculinity and femininity; only the direction of editing changes according to each of the two sexes. The proposed editing vector approach is based on the method proposed by Alaluf~{\it et al.}~\cite{shen_interpreting_2020}, that we applied to our data. 

\begin{figure}[hbtp!]
    \centering
    \includegraphics[width=6cm]{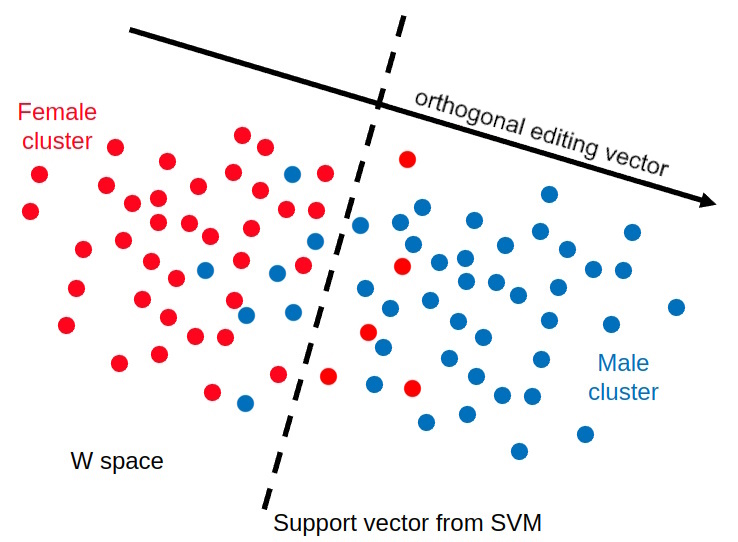}      
    \caption{Calculation of the editing vector from SVM on the clusters of female and male vectors in the space $W$.
        }
    \label{fig:svm}
\end{figure}

For this purpose, we first generated 100,000 images of mandrill faces with StyleGAN3-mandrill, to which their latent vectors are associated. Then we determine their sex with a MobileNet architecture CNN classifier~\cite{howard_mobilenets_2017} trained for this task on sex-labeled images from the database of real mandrill images. Then we train an SVM to classify the latent vectors according to their associated sex. Thus we obtain the vector $\boldsymbol{\overrightarrow{v}}$, orthogonal to the support vector separating the \say{male} cluster from the \say{female} cluster of latent vectors, as illustrated in Fig.~\ref{fig:svm}. Applied to the latent vector of an image as a scalar product, this vector $\boldsymbol{\overrightarrow{v}}$ gives the direction in which to modify a latent vector in order to change the variation of the sex level of the associated image.

The next step is to quantify the variation of the sex level caused by the editing vector $\boldsymbol{\overrightarrow{v}}$ in relation to the distribution of real images, as a ground truth. Then we encode all the real images from the Mandrillus Face Database (MFD) with the pSp-mandrill encoder in space $W+$. We project the vectors of these images onto the edit vector $\boldsymbol{\overrightarrow{v}}$, which we then consider to be a sex axis. This allows us to estimate the distribution of male and female images along the sex axis.

%%%%%%%%%%%%%%%%%%%%%%%%%%%%%%%%%%%%%%%%%%%%%%%%%%%%%%%
\subsection{Overview of the editing phase}
\label{sec:overviewEditing}

\begin{figure}[hbtp!]
    \centering
        \includegraphics[width=12cm]{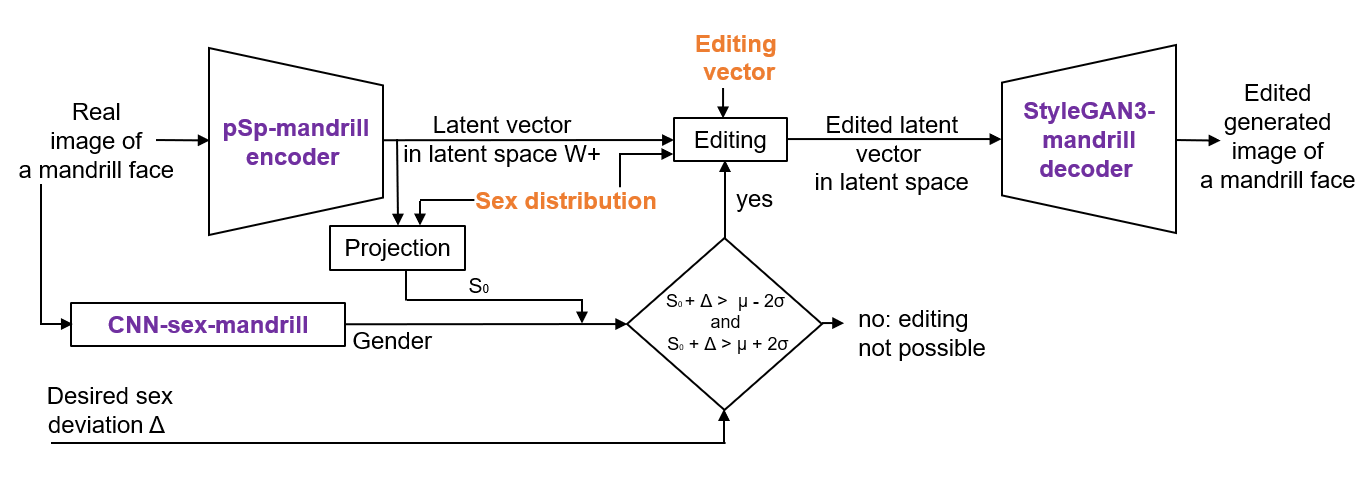}
        \caption{Overview of the editing phase.}
    \label{fig:editing}
\end{figure}

Now that all the parts are trained, we can integrate them into a pipeline for our mandrill portrait editing application as illustrated in Fig.~\ref{fig:editing}. First we select a real image of a mandrill face. Its sex is determined using the CNN-sex-mandrill classifier presented in Section~\ref{sec:Editing}. The image is then encoded with the pSp-mandrill encoder in the space $W+$ and projected onto the sex axis. From its sex level on this axis, we compute a likely editing range corresponding to ± 2 standard deviations around the mean of the distribution of all real images on this axis. From a given desired sex deviation \textit{$\Delta_d$} of editing, if this is within the range, the latent vector is then edited along the direction given by the sex axis. The new edited latent vector is obtained and decoded with the StyleGAN3-mandrill decoder to obtain a new edited image of a mandrill face.

%%%%%%%%%%%%%%%%%%%%%%%%%%%%%%%%%%%%%%%%%%%%%%%%%%%%%%%
\subsection{Possibility of the sex level deviation}
\label{sec:sex level}

The editing must modify sex level within a range that exists in reality. In Section~\ref{sec:Editing}, we explain how to obtain the distribution of real images on the sex axis. We analyze two distributions separately: that of males and that of females. For each, we calculate the mean $\mu$, more precisely $\mu_M$ for males and $\mu_F$ for females respectively, its standard deviation {$\sigma$}, $\sigma_M$ for males and $\sigma_F$ for females respectively, and determine bounds corresponding to $\pm 2 \sigma$. 

\begin{figure}[hbtp!]
    \centering
        \includegraphics[width=9cm]{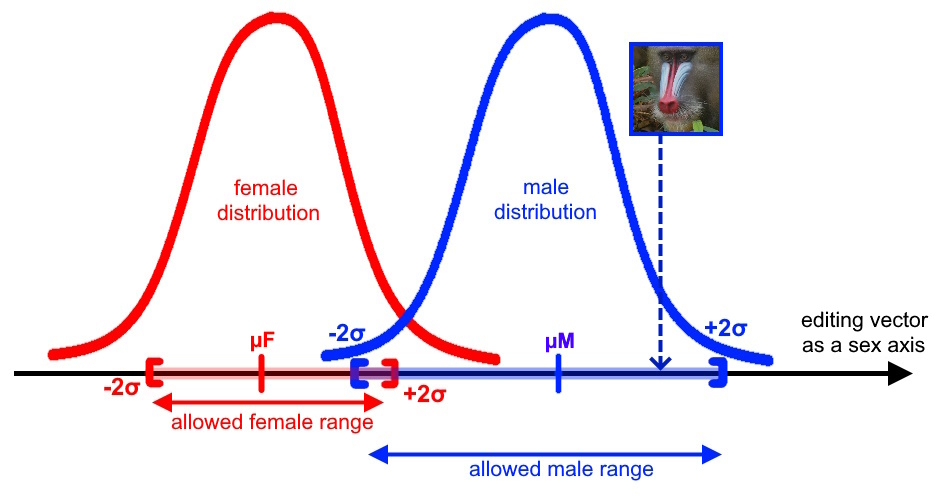}      
        \caption{Calculation of the upper and lower edit bounds for each sex according to the distributions of real images projected on the sex axis. }
    \label{fig:range}
\end{figure}

Next, we project an image encoded in the space $W+$ onto the sex axis, and calculate the two distances between its position and the upper and lower bounds of its own sex distribution as shown in Fig.~\ref{fig:range}. These distances are used to calculate whether or not editing is possible with a desired sex deviation $\Delta_d$, as presented in detail in Algorithm~\ref{algo:alg1}.

%%%%%%%%%%%%%%%%%%%%%%%%%%%%%%%%%%%%%%%%%%%%%%%%%%%%%%%
\begin{algorithm}
\caption{Calculating whether editing is possible or not.}
\label{algo:alg1}
\begin{algorithmic}[1]
\STATE \textbf{Input:} Original image $I_o$; Desired sex deviation $\Delta_d$
\STATE\textbf{Output:} Boolean $Editing$: Editing possible or not
\STATE $Sex \leftarrow$ CNNSexMandrill($I_o$);
\hspace{0.1cm} \textit{// Sex label classification}
\STATE $\boldsymbol{v_o} \leftarrow$ encodeWithpSp$(I_o)$; 
\STATE $S_o \leftarrow$ projection$(\boldsymbol{v_o})$;
%\STATE $sex level$  $\leftarrow$ Typicalitysex level((ImageCoordinates($I$));
\hspace{0.1cm} \textit{// Sex level computation}
\IF{$Sex = Male$}
\STATE $\mu_M \leftarrow$ averageMale();
\STATE $\sigma_{M} \leftarrow$ standardDeviationMale();
\IF{$S_o + \Delta_d < \mu_M - 2 \sigma_{M} $ \OR $S_o + \Delta_d > \mu_M + 2 \sigma_{M} $}
\STATE $Editing \leftarrow$ FALSE;
%\RETURN $Editing$ 
\ELSE
\STATE $Editing \leftarrow$ TRUE;
%\RETURN $Editing$ 
\ENDIF
\ELSE % IF FEMALE
\STATE $\mu_F \leftarrow$ averageFemale();
\STATE $\sigma_{F} \leftarrow$ standardDeviationFemale();
\IF{$S_o + \Delta_d < \mu_F - 2 \sigma_{F} $ \OR $S_o + \Delta_d > \mu_F + 2 \sigma_{F} $}
\STATE $Editing \leftarrow$ FALSE;
%\RETURN $Editing$ 
\ELSE
\STATE $Editing \leftarrow$ TRUE;
%\RETURN $Editing$ 
\ENDIF
\ENDIF
\RETURN $Editing$ 
\end{algorithmic}
\end{algorithm}

%%%%%%%%%%%%%%%%%%%%%%%%%%%%%%%%%%%%%%%%%%%%%%%%%%%%%%%
\subsection{Calculation and assessment of the sex level from the space $W$ to to the space $W+$  of an edited generated image of a mandrill face}
\label{sec:editing}

In this last step, we explain how to edit and assess the encoded image according to a desired deviation on the sex axis as illustrated in Fig.~\ref{fig:editing}. From the original image $I_o$, the desired deviation $\Delta_d$, two optimization parameters, the step $s$ corresponding to the modification intensity of the editing deviation to be optimized, the threshold $T$ corresponding to the editing precision tolerance, and the standard deviation $\sigma$, we want to obtain the edited image $I_e$ with the optimized editing deviation $\Delta_e$ that enabled this editing.

In order to edit an image, we cannot directly modify its coordinates according to the latent vector $\boldsymbol{v}$ multiplied by the desired deviation. In fact, in the space $W+$, the magnitude of the editing intensity is not the same depending on where it is measured. It is not the same for an encoded image directly after its edition in the space $W+$ or for this same image decoded then re-encoded with pSp-mandrill after its edition. As the editing vector is calculated in the space $W$, it does not necessarily correspond to the optimal direction of sex level in the space $W+$.

%%%%%%%%%%%%%%%%%%%%%%%%%%%%%%%%%%%%%%%%%%%%%%%%%%%%%%%
\begin{algorithm}[hbtp!]
\caption{Optimal research of the editing step.}
\label{algo:alg2}
\begin{algorithmic}[1]
\STATE \textbf{Input:} Image $I_o$; Desired deviation $\Delta_d$; 
\STATE Step $s$; Threshold $T$; standard deviation $\sigma$;
\STATE\textbf{Output:} Edited image $I_e$; Optimized deviation $\Delta_e$; 
\IF{editingPossible($I_o, \Delta_d$)}
\STATE \textit{// Original image sex level computation}
\STATE $\boldsymbol{v_o} \leftarrow$ encodeWithpSp$(I_o)$; $S_o \leftarrow$ projection$(\boldsymbol{v_o})$; 
\STATE \textit{// Initialization}
\STATE $S_r \leftarrow S_o + \Delta_d$; $\Delta_e \leftarrow \Delta_d$; 
\STATE \textit{// Generation of the initial edited image}
\STATE $\boldsymbol{v_e} \leftarrow  \boldsymbol{v_o} + \Delta_e \boldsymbol{\overrightarrow{v}}$; 
\STATE $I_e \leftarrow$ decodeWithStyleGAN3-mandrill$(\boldsymbol{v_e})$;
\STATE \textit{// Initial edited image sex level computation}
\STATE  $\boldsymbol{v_e} \leftarrow$ encodeWithpSp$(I_e)$; $S_e \leftarrow$ projection$(\boldsymbol{v_e})$;
\WHILE{$S_e \ne S_r \pm T \times \sigma$}
\IF{$S_r > S_o$}
    \IF{$S_e > S_o$}
        \IF{$S_e > S_r$}
        \STATE $\Delta_e \leftarrow \Delta_e - \Delta_e / s$ \textit{// Condition A}
        \ELSE
        \STATE $\Delta_e \leftarrow \Delta_e + \Delta_e / s$ \textit{// Condition B}
        \ENDIF
    \ELSE
\STATE $\Delta_e \leftarrow |\Delta_e| +|\Delta_e| / s$ \textit{// Condition C}
    \ENDIF
\ELSE 
    \IF{$S_e < S_o$}
        \IF{$S_e > S_r$} 
        \STATE $\Delta_e \leftarrow \Delta_e + \Delta_e / s$ \textit{// Condition D}
        \ELSE
        \STATE $\Delta_e \leftarrow \Delta_e - \Delta_e / s$ \textit{// Condition E}
        \ENDIF
    \ELSE
\STATE $\Delta_e \leftarrow -|\Delta_e| +\Delta_e / s$ \textit{// Condition F}
    \ENDIF
\ENDIF
\STATE \textit{// Generation of the current edited image}
\STATE $\boldsymbol{v_e} \leftarrow  \boldsymbol{v_o} + \Delta_e \boldsymbol{\overrightarrow{v}}$ ;  
\STATE $I_e \leftarrow$ decodeWithStyleGAN3-mandrill$(\boldsymbol{v_e})$;
\STATE \textit{// Current edited image sex level computation}
\STATE  $\boldsymbol{v_e} \leftarrow$ encodeWithpSp$(I_e)$; $S_e \leftarrow$ projection$(\boldsymbol{v_e})$; 
\ENDWHILE
\RETURN $I_e$ and $\Delta_e$
\ELSE 
\RETURN Edition not possible
\ENDIF
\end{algorithmic}
\end{algorithm}

To overcome this problem, we have developed an algorithm, presented in Algorithm~\ref{algo:alg2}, which optimizes the editing intensity to be applied to the latent vector of an encoded image by decoding and re-encoding the edited image, then comparing the sex level of the re-encoded image with the desired sex level. 
The principle of Algorithm~\ref{algo:alg2} is to edit an image with a certain deviation $\Delta$, see if its sex level corresponds to the desired sex level, then according to this, modify the value of the deviation $\Delta$ by increasing it, decreasing it, or changing its sign and repeating the same steps until the desired deviation $\Delta_d$ is reached.

\begin{figure}[hbtp!]
    \centering
        \includegraphics[width=7.5cm]{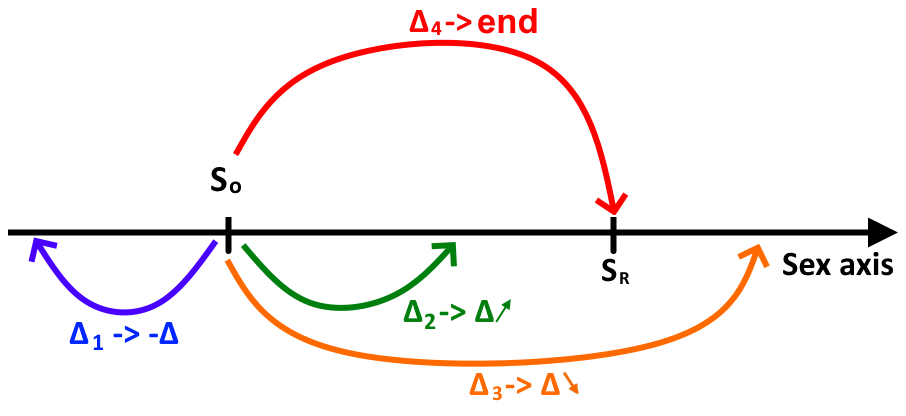}  
        \caption{Example of the research of the optimal editing step $\Delta_e$ for editing the sex level of a mandrill face. Starting from the sex level $S_0$ of the original generated image, to reach the sex level $S_R$, in the first step, $\Delta_1$ corresponds to condition $C$ of the Alg.~\ref{algo:alg2}, for the second step $\Delta_2$ to condition $B$, while for the third step $\Delta_3$ to condition $A$ and finally $\Delta_4$ to the optimal $\Delta_e$.}
    \label{fig:optimisation}
\end{figure}

Fig ~\ref{fig:optimisation} illustrates some of the conditions specified in Algorithm~\ref{algo:alg2}. This example, which searches for the optimal editing step $\Delta_e$ for editing the sex level of a mandrill face, starts from the sex level $S_0$ of the original generated image, to reach the sex level $S_R$. In the first step, $\Delta_1$ corresponds to condition $C$ of the Alg.~\ref{algo:alg2}, for the second step $\Delta_2$ to condition $B$, while for the third step $\Delta_3$ to condition $A$ and finally $\Delta_4$ to the optimal $\Delta_e$. Note that for each step, the algorithm restarts from $S_0$.

%%%%%%%%%%%%%%%%%%%%%%%%%%%%%%%%%%%%%%%%%%%%%%%%%%%%%%%
%                Results                 %
%%%%%%%%%%%%%%%%%%%%%%%%%%%%%%%%%%%%%%%%%%%%%%%%%%%%%%%
\section{Experimental results}
\label{sec:Results}

In this section, we present the results obtained with our proposed approach applied to real images of mandrill faces. After presenting the database of mandrill images used in Section~\ref{subsec:bdd}, we apply in Section~\ref{subsec:étapes} the developed method to two examples, an image of a male mandrill and an image of a female mandrill, carefully detailing all the steps. We then show the results of the method on various examples in Section~\ref{subsec:plus d'exemples}, before discussing some unexpected results and taking a step back from our work in Section~\ref{subsec:discussion}.

%%%%%%%%%%%%%%%%%%%%%%%%%%%%%%%%%%%%%%%%%%%%%%%%%%%%%%%
\subsection{The Mandrillus Face Database~\cite{TIEO2023108939}}
\label{subsec:bdd}
\begin{figure}[hbtp!]
    \centering    
    \includegraphics[width=0.48\textwidth]{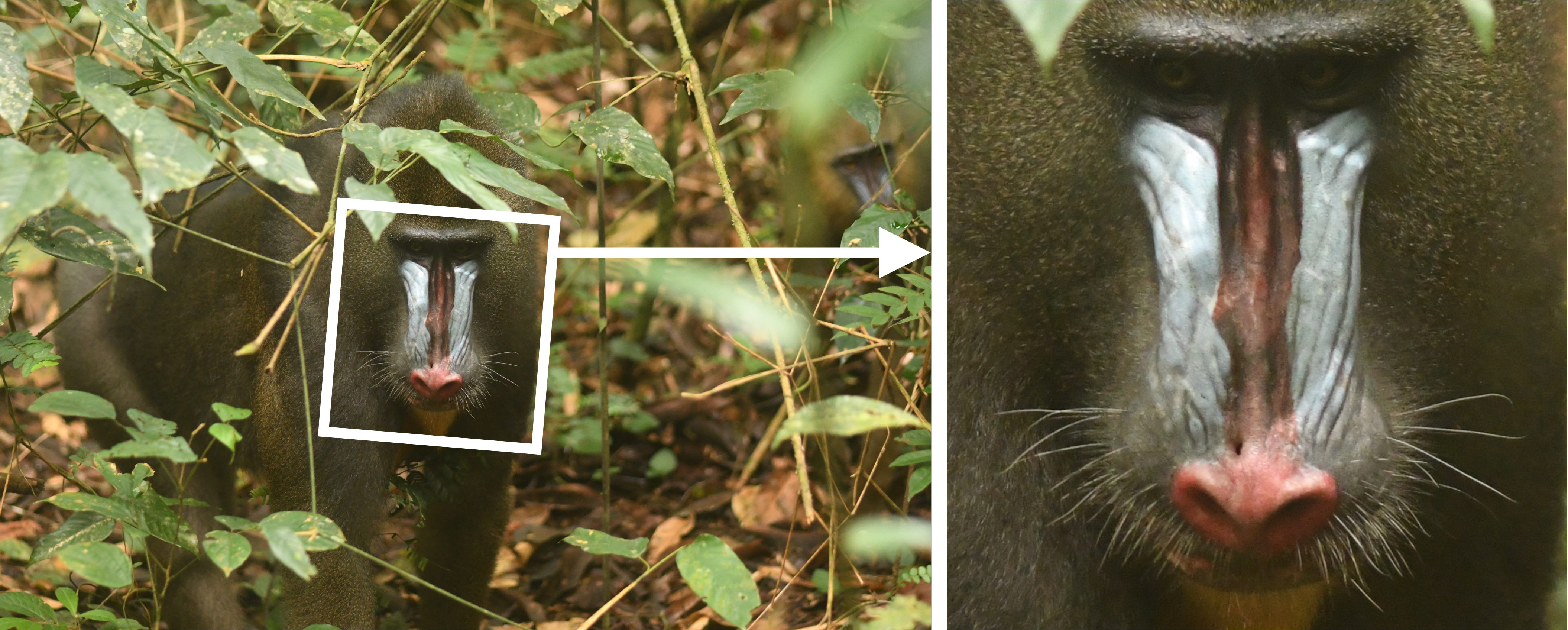}
    \caption{Mandrill \#20230104\_id203(2): a) In its environment, b) Cropped and straightened image of its face} 
 \label{fig:cropping1}
\end{figure}

\begin{figure}[hbtp!]
    \centering    
    \includegraphics[width=0.48\textwidth]{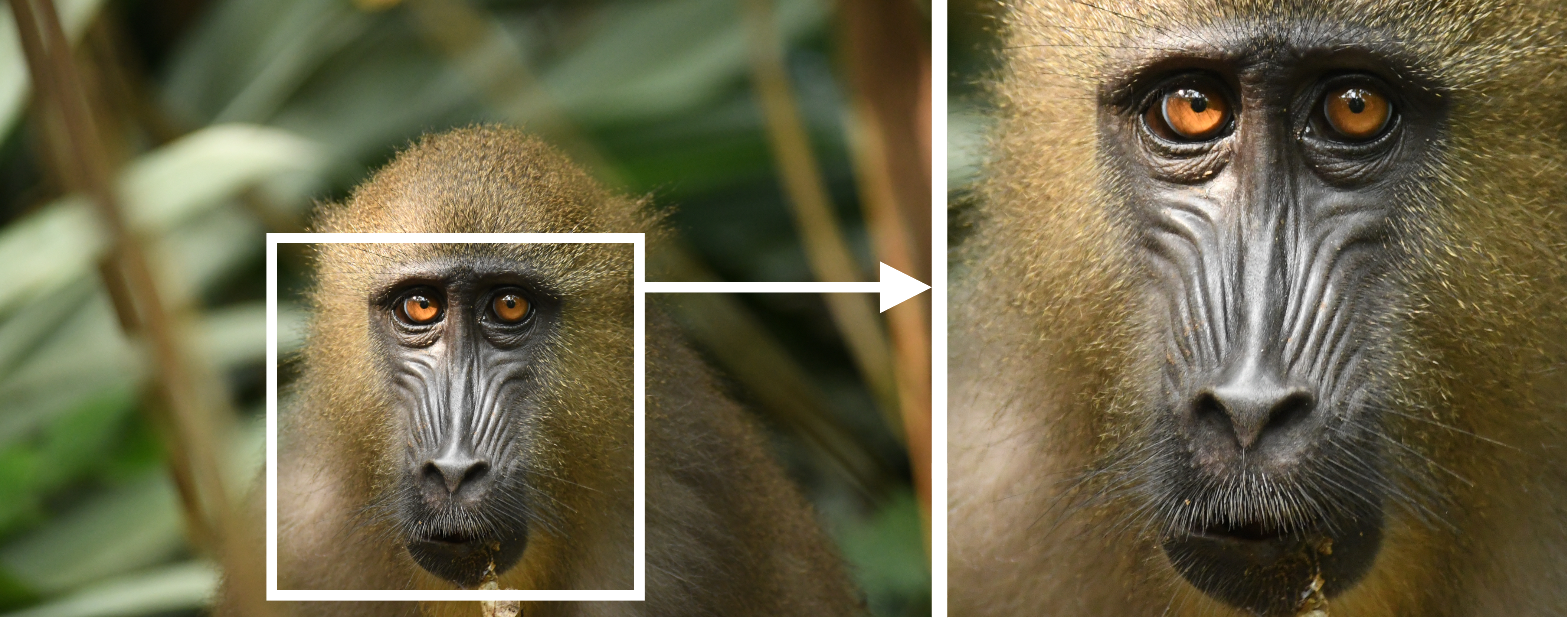}
    \caption{Mandrill \#20230104\_id205(5): a) In its environment, b) Cropped and straightened image of its face.} 
%    \caption{Straightening and cropping of a mandrill face from a full picture of a mandrill in its rainforest environment}
    \label{fig:cropping2}
\end{figure}

The Mandrillus Face Database (MFD)~\cite{TIEO2023108939} is one of the largest non-human animal face databases, as it contains 29495 images, representing 397 various mandrills, from pictures taken over  a period of 10 years. The images were taken by volunteers following a group of mandrills through a rainforest in the Lékédi Park and its surroundings, in southern Gabon (near to the village of Bakoumba).
As shown in Fig.~\ref{fig:cropping1} and Fig.~\ref{fig:cropping2}, from a full picture of a mandrill in its natural rainforest environment, the mandrill faces are manually aligned and cropped. Images are first manually oriented to align the pupils of the eyes horizontally, and then centered and cropped to keep only the face, the neck and ears are then removed. 

In our work, to train our machine learning algorithms, we use a database subset corresponding to good or high quality images of 4820 adult male and 13538 female mandrill faces, making a total of 18358 images. Good quality and high quality images correspond to \say{Quality 2} and \say{Quality 3} images respectively in the MFD database~\cite{TIEO2023108939}.

%%%%%%%%%%%%%%%%%%%%%%%%%%%%%%%%%%%%%%%%%%%%%%%%%%%%%%%
\subsection{Detailed examples, from real mandrill faces to edited generated mandril faces}
\label{subsec:étapes}

\begin{figure}[hbtp!]
\center
\begin{tabular}{ccccc}
 &\scriptsize{\#20191028\_id102\_femadu\_(1)}  & \scriptsize{\#20220629\_id214\_(2)} & \scriptsize{\#20220701\_id21} & \scriptsize{\#20220709\_id41}  \\   
    a) &\includegraphics[scale=0.08]{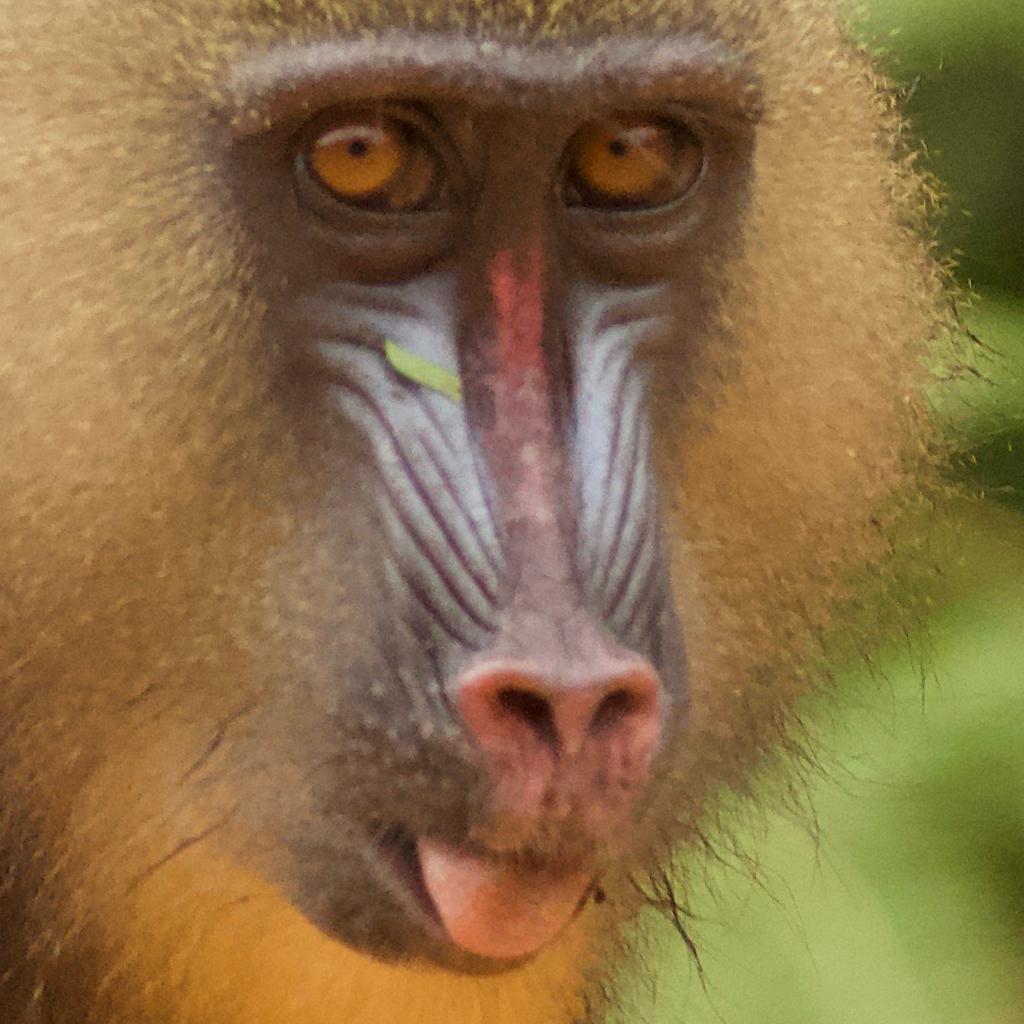} & 
    \includegraphics[scale=0.08]{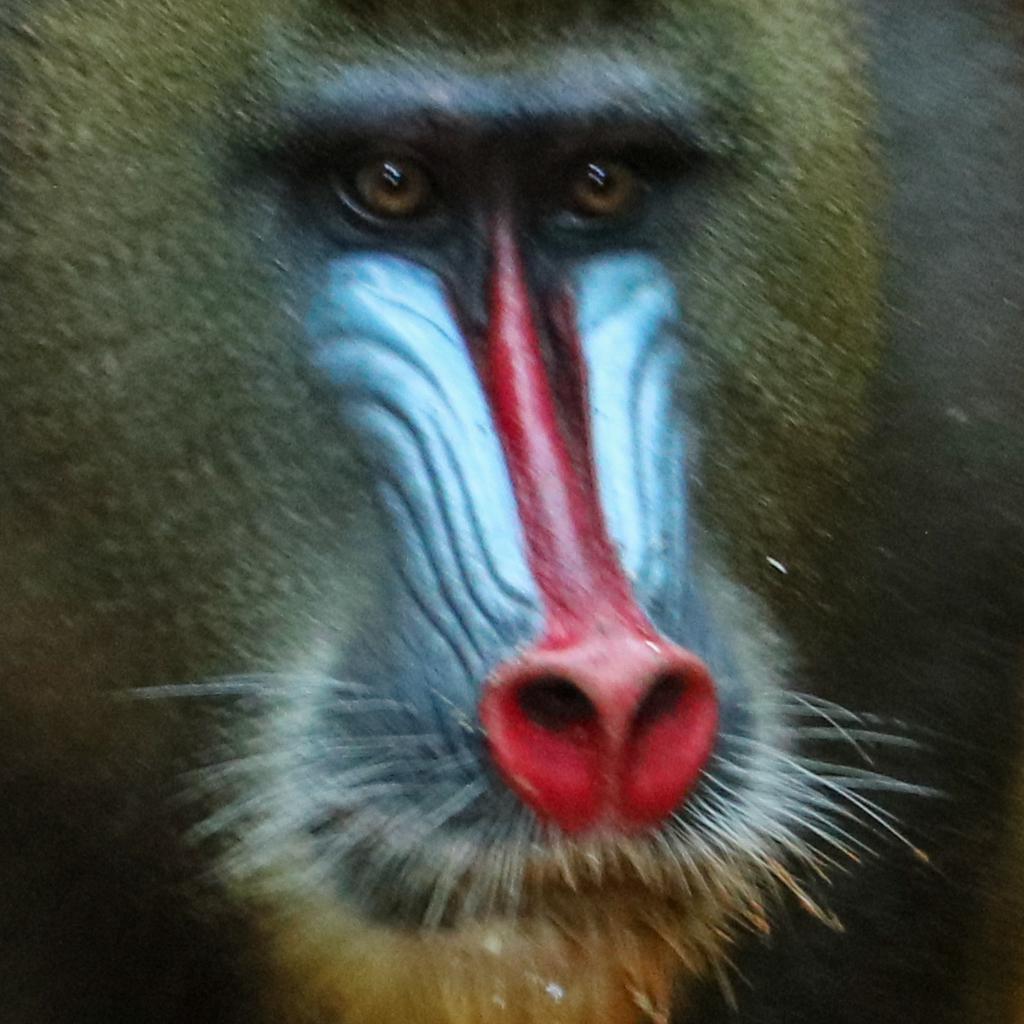} &
    \includegraphics[scale=0.08]{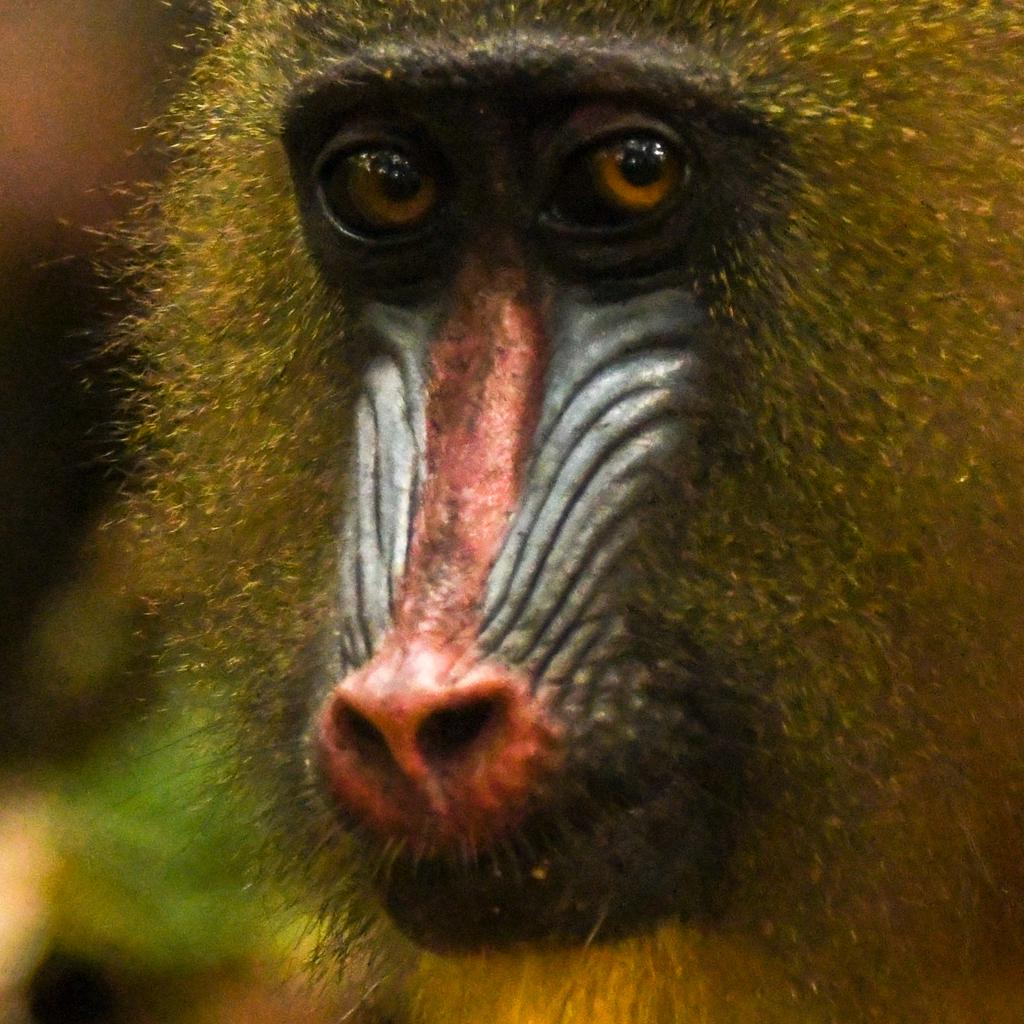} & 
    \includegraphics[scale=0.08]{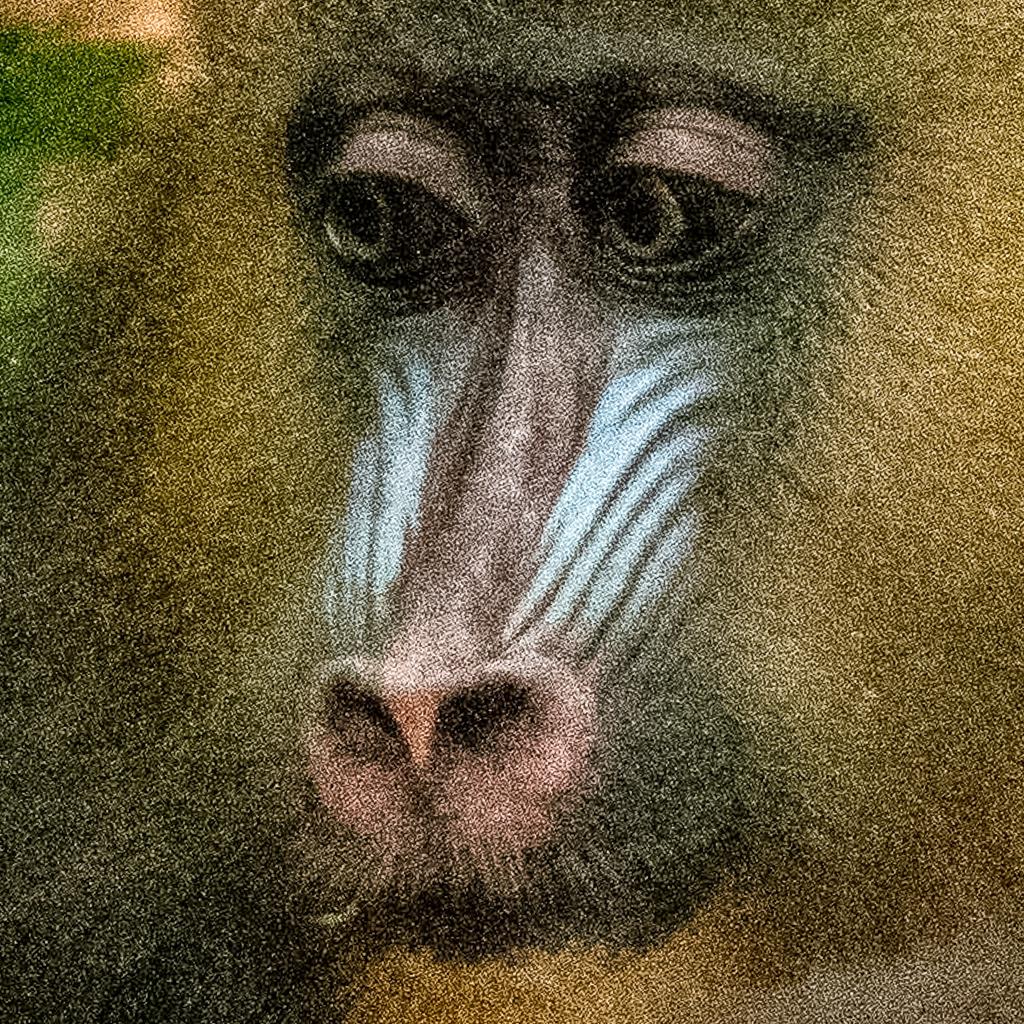}\\
    b) &\includegraphics[scale=0.108]{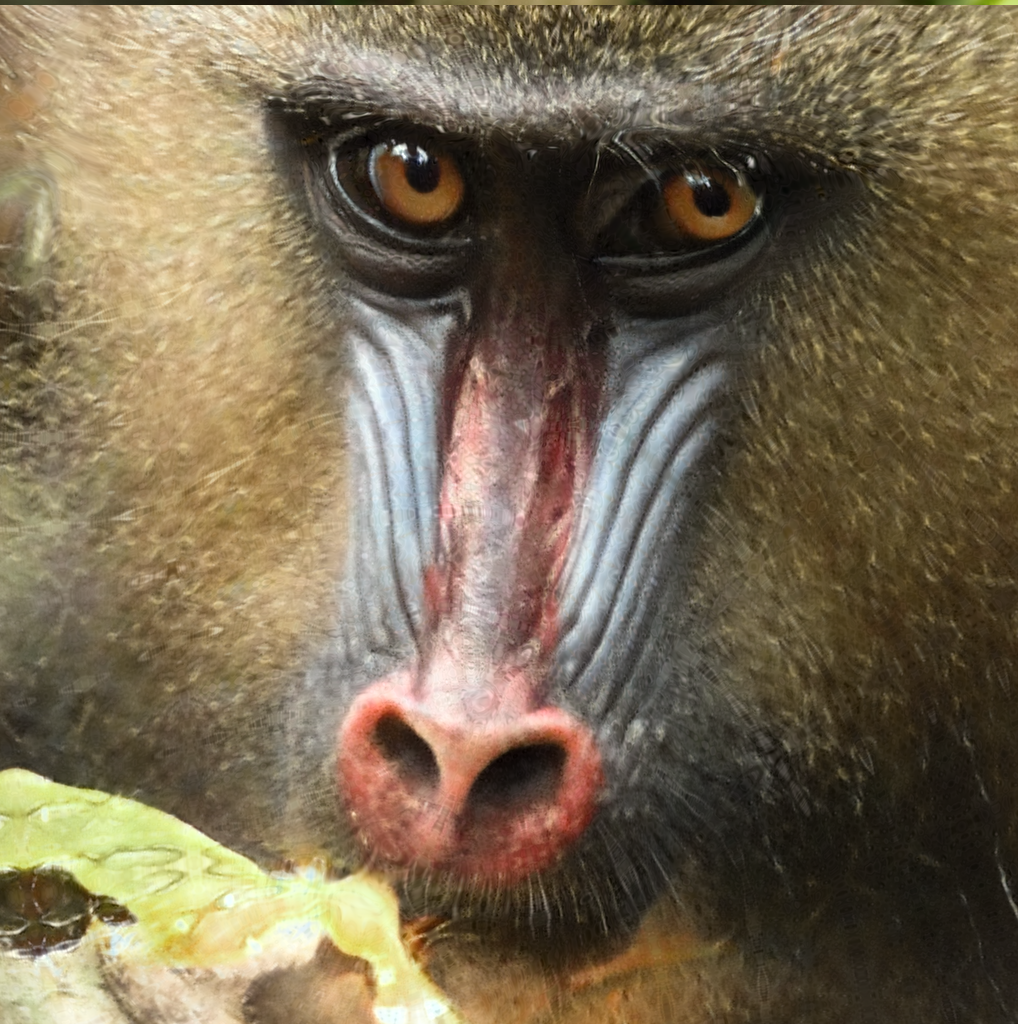} & 
    \includegraphics[scale=0.108]{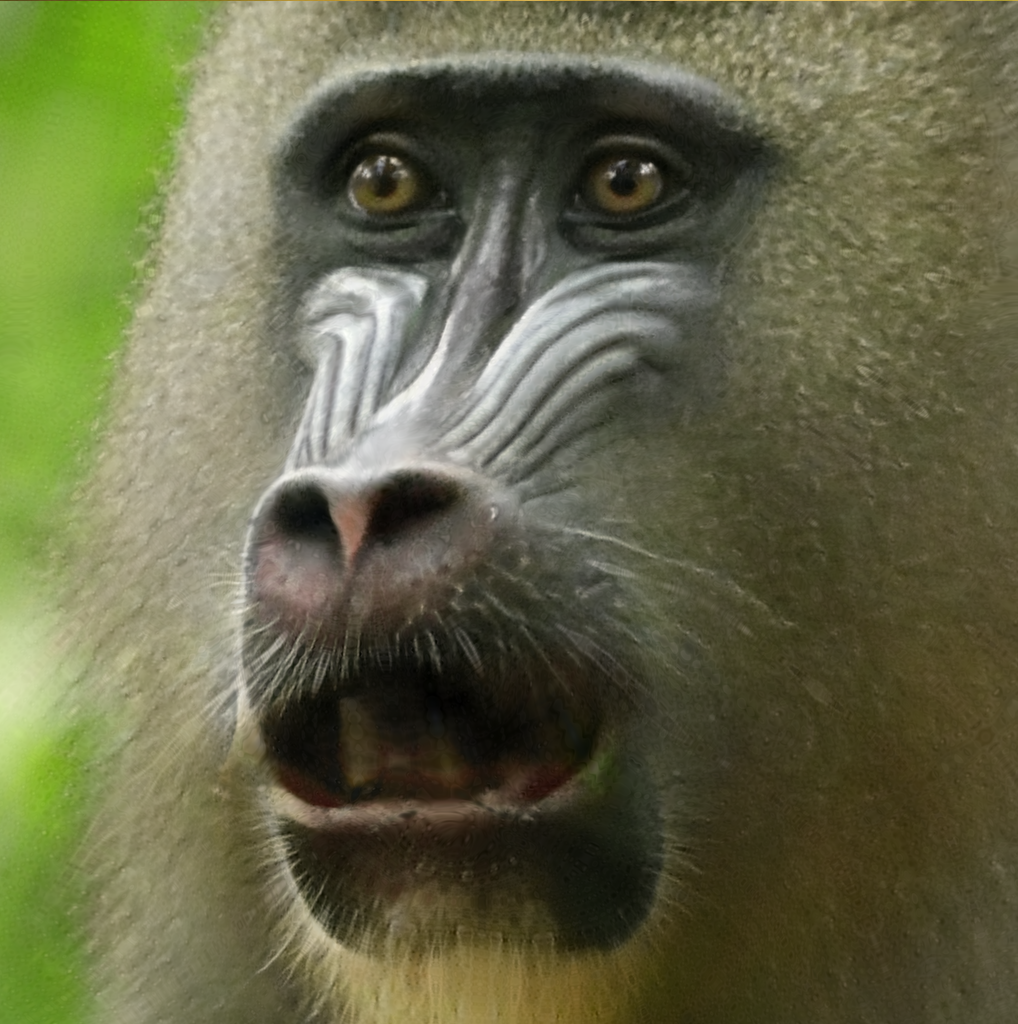} &
    \includegraphics[scale=0.108]{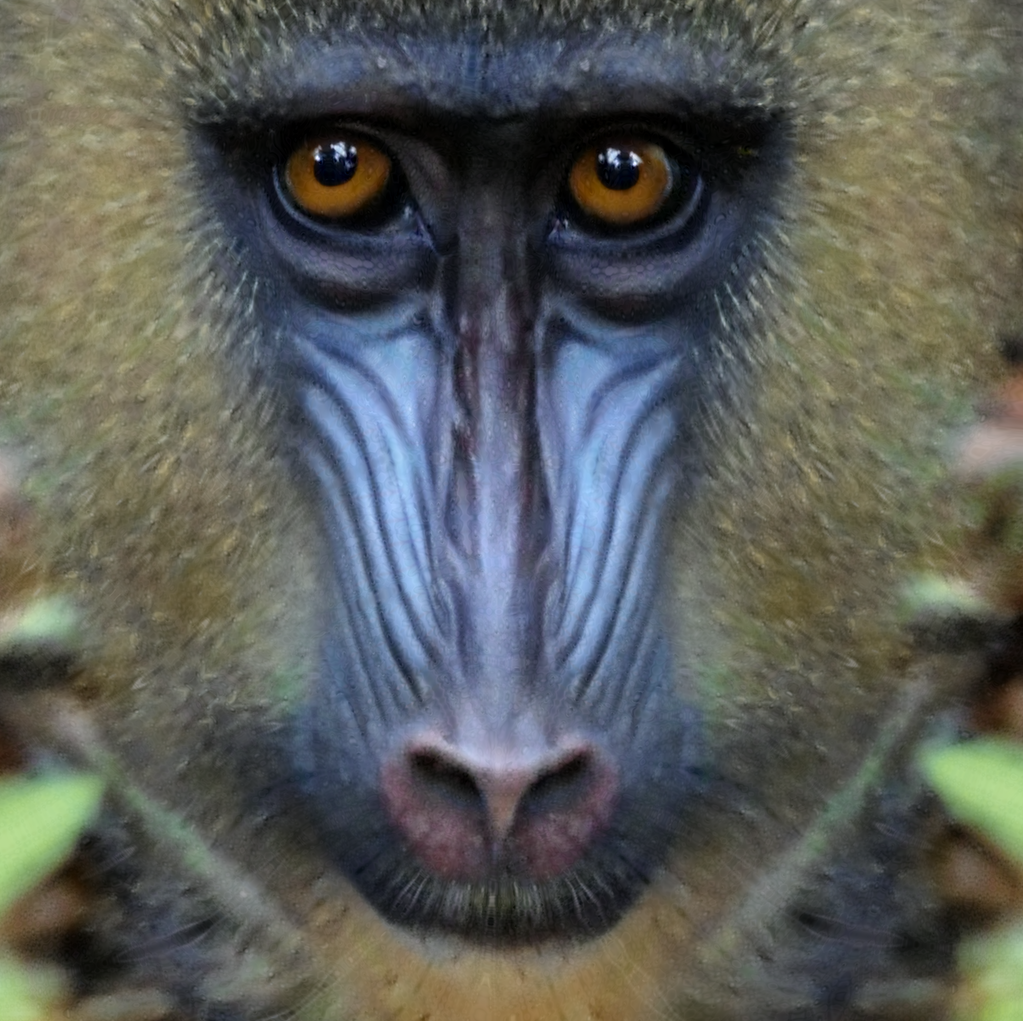} & 
    \includegraphics[scale=0.108]{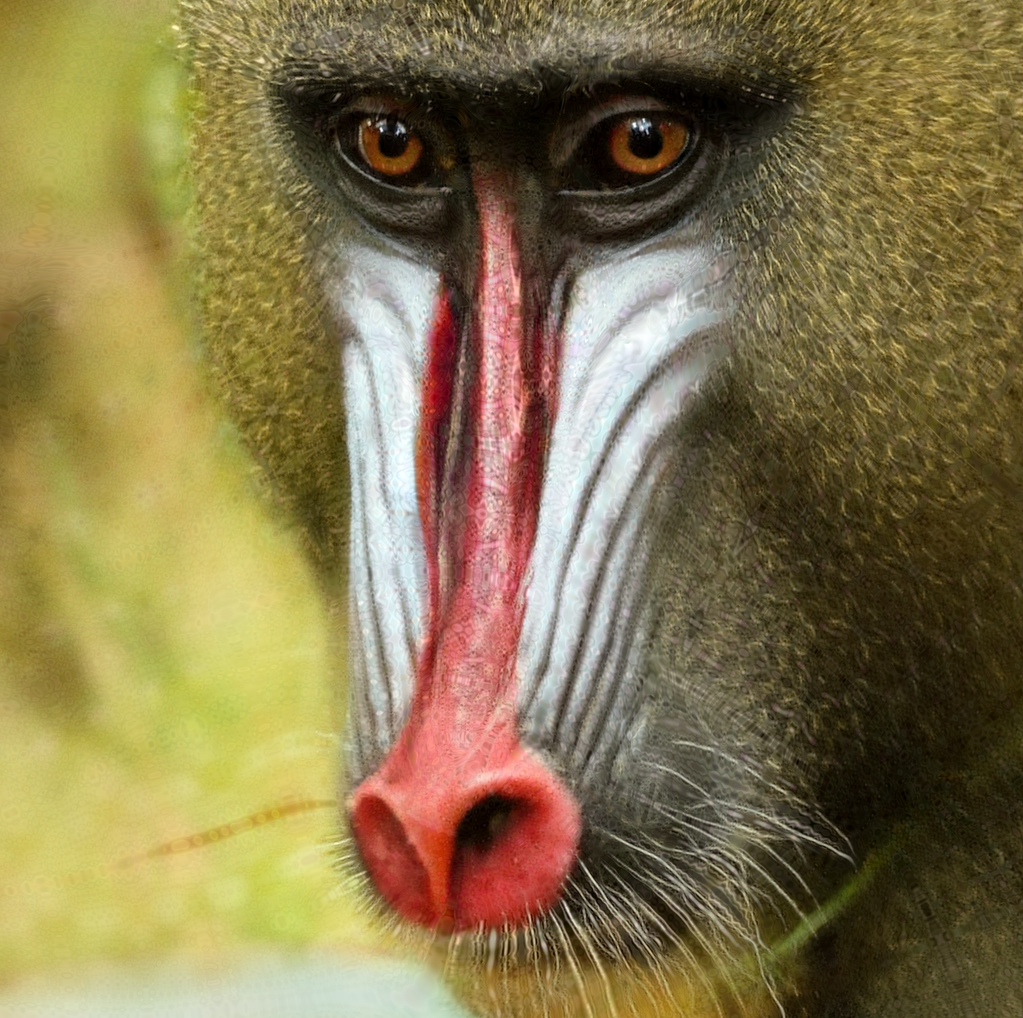}
\end{tabular}    
\caption{Comparison between: a) Real mandrill faces from the MFD database~\cite{tieo:04240893}, and b) Synthetic non-existing mandrill faces generated from StyleGAN3-mandrill}
\label{fig:SyntheticMandrillFaces} 
\end{figure}

In this section, we detail the experimental results we obtained by our proposed method. First, the training step of the developped StyleGAN3-mandrill takes 110 epochs, freezing 11 layers, \textit{i.e.} keeping the pre-trained StyleGAN3 weights on FFHQ (human faces) for the first 11 layers of the GAN generator. Before this training, we have a $FID = 266.26$,  calculated between synthetic images generated by StyleGAN3 and real images of the MFD database subset. We stop our training when the images generated by the StyleGAN3-mandrill are visually plausible, as shown in Fig.~\ref{fig:SyntheticMandrillFaces}, and the FID no longer appears to be decreasing. At the end of training, the FID is 3.65. 
In Fig.~\ref{fig:SyntheticMandrillFaces}.a we show four real mandrill faces from the MFD database~\cite{tieo:04240893} while in Fig.~\ref{fig:SyntheticMandrillFaces}.b we illustrate four synthetic mandrill faces, that is faces of mandrills that do not exist, generated from StyleGAN3-mandrill. These synthetic images are generated from a random latent vector of a size 512, the number of dimensions of the space $W$. These synthetic mandrill faces, Fig.~\ref{fig:SyntheticMandrillFaces}.b, have been shown to experts and they were not able to distinguish the synthetic from the real.

\begin{figure}[hbtp!]
\center
\begin{tabular}{cc}
    \includegraphics[scale=0.08]{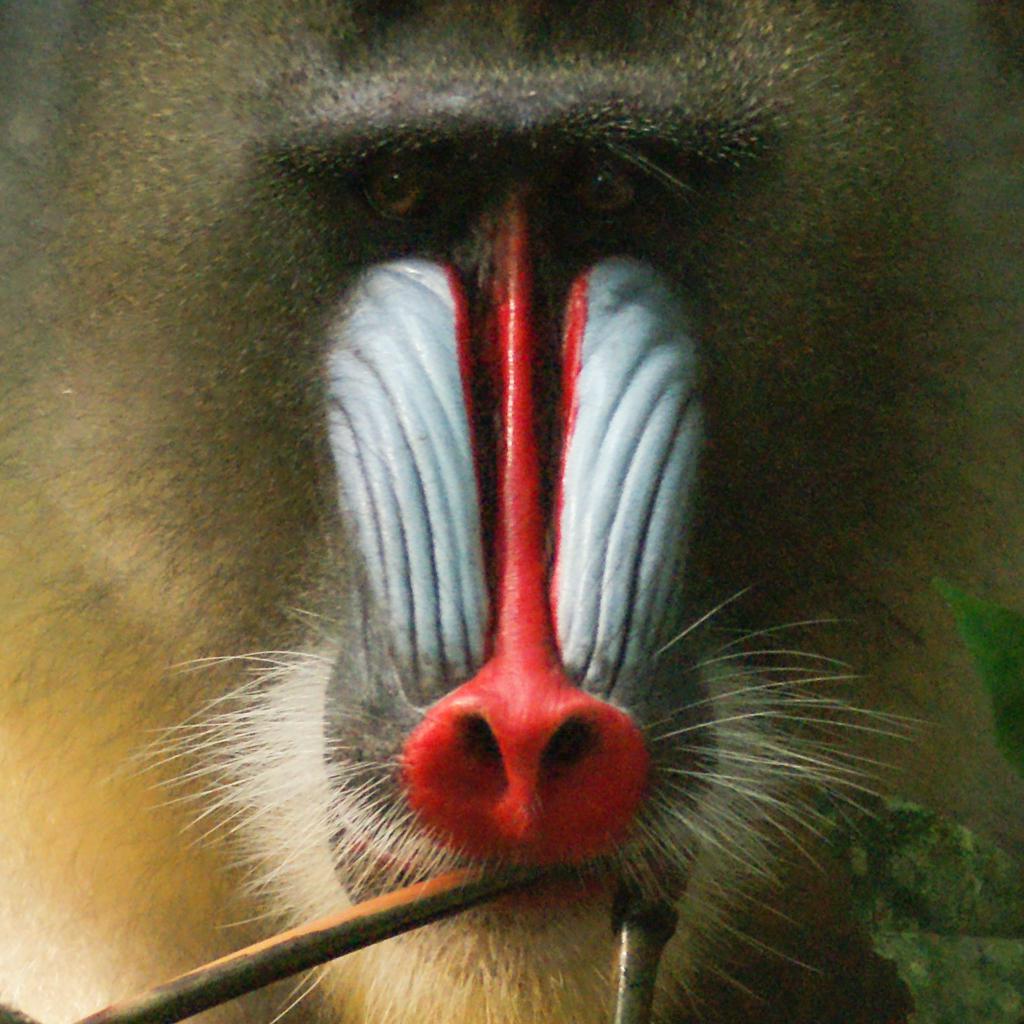} & 
    \includegraphics[scale=0.08]{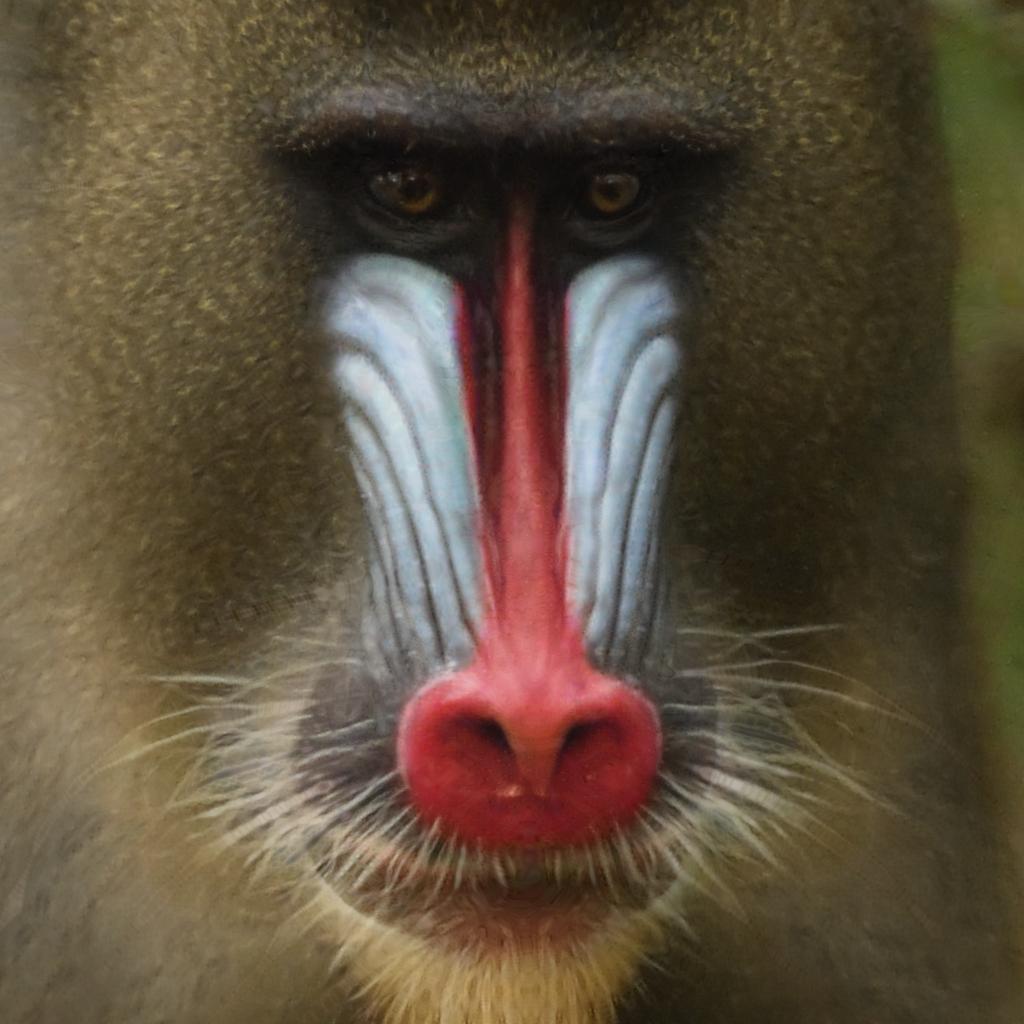} \\ 
      (a) &  (b) 
\end{tabular}    
\caption{Male mandrill face \#ID-57\_20150517(13)~\cite{TIEO2023108939}: a) Real image, b) Synthetic image generated from (a) encoded with pSp-mandrill then decoded with StyleGAN3-mandrill, with a sex level $S_o = 0.59 $.}
\label{fig:maleMandrillFace} 
\end{figure}

From a real image of a mandrill face, we can then generate its corresponding synthetic image with an encoding based on pSp-mandrill followed by a decoding with StyleGAN3-mandrill as presented in Section~\ref{sec:GenerationMandrillFaces}. Remember, face editing cannot be applied to a real image, but only to a generated one, in order to edit its latent vector. Fig.~\ref{fig:maleMandrillFace} illustrates the generation of a synthetic mandrill face from an image of a real mandrill face. From the real image of the male mandrill face \#ID-57\_20150517(13)~\cite{TIEO2023108939} shown in Fig.~\ref{fig:maleMandrillFace}.a, we can generate the corresponding synthetic image, shown in Fig.~\ref{fig:maleMandrillFace}.b, which has been encoded with pSp-mandrill to obtain its latent vector in the space $W+$, with a size 18 times larger than that of the space $W$ (size of 9216), and decoded with StyleGAN3-mandrill to visualize it. In the generated image Fig.~\ref{fig:maleMandrillFace}.b, we can see that the branch under the nose of the mandrill disappears during this process, and that colors and shadows are slightly modified. This is because the encoder positions the image in the space $W+$, according to the distribution of features found in the MFD database subset and learned by StyleGAN3-mandrill. As the features of an atypical object like a branch are poorly represented in this distribution, such features disappear during the encoding step. Note that the decoded image has not been modified in any way. To improve the preservation of shadows and colors, a histogram specification step could be implemented. After the projection on the sex axis, the sex level of this generated mandrill face image, is determined as sex level $S_o = 0.59 $.  

\begin{figure}[hbtp!]
    \centering
        \includegraphics[width=7.5cm]{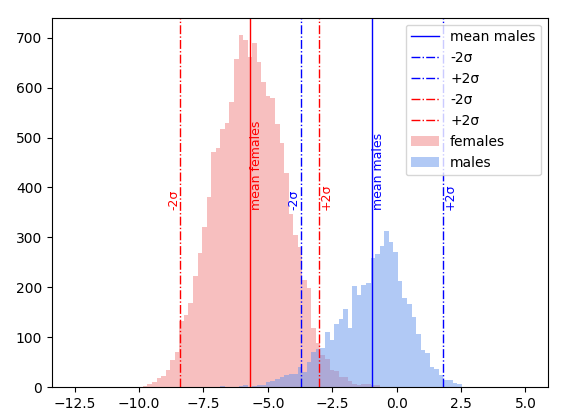}
        \caption{Distribution of the real mandrill faces from the MFD database subset~\cite{tieo:04240893} projected on the sex axis, with the positions of the sex level means ($\mu_F = -5.70$ and $\mu_M =-0.96$) and the upper and lower bounds (between [-8.40, -2.99] for females and between [-3.72, 1.81] for males)}.  
    \label{fig:histo_projection}
\end{figure}

As shown in Fig.~\ref{fig:histo_projection}, all the images used in the MFD database subset are encoded with pSp-mandrill and projected onto the sex axis. This allows us to obtain the distribution of their sex level on this axis. For the projection, as assumed in Section~\ref{sec:Editing}, the sex axis is the same for both males and females. Even with this assumption, we can see on Fig.~\ref{fig:histo_projection} that the two distributions are relatively easy to seperate, and that on this axis, the sex levels for females are generally lower than those for males. Indeed, for the female distribution, from 13538 female mandrill faces, the mean sex level is $\mu_F = -5.70$, with a  standard deviation of $\sigma_F=1.35$, while for the male distribution, from 4820 images,  the mean sex level is $\mu_M =-0.96$, with a standard deviation is $\sigma_M=1.38$.

%males
\begin{figure}[hbtp!]
\center
\begin{tabular}{cccc}
    \includegraphics[scale=0.08]{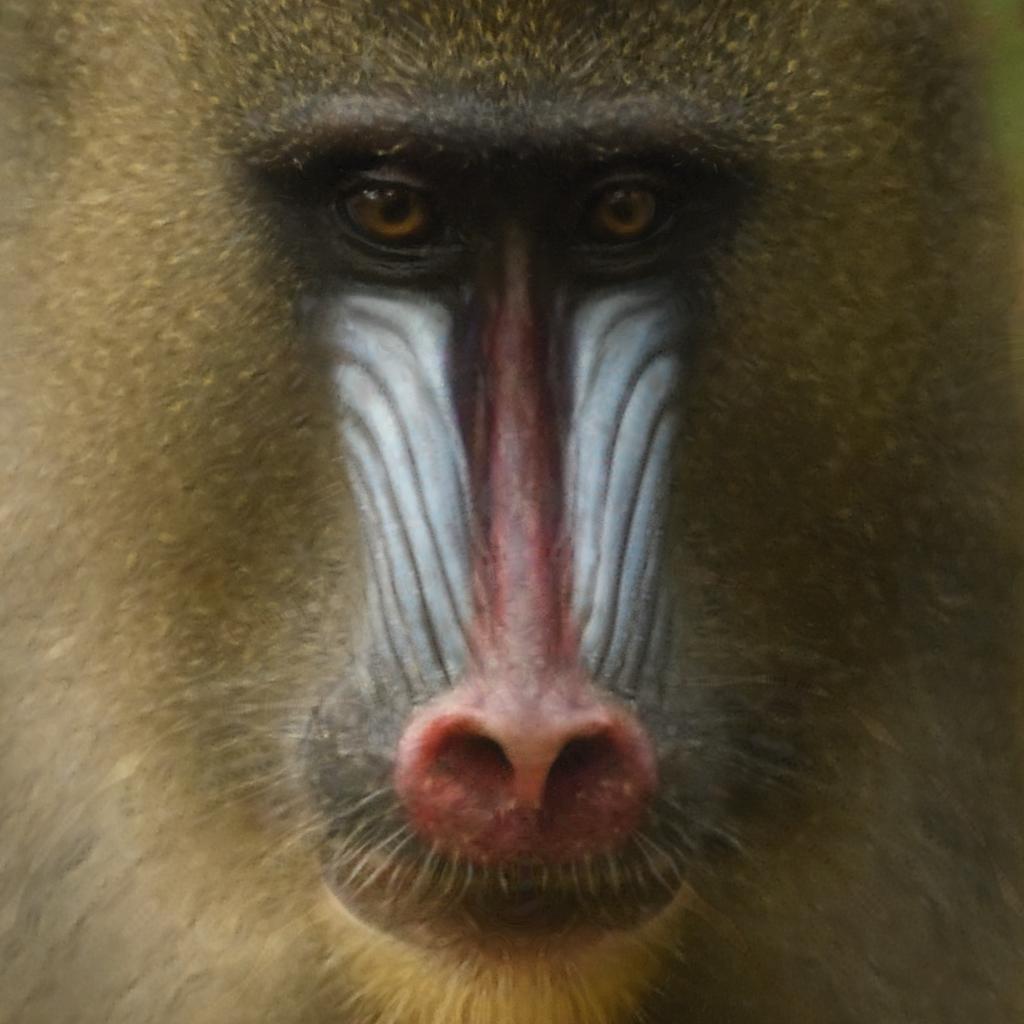} & 
    \includegraphics[scale=0.08]{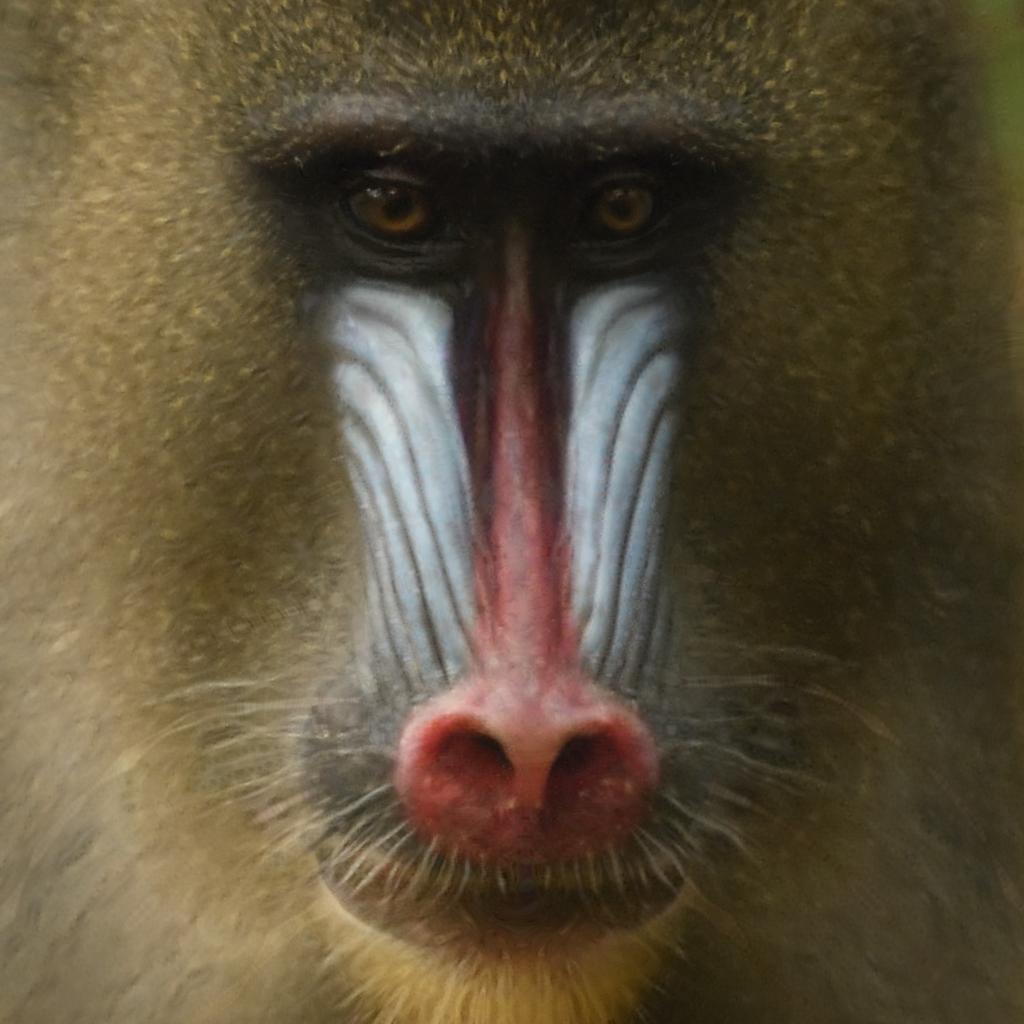} &
    \includegraphics[scale=0.08]{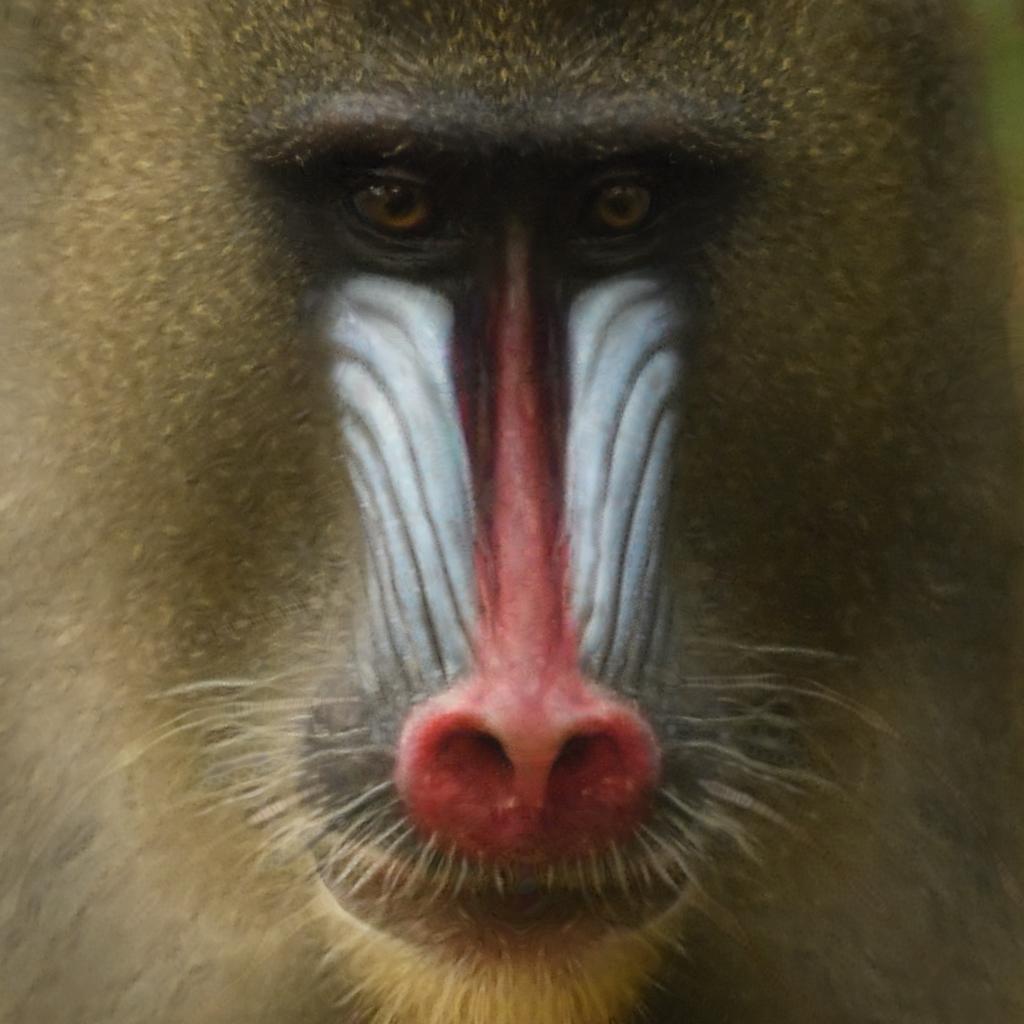} & 
    \includegraphics[scale=0.08]{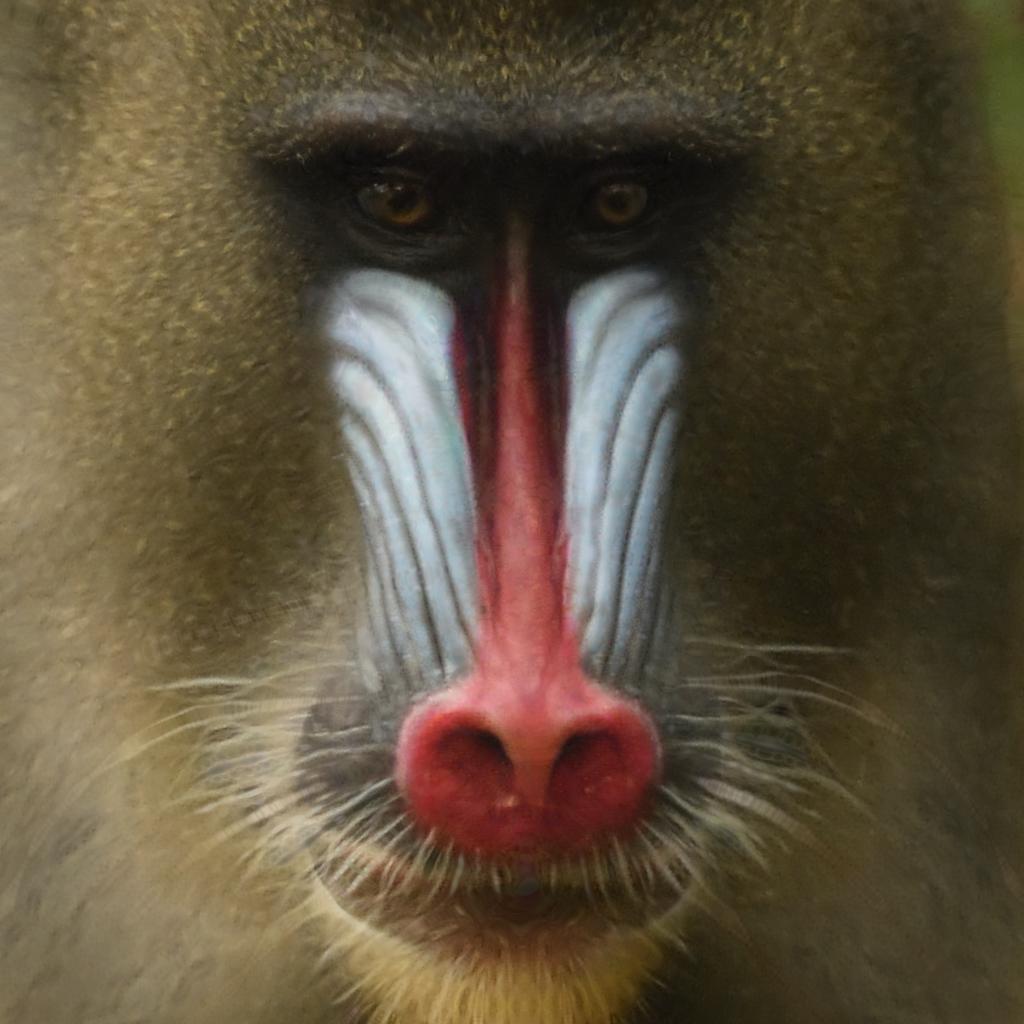}\\
    $S_r = -3.35$ & $S_r = -2.66$ & $S_r = -1.97$ & $S_r = -1.28$ \\
     $S_e = -3.44$, $i=11$ & $S_e = -2.78$, $i=5$ & $S_e = -1.99$, $i=5$ & $S_e = -1.39$, $i=6$ \\ 
     (a) &  (b) & (c) & (d) \\ 
    \multicolumn{1}{|c|}{\includegraphics[scale=0.08]{Images/57encodage_0_score-0.589.jpg}} & 
    \includegraphics[scale=0.08]{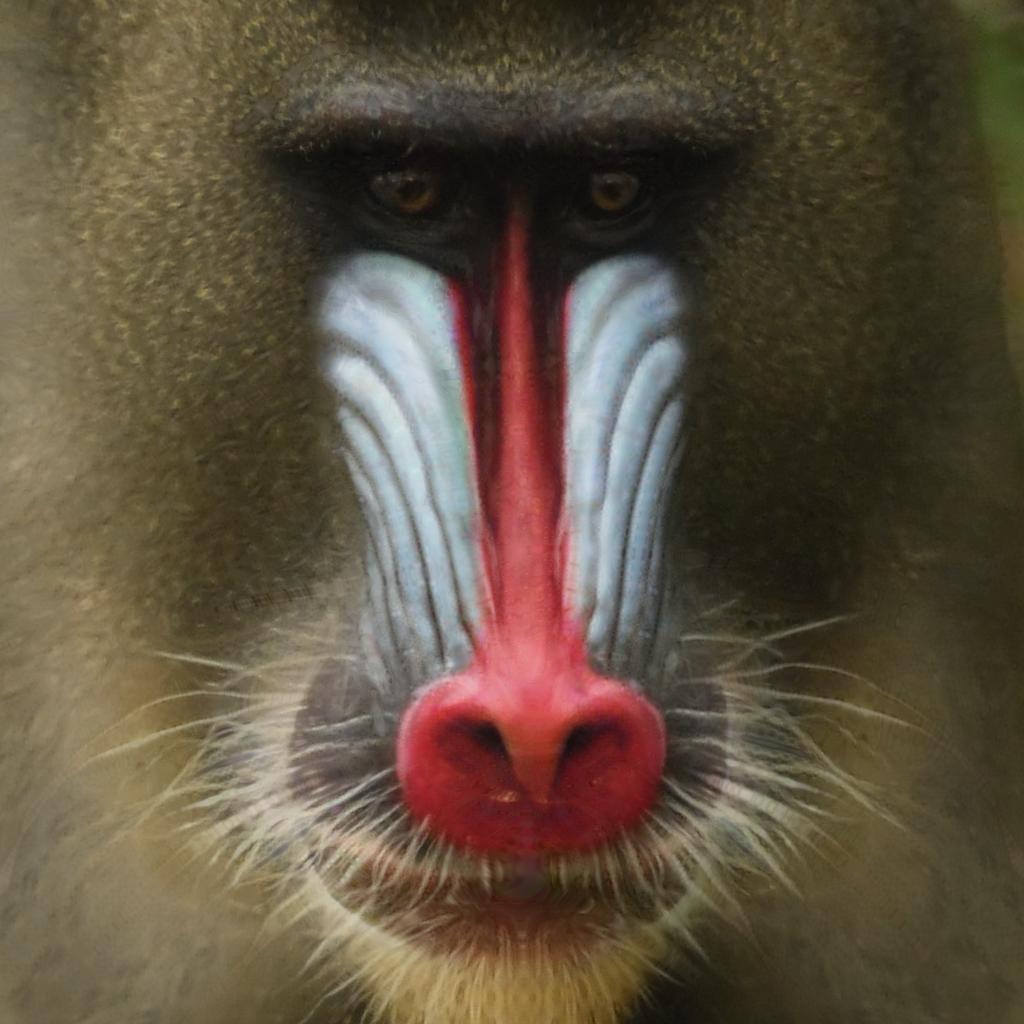} &
    \includegraphics[scale=0.08]{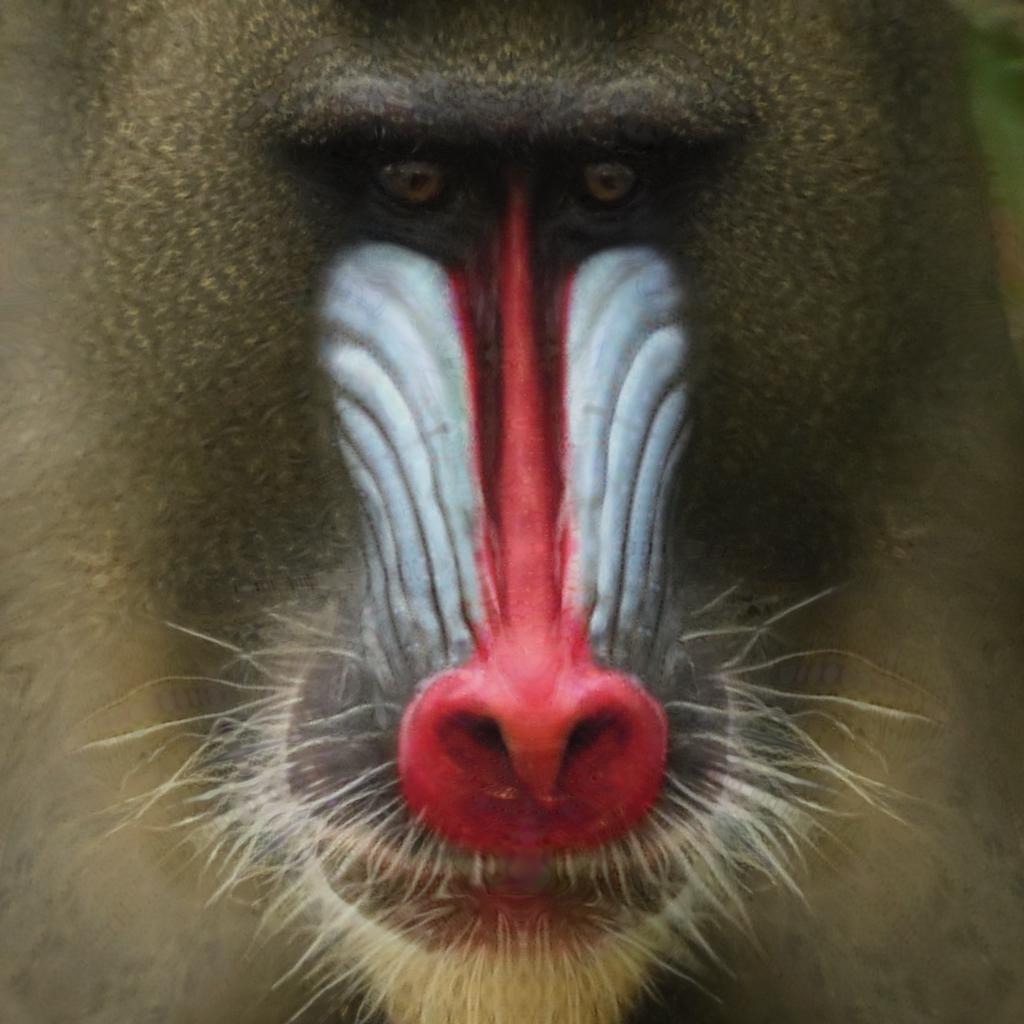} & 
    \includegraphics[scale=0.08]{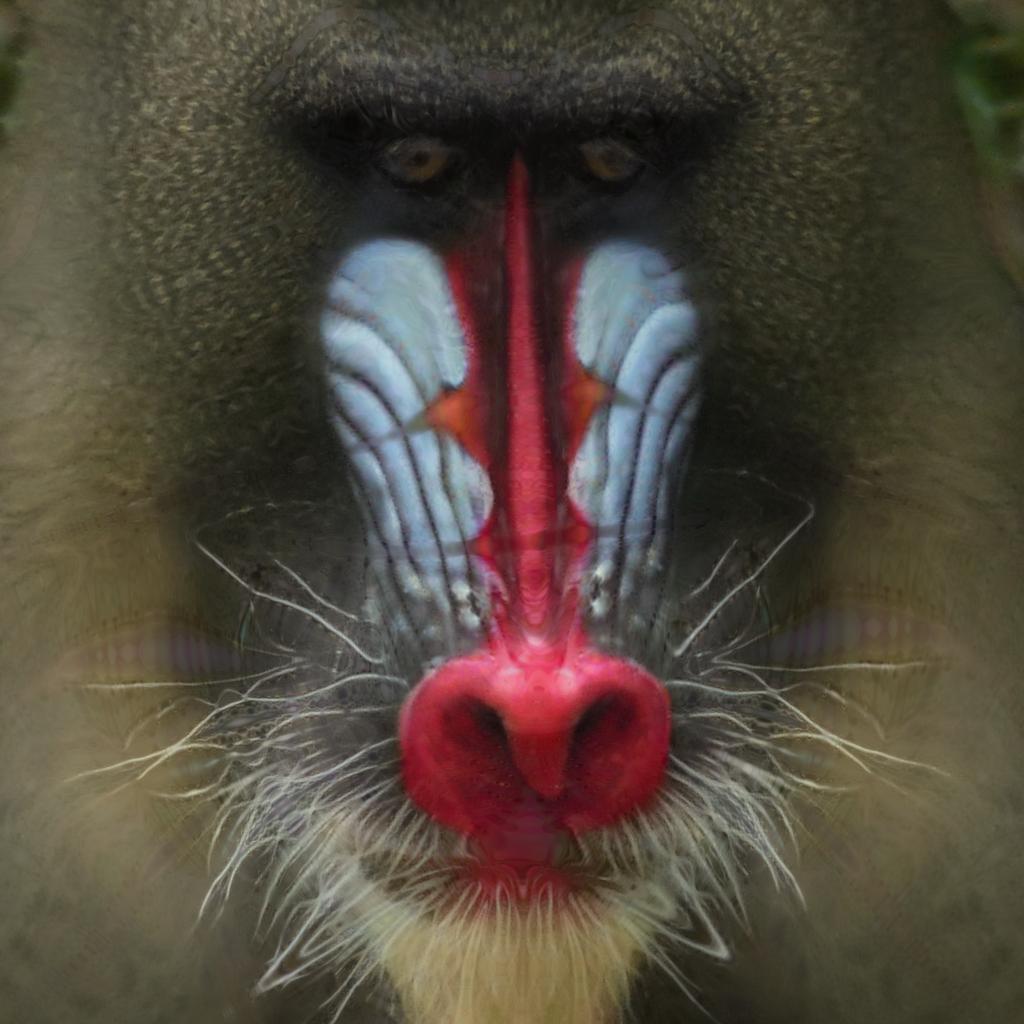}\\
    \multicolumn{1}{|c|}{$S_o = 0.59$} & $S_r = 0.10$ & $S_r = 0.79$ & $S_r = 1.49$ \\
    \multicolumn{1}{|c|}{Original} & $S_e = 0.02$, $i=5$ & $S_e = 0.87$, $i=5$ & $S_e = 1.57$, $i=9$ \\
    \multicolumn{1}{c}{(e)} & (f) & (g) & (h)
\end{tabular}    
\caption{Editing of the male mandrill face \#-57\_20150517(13)~\cite{TIEO2023108939} from the generated image illustrated~Fig.~\ref{fig:maleMandrillFace}.b : a) Edited generated image with a sex level $S_e = S_o -2 \sigma_M = -3.44$, b) Edited generated image with a sex level $S_e = S_o -1.5 \sigma_M = -2.78$, c) Edited generated image with a sex level $S_e = S_o - \sigma_M = --1.99$, d)Edited generated image with a sex level $S_e = S_o -0.5 \sigma_M = -1.28$,  e) Original encoded image (Fig.~\ref{fig:maleMandrillFace}.b) with a sex level $S_o = 0.59 $, f) Edited generated image with a sex level $S_e = S_o + 0.5\sigma_M = 0.02$, g) Edited generated image with a sex level $S_e = S_o + \sigma_M = 0.81$, h) Edited generated image with a sex level $S_e = S_o + 1.5 \sigma_M = 1.57$. For each image, $i$ is the number of iterations in Algorithm~\ref{algo:alg2} to obtain the desired deviation $\Delta_d$.}
\label{fig:EditedMaleMandrillFace8} 
\end{figure}

In the rest of this section we present the results obtained at the different steps of editing real images of mandrill faces, with two examples, one of a male and one of a female. 
Images are then edited at the desired deviation $\Delta_d$. In Fig.~\ref{fig:EditedMaleMandrillFace8} we illustrate an example with the male mandrill face generated and illustrated in Fig.~\ref{fig:maleMandrillFace}.b. From this generated male mandrill face, illustrated in Fig.~\ref{fig:EditedMaleMandrillFace8}.e. we choose deviations $\Delta_e = -2 \sigma$, in Fig.~\ref{fig:EditedMaleMandrillFace8}.a, $\Delta_e = -1.5\sigma$, in Fig.~\ref{fig:EditedMaleMandrillFace8}.b, $\Delta_e = -\sigma$, in Fig.~\ref{fig:EditedMaleMandrillFace8}.c, $\Delta_e = -0.5\sigma$, in Fig.~\ref{fig:EditedMaleMandrillFace8}.d, $\Delta_e = 0.5\sigma$, in Fig.~\ref{fig:EditedMaleMandrillFace8}.f, $\Delta_e = \sigma$, in Fig.~\ref{fig:EditedMaleMandrillFace8}.g, and $\Delta_e = 1.5\sigma$, in Fig.~\ref{fig:EditedMaleMandrillFace8}.h. Note, that based on the conditions presented in Algorithm~\ref{algo:alg1}, it is not possible to go as far as $\Delta_e = -2.5 \sigma$ or $\Delta_e = 2 \sigma$ in order  not to exceed the bounds of $2 \sigma$ around the mean sex level $\mu_M$.

\begin{figure}[hbtp!]
\center
\begin{tabular}{c}
    \includegraphics[width=6.5cm]{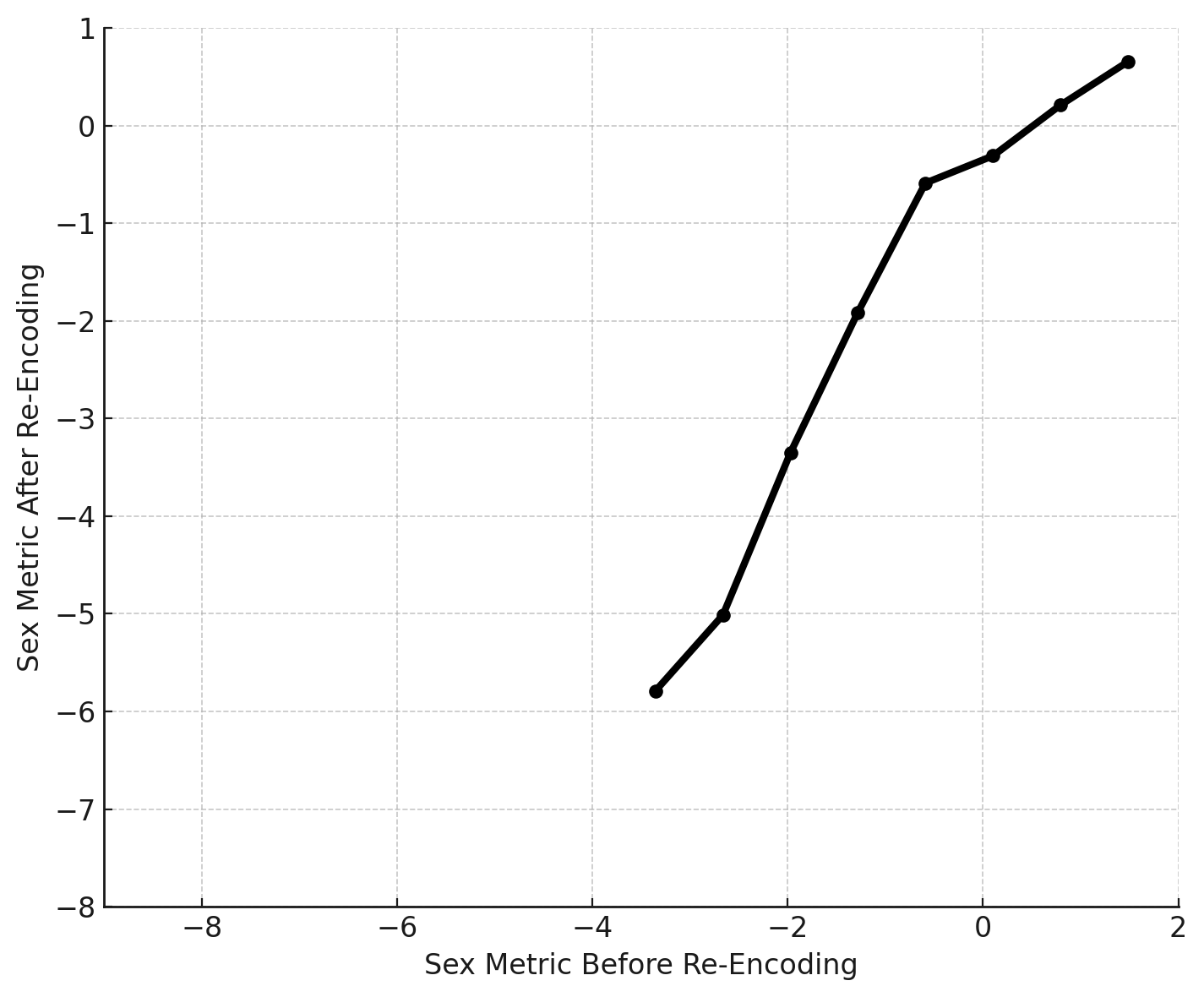} 
\end{tabular}    
\caption{Sex level editing as a function of the obtained sex levels after decoding and re-encoding for the image \#-57\_20150517(13) shown Fig.~\ref{fig:EditedMaleMandrillFace8}.}
\label{fig:maleMandrillcurve} 
\end{figure}

To obtain the edited images shown in Fig.~\ref{fig:EditedMaleMandrillFace8} with the optimized deviation $\Delta_e$, we used Algorithm~\ref{algo:alg2}. Experimentally, we set the value of the threshold $T = 0.1$. This threshold $T$ is multiplied by the standard deviation in the algorithm, giving a value of $0.138$. This value guarantees visual accuracy, knowing that a greater precision would be useless as it would be imperceptible to the naked eye, even for experts. What's more, the relationship between the sex levels of the edited images and the sex levels of the same images decoded and then re-encoded is not monotonic on a scale smaller than $T = 0.1$. Consequently, decreasing the value of threshold $T$ would have no impact on accuracy. We set the value of the step size $S =7$, which means that fewer iterations are needed than with a higher value of $S$, while avoiding diverging out of the solution space, which would have been the case with a lower value of $S$. For the editing, shown in Fig.~\ref{fig:EditedMaleMandrillFace8}, the number of iterations of Algorithm~\ref{algo:alg2} varies from $5$ to $11$ with these parameter values. By using Algorithm~\ref{algo:alg2}, the first value of the deviation $\Delta_e$ before optimization corresponds to the desired editing. There seems to be a monotonic correlation between this edition and the value corresponding to what we obtain after optimization, as can be seen in Fig.~\ref{fig:maleMandrillcurve} with several editing values. In Fig.~\ref{fig:maleMandrillcurve}, for the image \#-57\_20150517(13) shown in Fig.~\ref{fig:EditedMaleMandrillFace8}, we can see that for a range between $-2 \sigma$ and $1.5 \sigma$ for the sex level editing, we obtain values for the sex level editing after decoding then re-encoding in a range between $-6$ and $1$. Note that visually, the increased masculinity of an image corresponds to more pronounced features, more contrasting colors, a more elongated face and a redder nose, as shown in Fig.~\ref{fig:EditedMaleMandrillFace8}.

\begin{figure}[hbtp!]
\center
\begin{tabular}{cc}
    \includegraphics[scale=0.08]{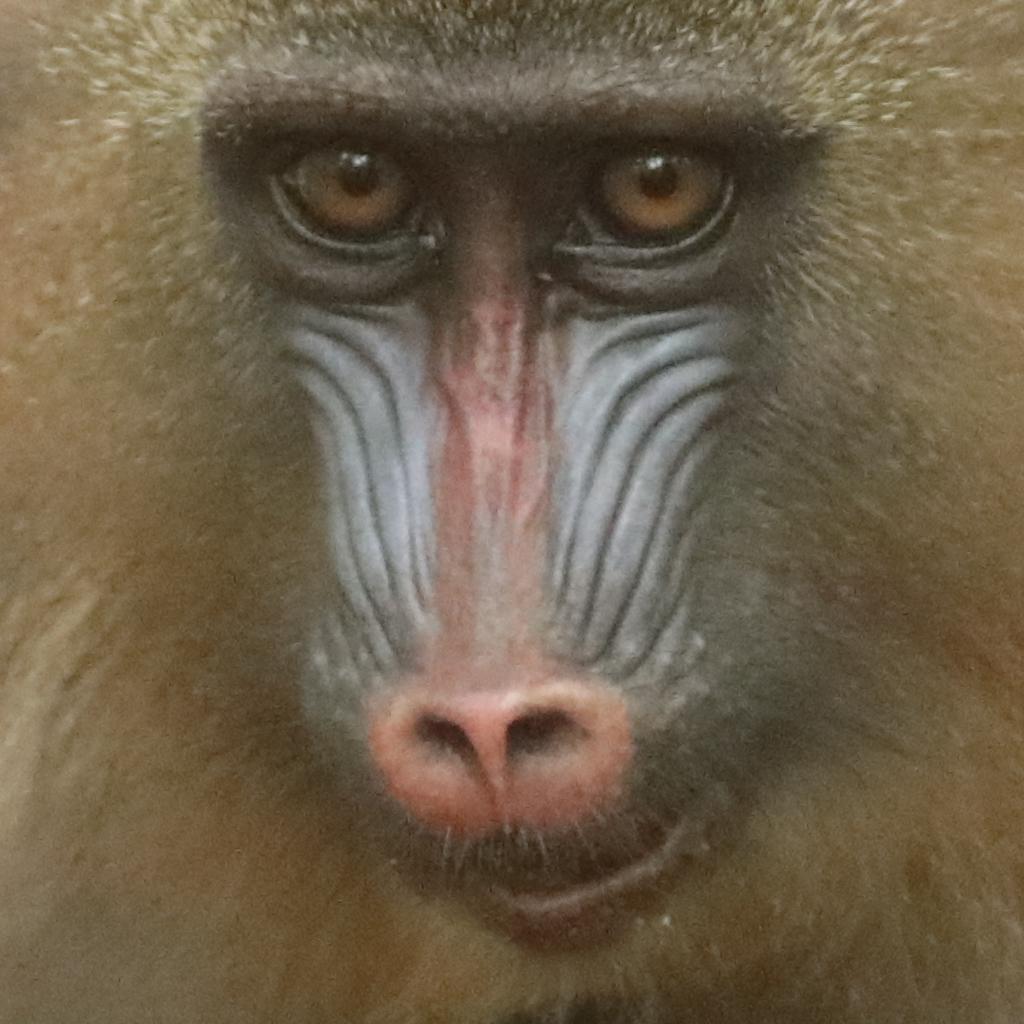} & 
    \includegraphics[scale=0.08]{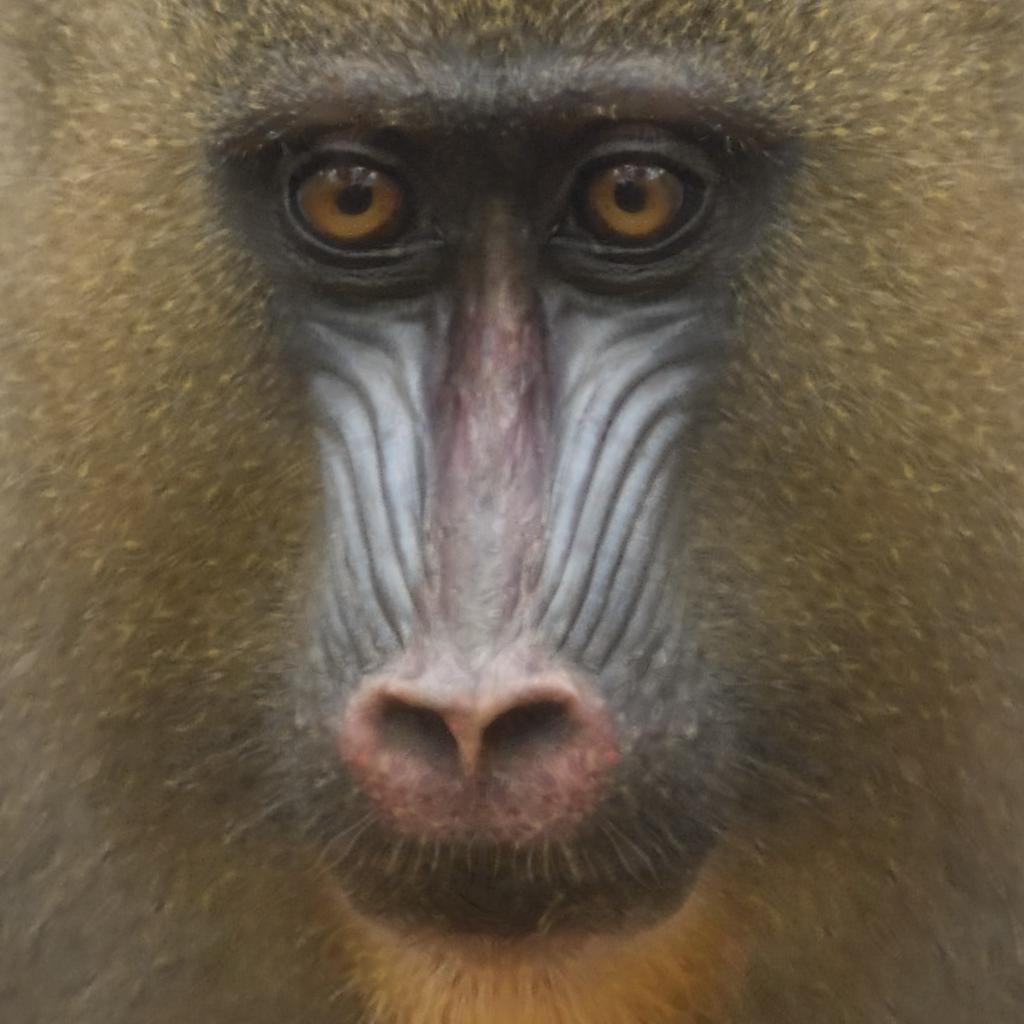} \\ 
      (a) &  (b) 
\end{tabular}    
\caption{Female mandrill face \#20210325\_id221\_femadu~\cite{TIEO2023108939}: a) Real image, b) Synthetic image generated from (a) encoded with pSp-mandrill then decoded with StyleGAN3-mandrill, with a sex level $S_o = -5.70 $.}
\label{fig:femaleMandrillFace} 
\end{figure}

%females
Similarly, as illustrated in Fig.~\ref{fig:femaleMandrillFace}, we have encoded an image of the female mandrill face\#20210325\_id221\_femadu, Fig.~\ref{fig:femaleMandrillFace}.a, and then encoded with pSp-mandrill then decoded it, Fig.~\ref{fig:femaleMandrillFace}.b, in the same way as the male mandrill example.
From the real image of the female mandrill face Fig.~\ref{fig:femaleMandrillFace}.a, note that the encoding, illustrated Fig.~\ref{fig:femaleMandrillFace}.b modifies the inclination of the head and makes the eyes rounder. this is again a result of feature normalization as observed with the example illustrated Fig. 9.~\ref{fig:maleMandrillFace}. 

\begin{figure}[hbtp!]
\center
\begin{tabular}{cccc}
    \includegraphics[scale=0.08]{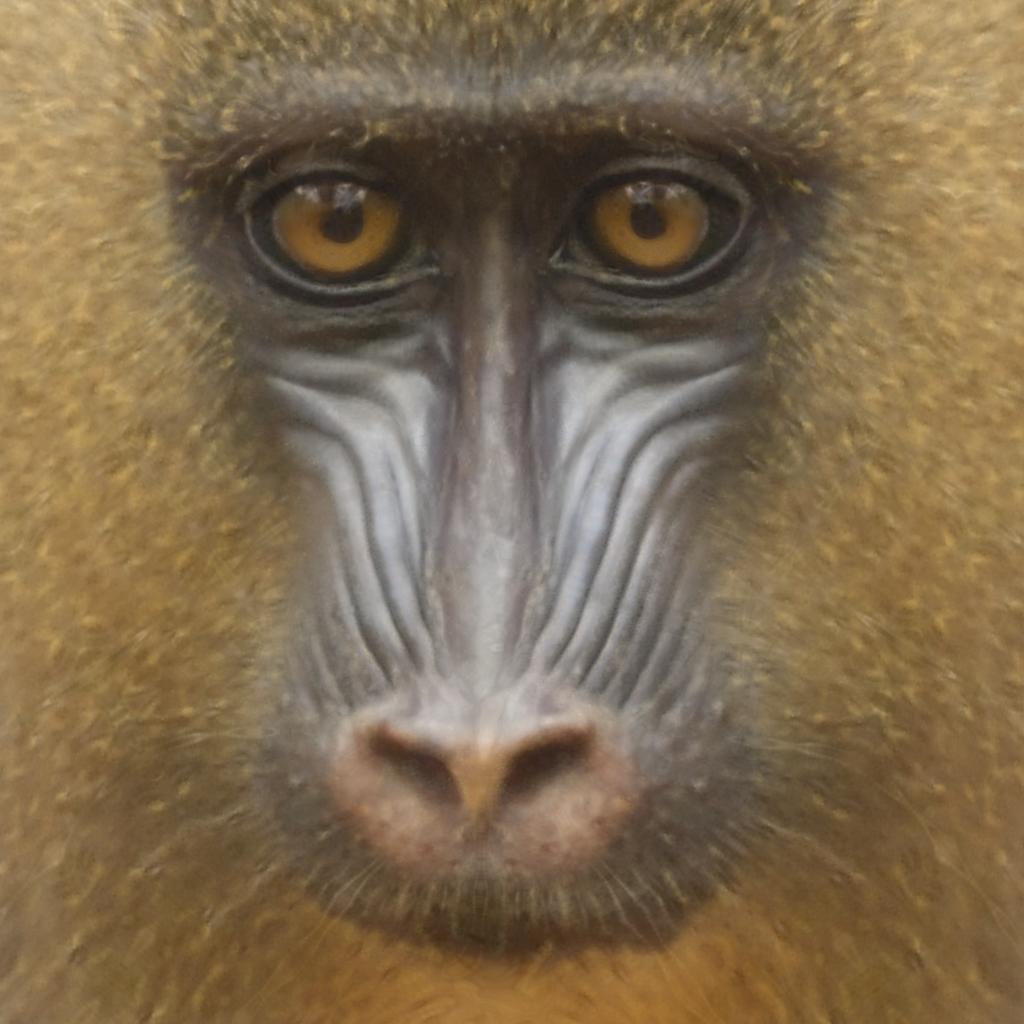} & 
    \includegraphics[scale=0.08]{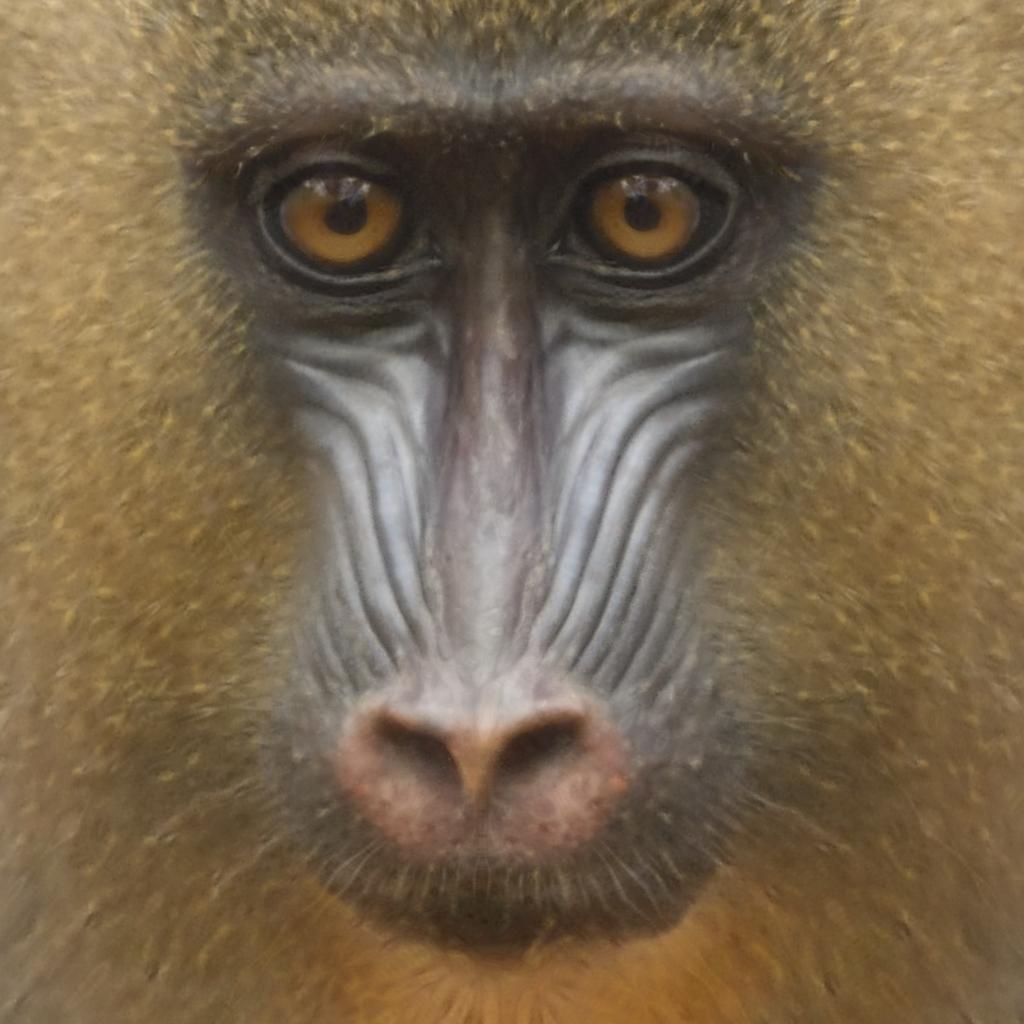} &
    \includegraphics[scale=0.08]{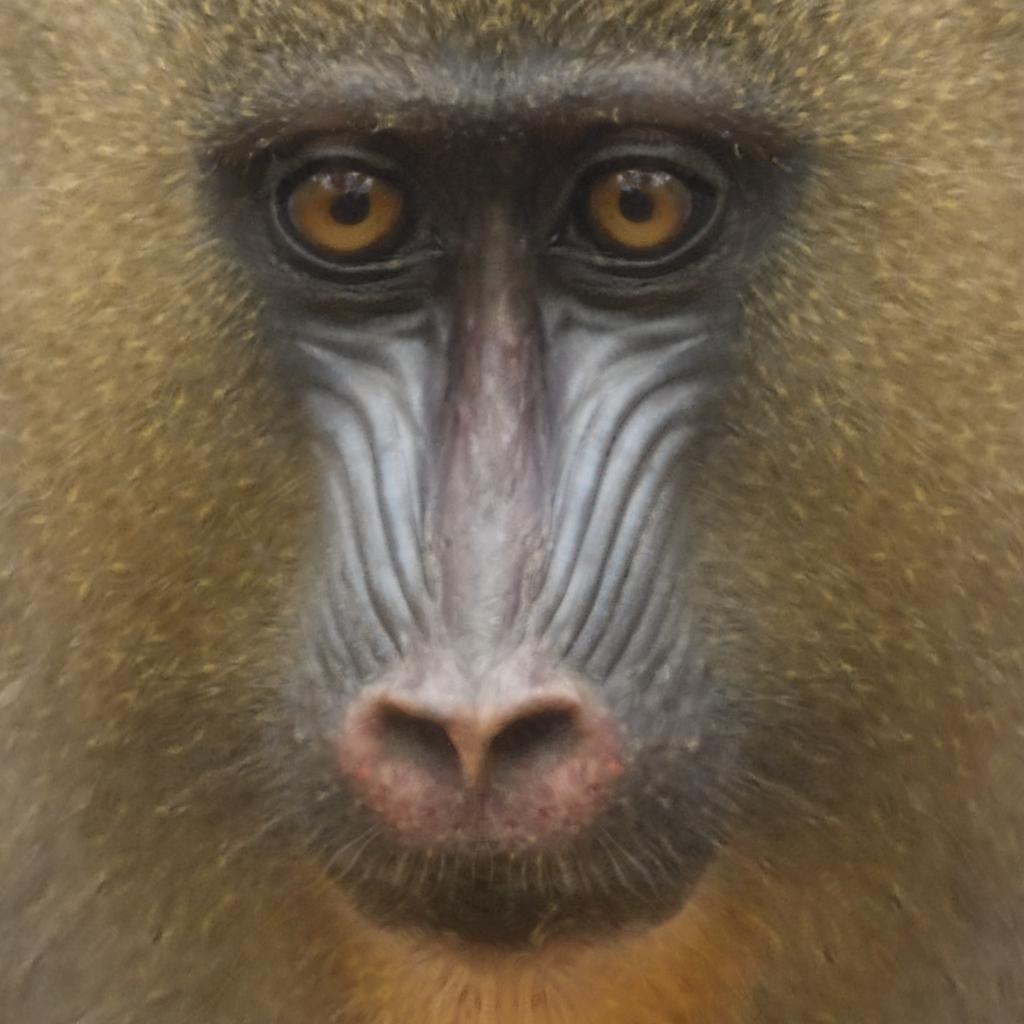} & 
    \includegraphics[scale=0.08]{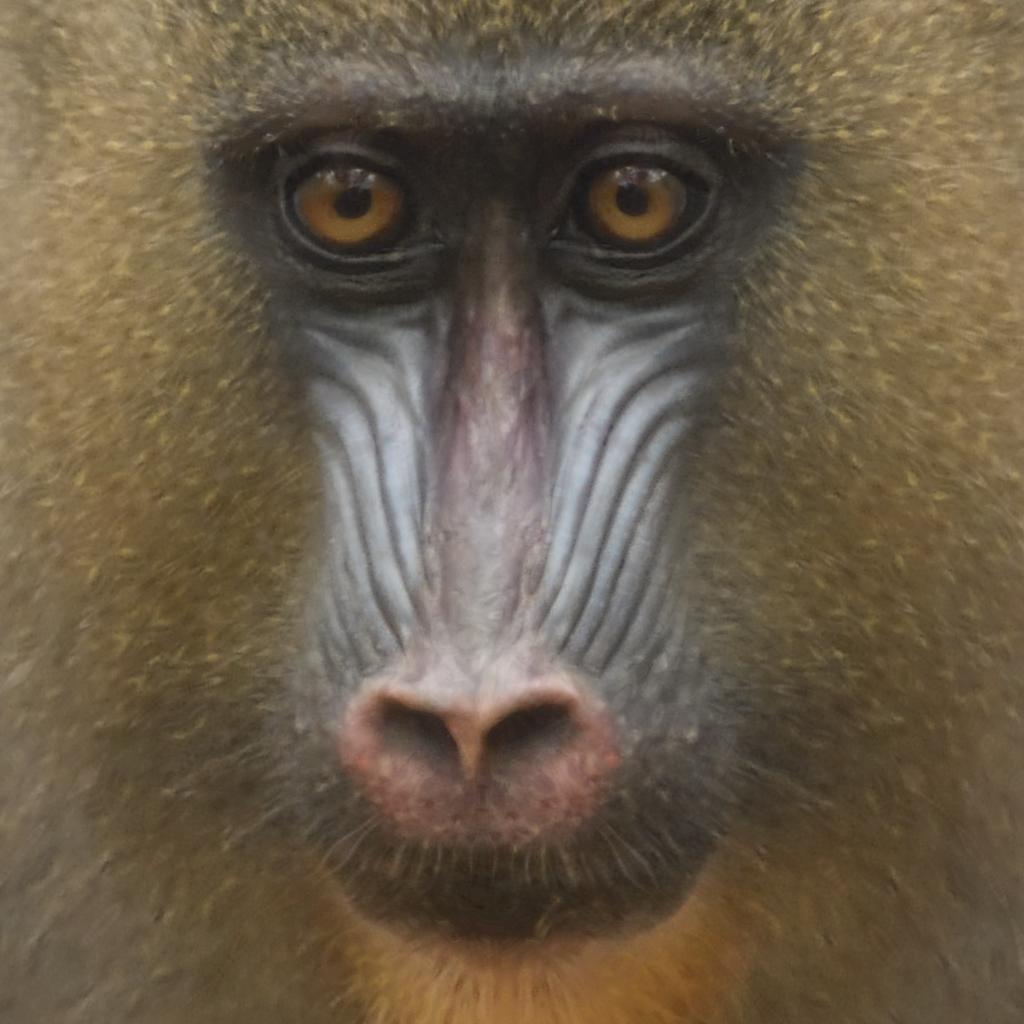}\\
    $S_r = -8.40$ & $S_r = -7.72$ & $S_r = -7.05$ & $S_r = -6.37$ \\
     $S_e = -8.33$, $i=8$ & $S_e = -7.70$, $i=6$ & $S_e = -6.93$, $i=3$ & $S_e = -6.47$, $i=2$ \\ 
     (a) &  (b) & (c) & (d) \\ 
    \multicolumn{1}{|c|}{\includegraphics[scale=0.08]{Images/221_encodage.jpg}} & 
    \includegraphics[scale=0.08]{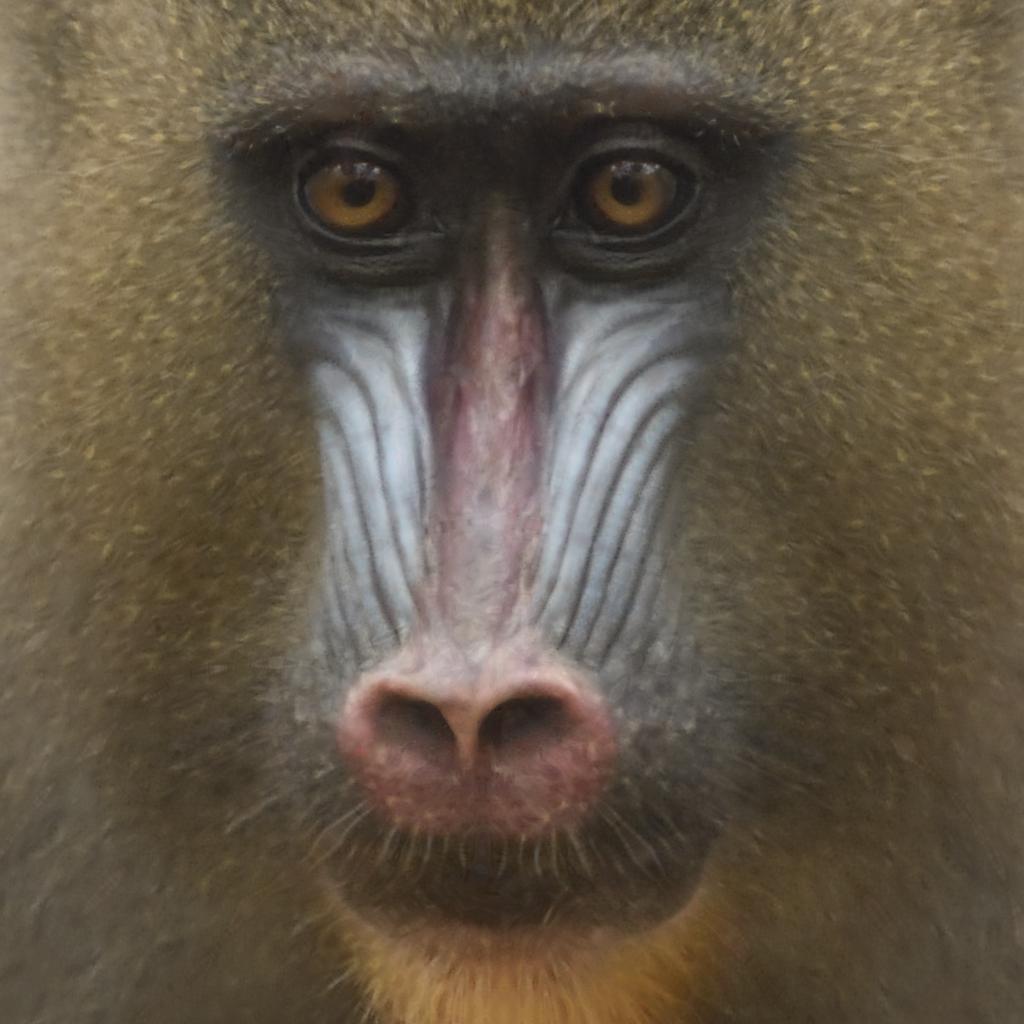} &
    \includegraphics[scale=0.08]{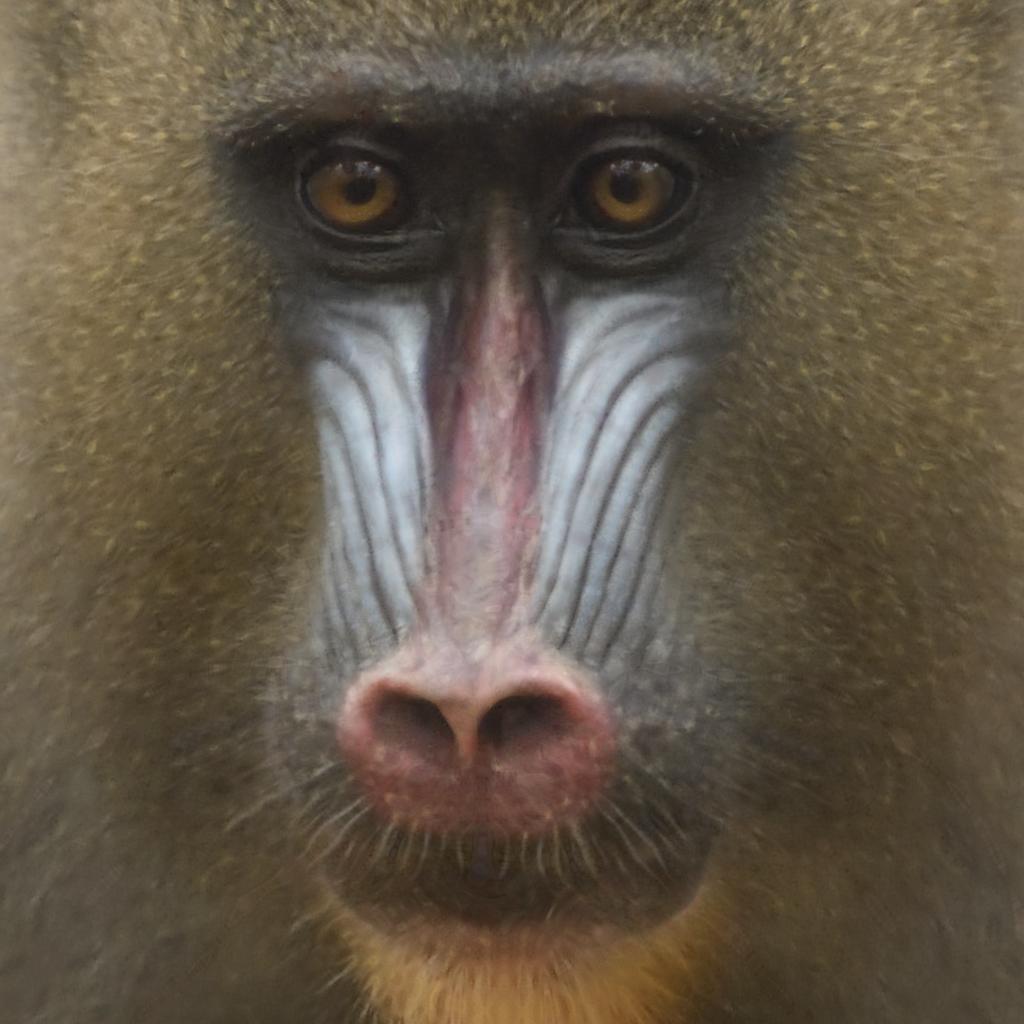} & 
    \includegraphics[scale=0.08]{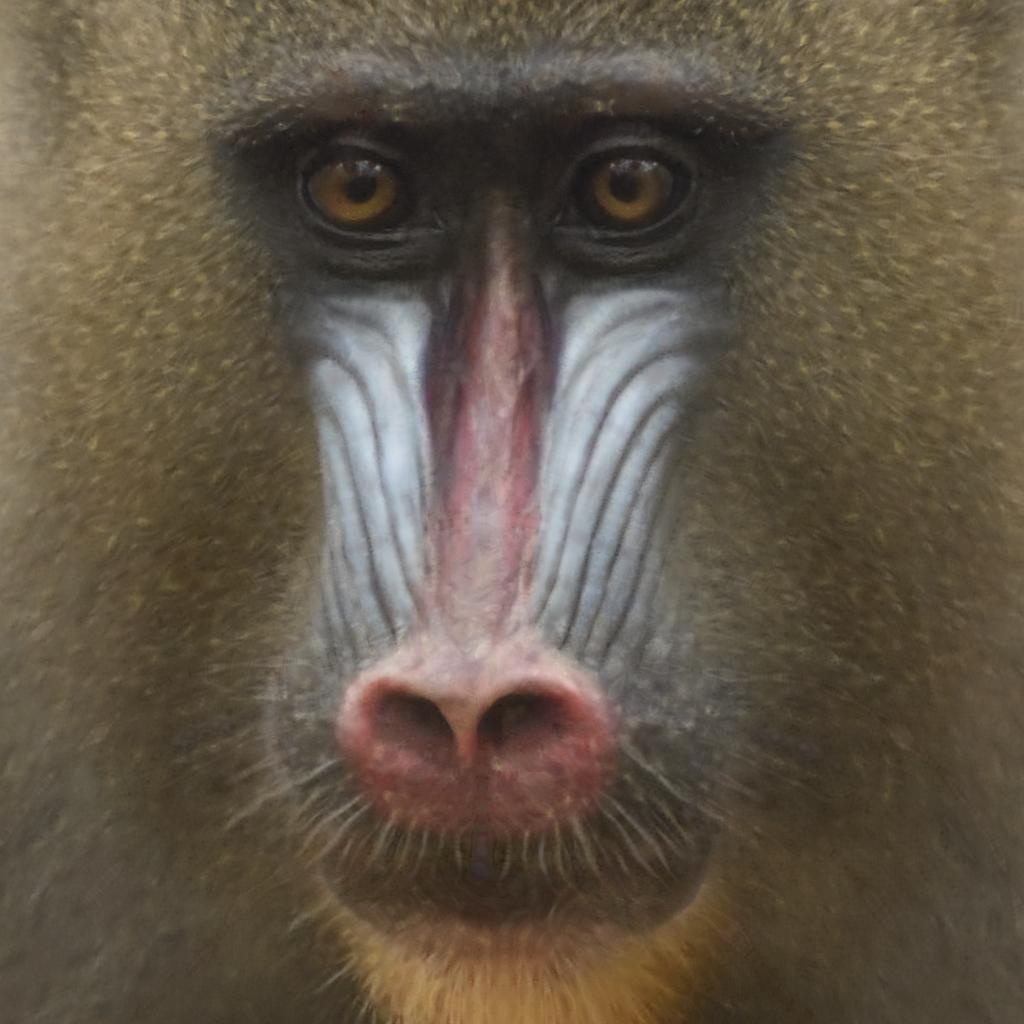}\\
    \multicolumn{1}{|c|}{$S_o = -5.70$} & $S_r = -5.02$ & $S_r = -4.34$ & $S_r = -3.67$ \\
    \multicolumn{1}{|c|}{Original} & $S_e = -5.10$, $i=5$ & $S_e = -4.25$, $i=6$ & $S_e = -3.56$, $i=4$ \\
    \multicolumn{1}{c}{(e)} & (f) & (g) & (h)
\end{tabular}    
\caption{Editing of the female mandrill face \#20210325\_id221\_femadu~\cite{TIEO2023108939} from the generated image illustrated~Fig.~\ref{fig:maleMandrillFace}.b : a) Edited generated image with a sex level $S_e = S_o -2 \sigma_M = -8.33$, b) Edited generated image with a sex level $S_e = S_o -1.5 \sigma_M = -7.72$, c) Edited generated image with a sex level $S_e = S_o - \sigma_M = -6.93$, d) Edited generated image with a sex level $S_e = S_o -0.5 \sigma_M = -6.47$,  e) Original encoded image (Fig.~\ref{fig:femaleMandrillFace}.b) with a sex level $S_o = -5.70 $, f) Edited generated image with a sex level $S_e = S_o + 0.5\sigma_M = -5.10$, g) Edited generated image with a sex level $S_e = S_o + \sigma_M = -4.25$, h) Edited generated image with a sex level $S_e = S_o + 1.5 \sigma_M = -3.56$.}
\label{fig:EditedFemaleMandrillFace8} 
\end{figure}

\begin{figure}[hbtp!]
\center
    \includegraphics[width=6.5cm]{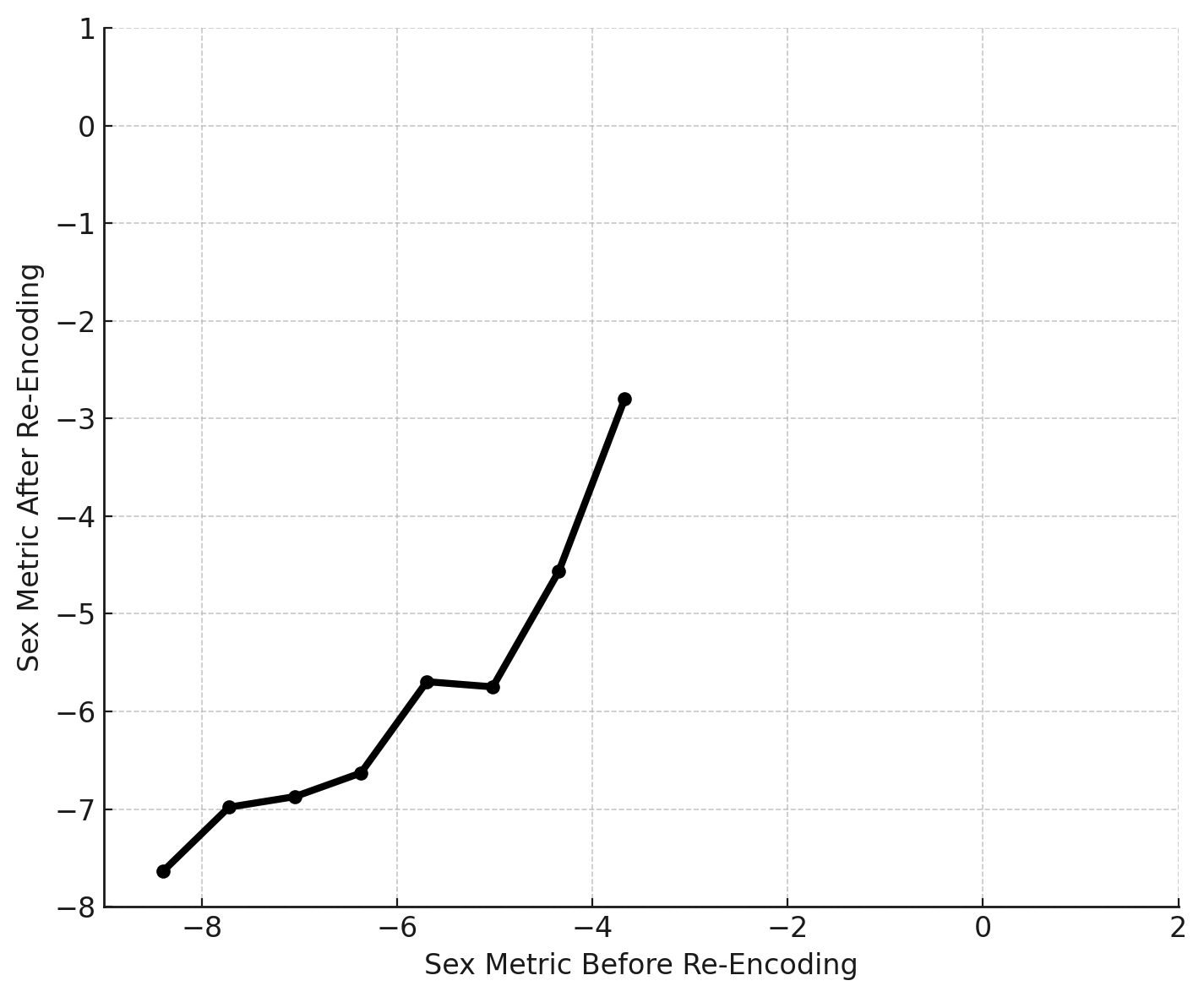} 
\caption{ Sex level editing as a function of the obtained sex levels after decoding and re-encoding for the image \#20210325\_id221\_femadu shown Fig.~\ref{fig:EditedFemaleMandrillFace8}.}
\label{fig:femaleMandrillcurve} 
\end{figure}

As shown in Fig.~\ref{fig:EditedFemaleMandrillFace8}, the generated female mandrill face Fig.~\ref{fig:femaleMandrillFace}.b is also edited with Algorithm~\ref{algo:alg2} according to the corresponding step range for a female. In order not to exceed $2 \sigma$ around the mean sex level $\mu_F$, the editing is within a range of -2 to +1.5$\sigma$. As with the male mandrill faces, overall, there seems to be a monotonic correlation between this editing and the value corresponding to what we obtain after optimization, as can be seen in Fig.~\ref{fig:femaleMandrillcurve} with several editing values. (with a small exception between 0 and 0.5$\sigma$). Visually, a more feminine face is rounder in shape, with lighter colors and proportionaly larger eyes.

%%%%%%%%%%%%%%%%%%%%%%%%%%%%%%%%%%%%%%%%%%%%%%%%%%%%%%%
\subsection{Additional examples from the MFD database subset}
\label{subsec:plus d'exemples}

\begin{figure}[hbtp!]
\center
\begin{tabular}{cccc} 
    \textbf{Real image} & \textbf{Encoded image} & \textbf{Edited image 1} & \textbf{Edited image 2} \\
    \includegraphics[scale=0.08]{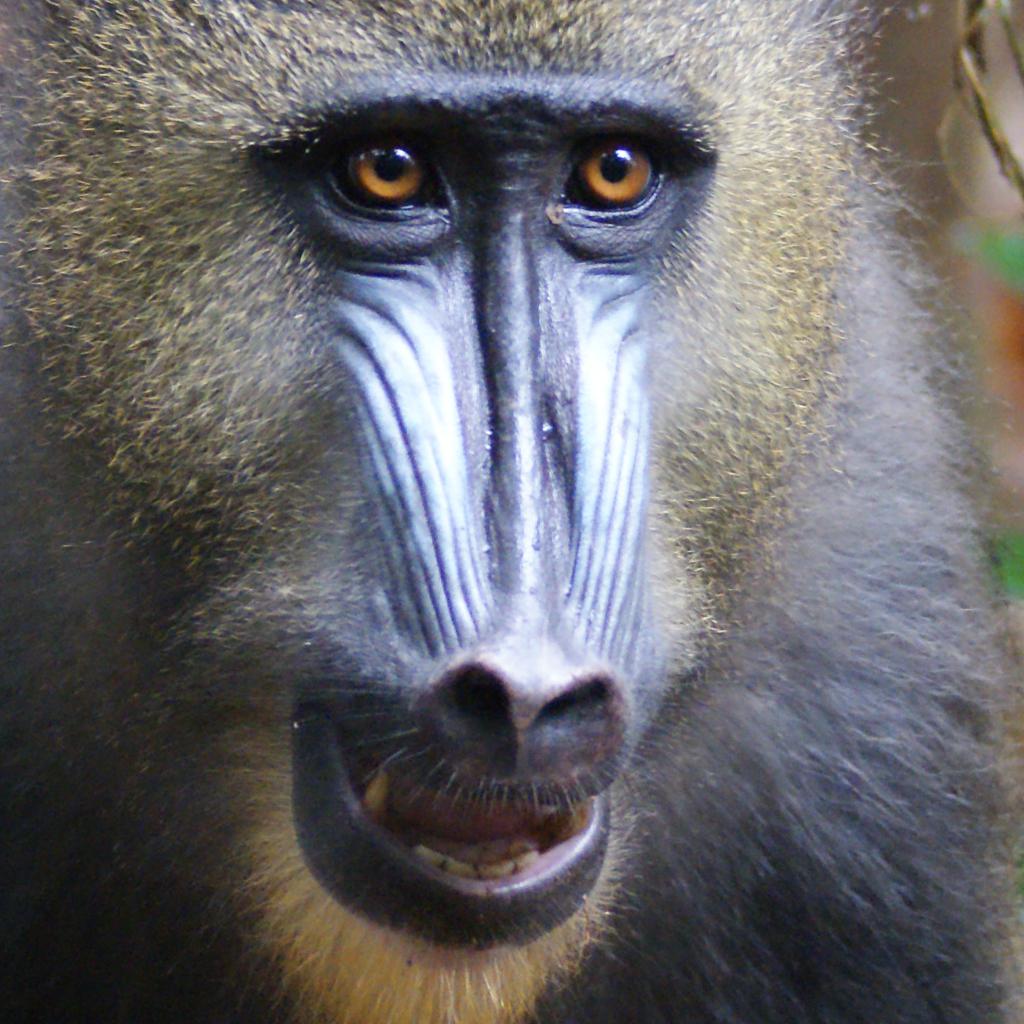} & 
    \includegraphics[scale=0.08]{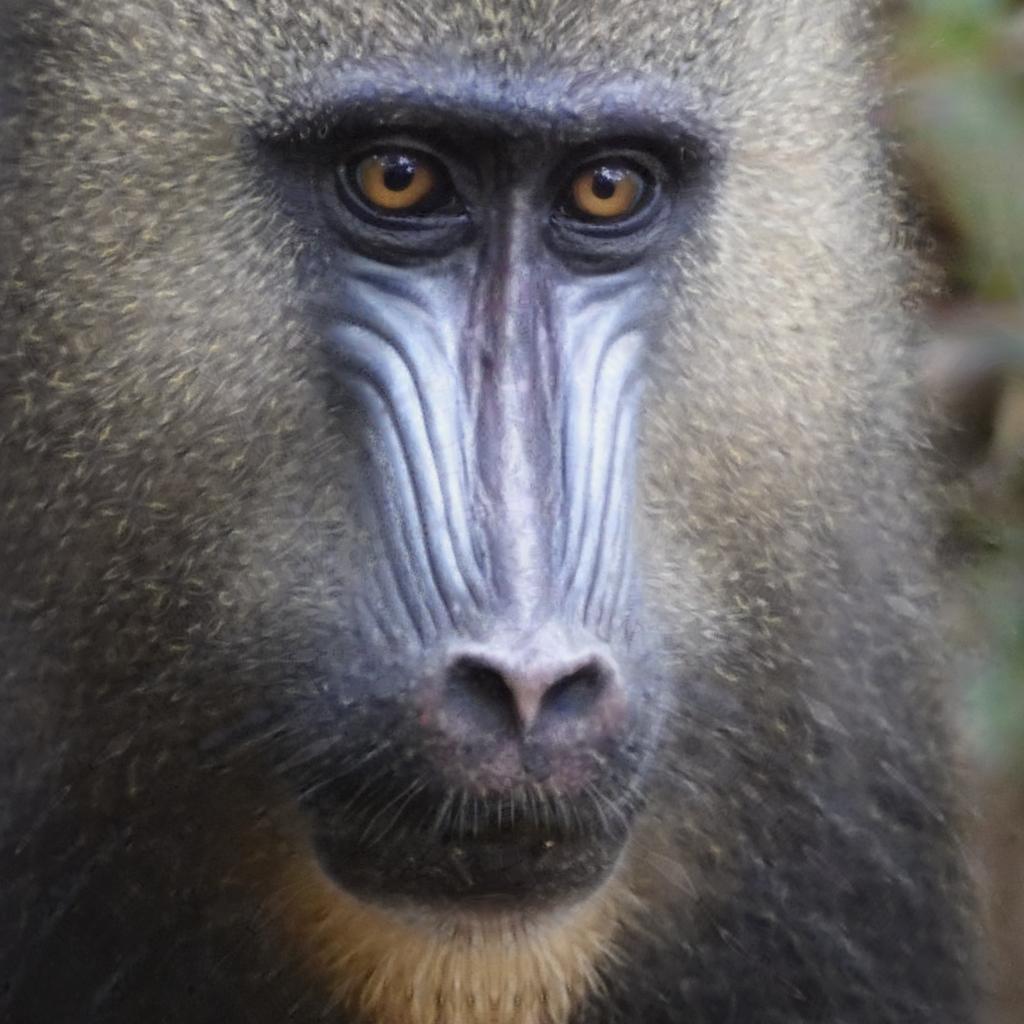} &
    \includegraphics[scale=0.08]{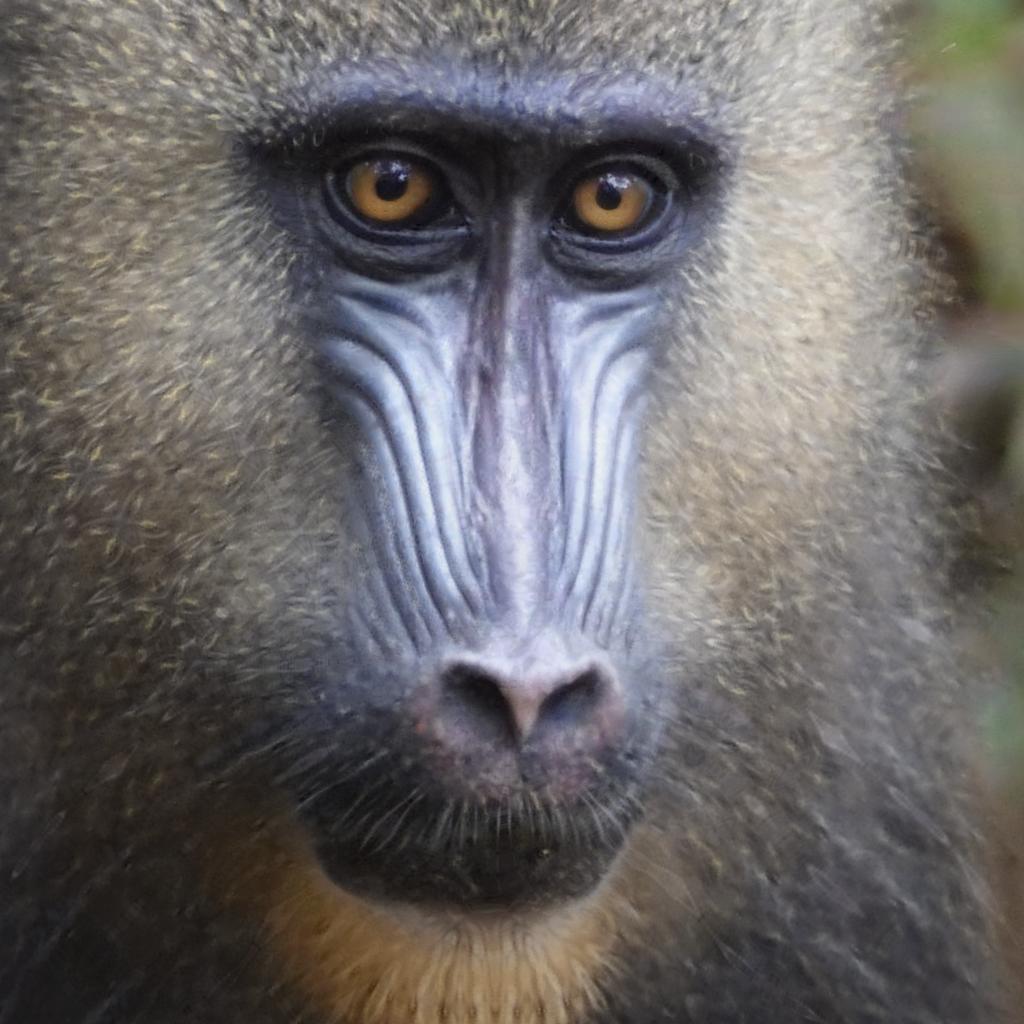} & 
    \includegraphics[scale=0.08]{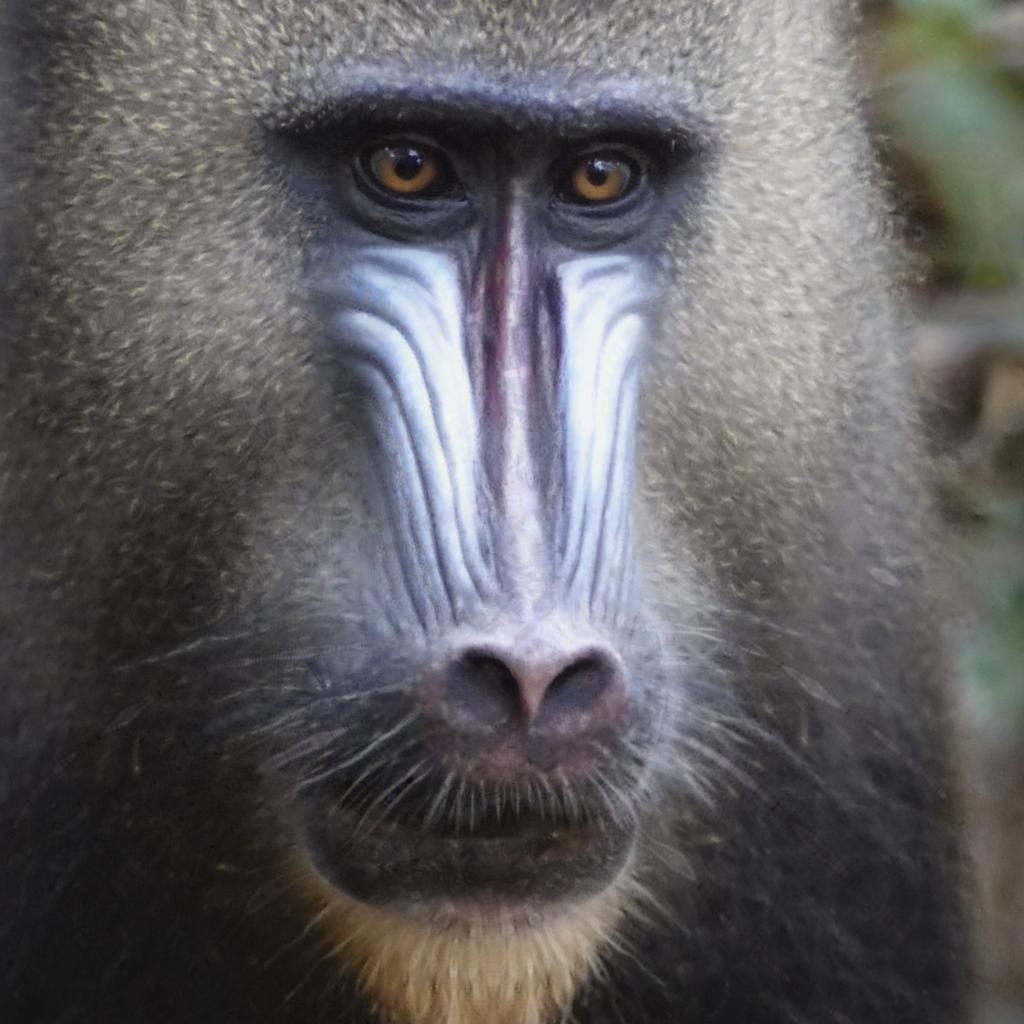}\\
    \scriptsize{\#20180409\_id103\_malado(17)} & $S_o = -3.38$ & $S_r = -4.76$ (-1$\sigma$) & $S_r = -1.99$ (+1$\sigma$) \\ 
     & & $S_e = -4.82$, $i=1$ & $S_e = -1.86$, $i=6$ \\     
    \includegraphics[scale=0.08]{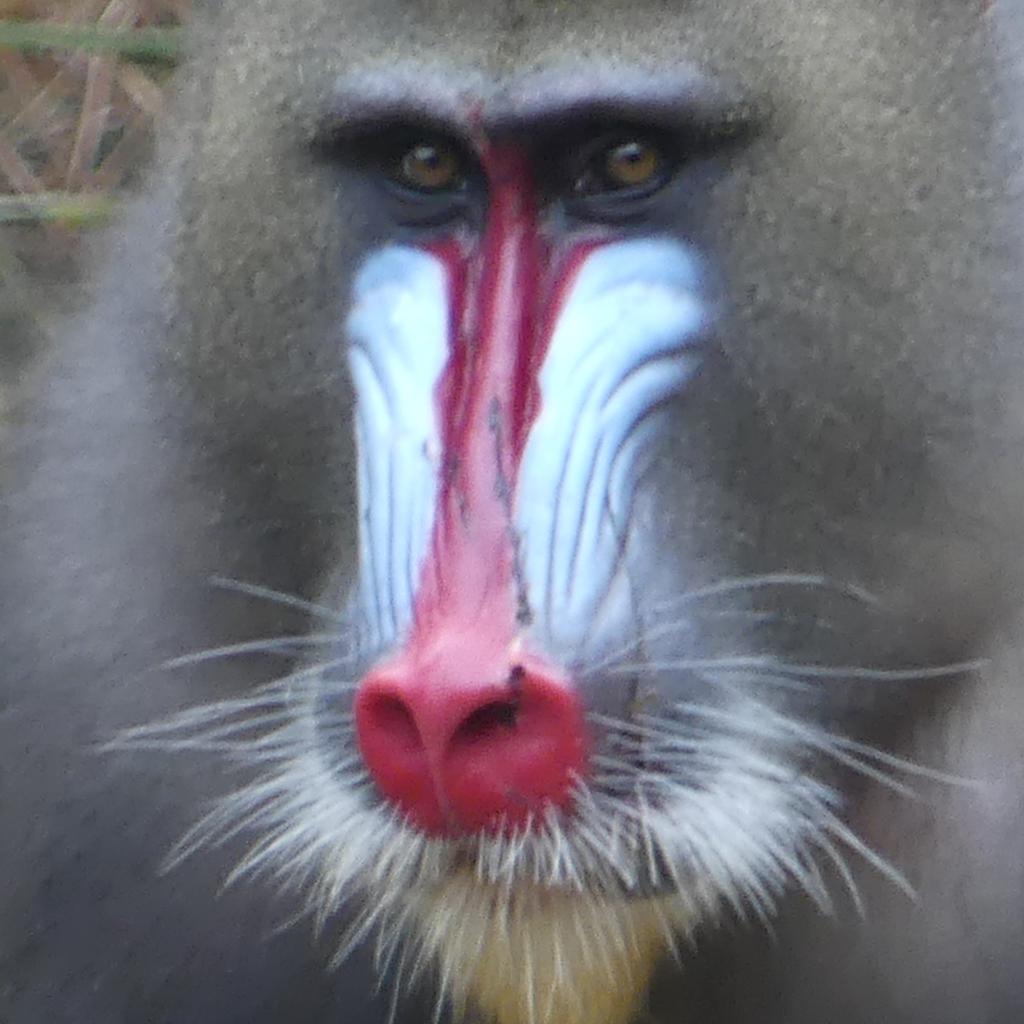} & 
    \includegraphics[scale=0.08]{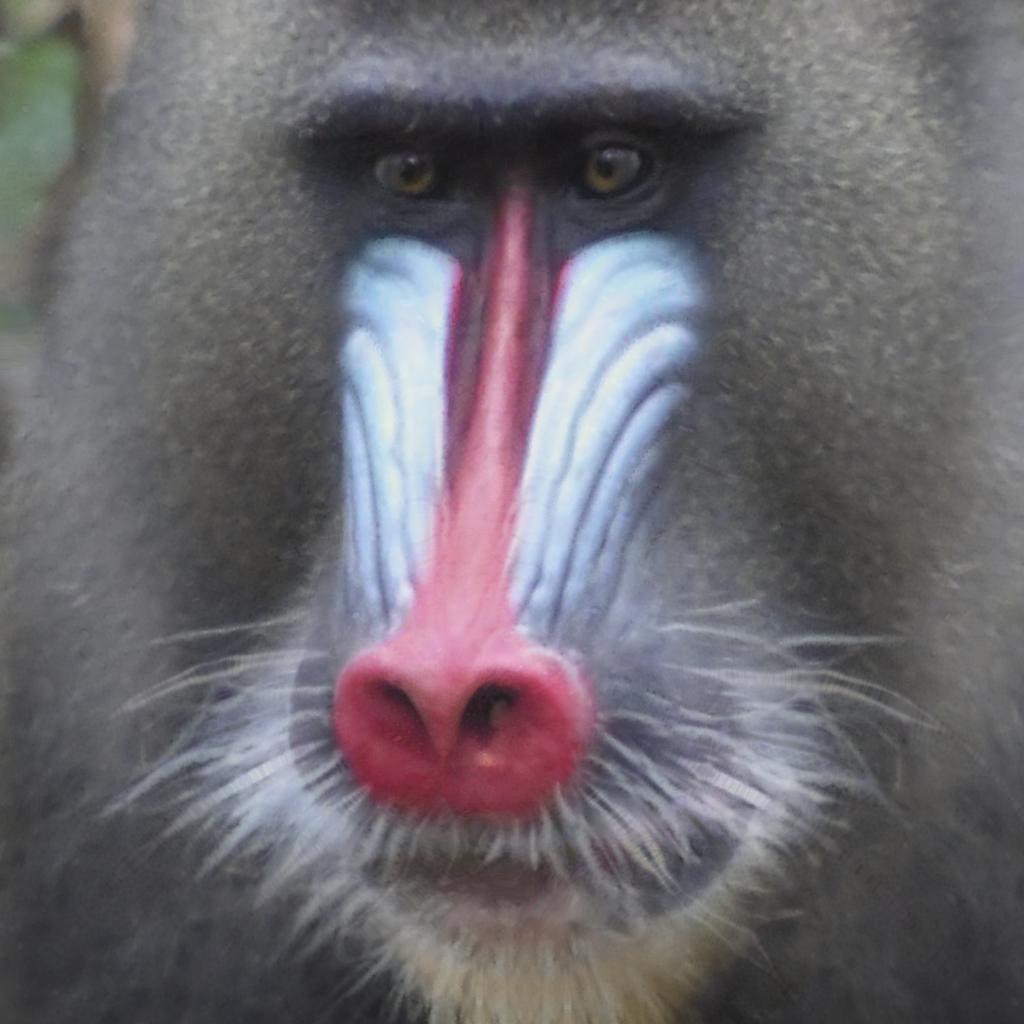} &
    \includegraphics[scale=0.08]{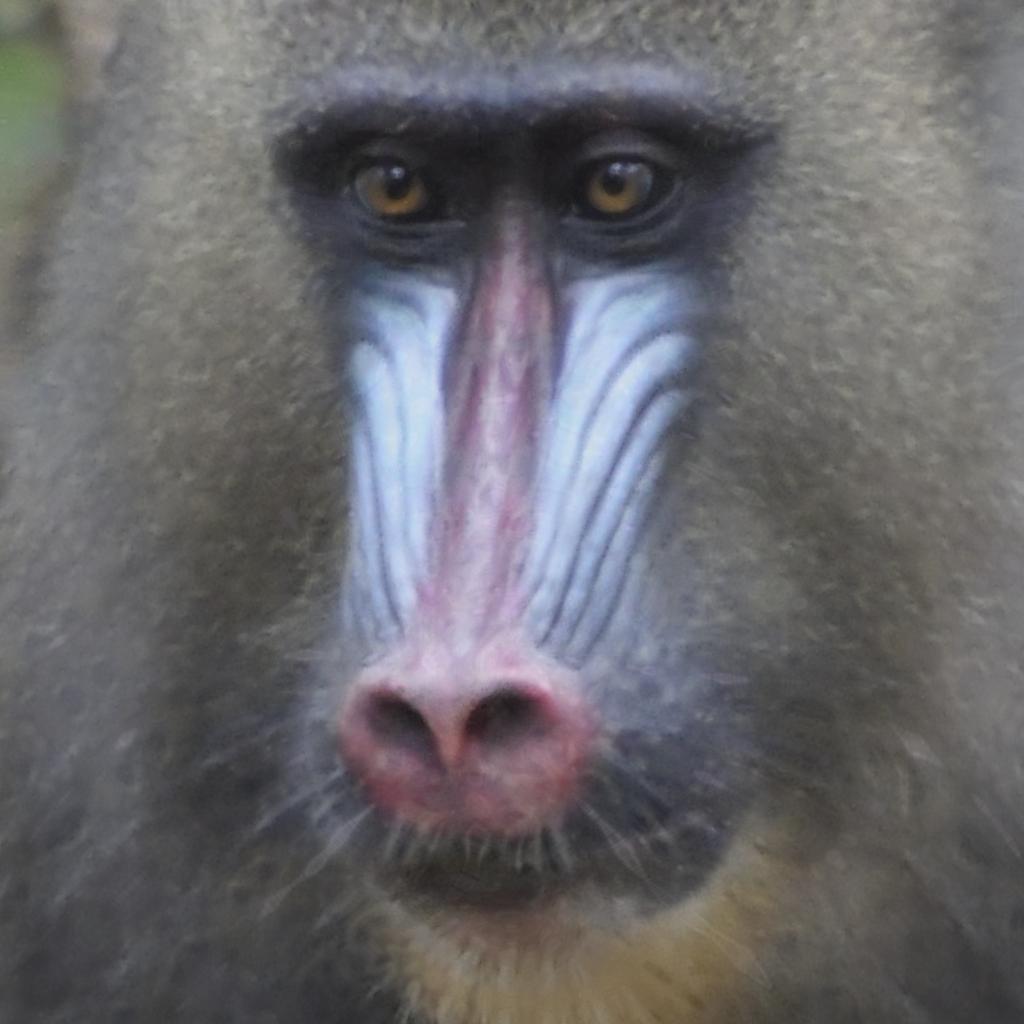} & 
    \includegraphics[scale=0.08]{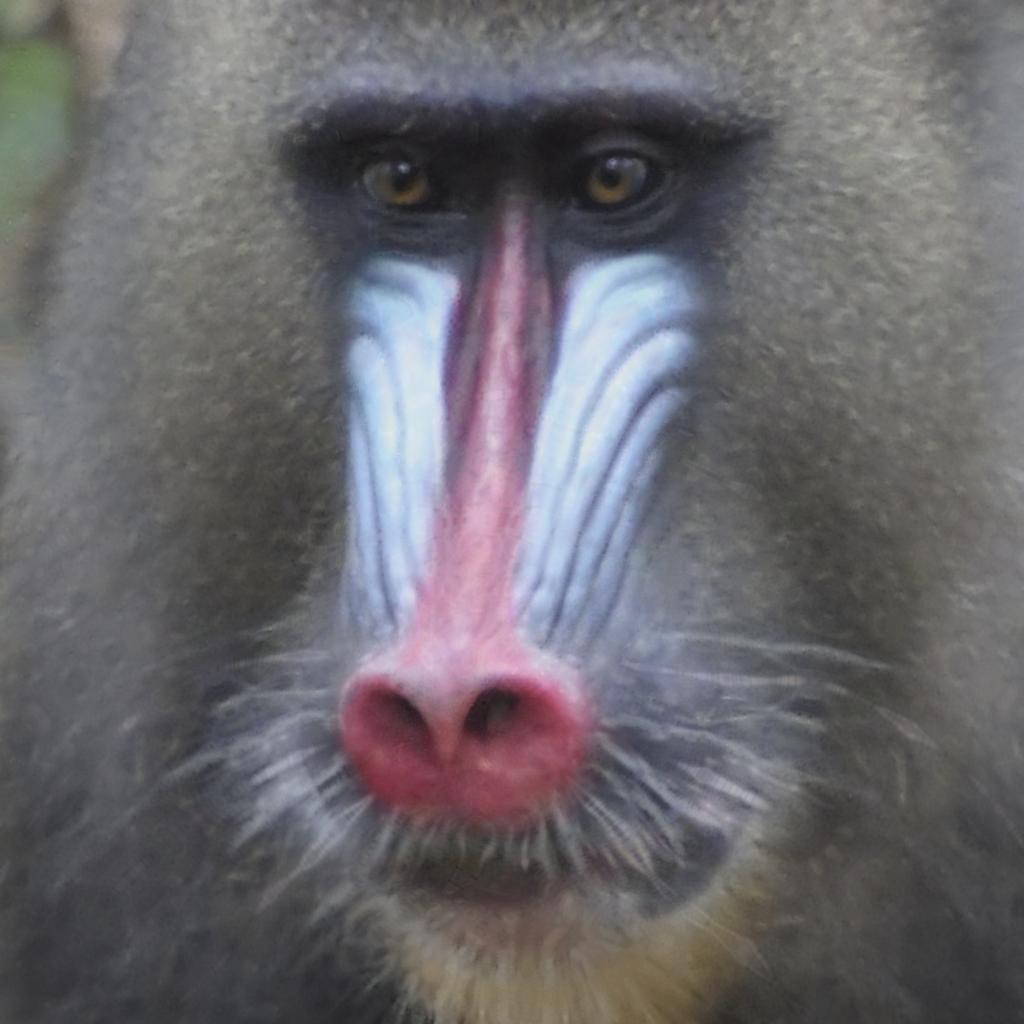}\\
    \scriptsize{\#20181008\_id152\_maladuCENTRE(12)} & $S_o = 0.92$ & $S_r = -1.85$ (-2$\sigma$) & $S_r = -0.46$ (-1$\sigma$) \\ 
     &            & $S_e = -1.75$, $i=3$ & $S_e = -0.44$, $i=2$ \\     
    \includegraphics[scale=0.08]{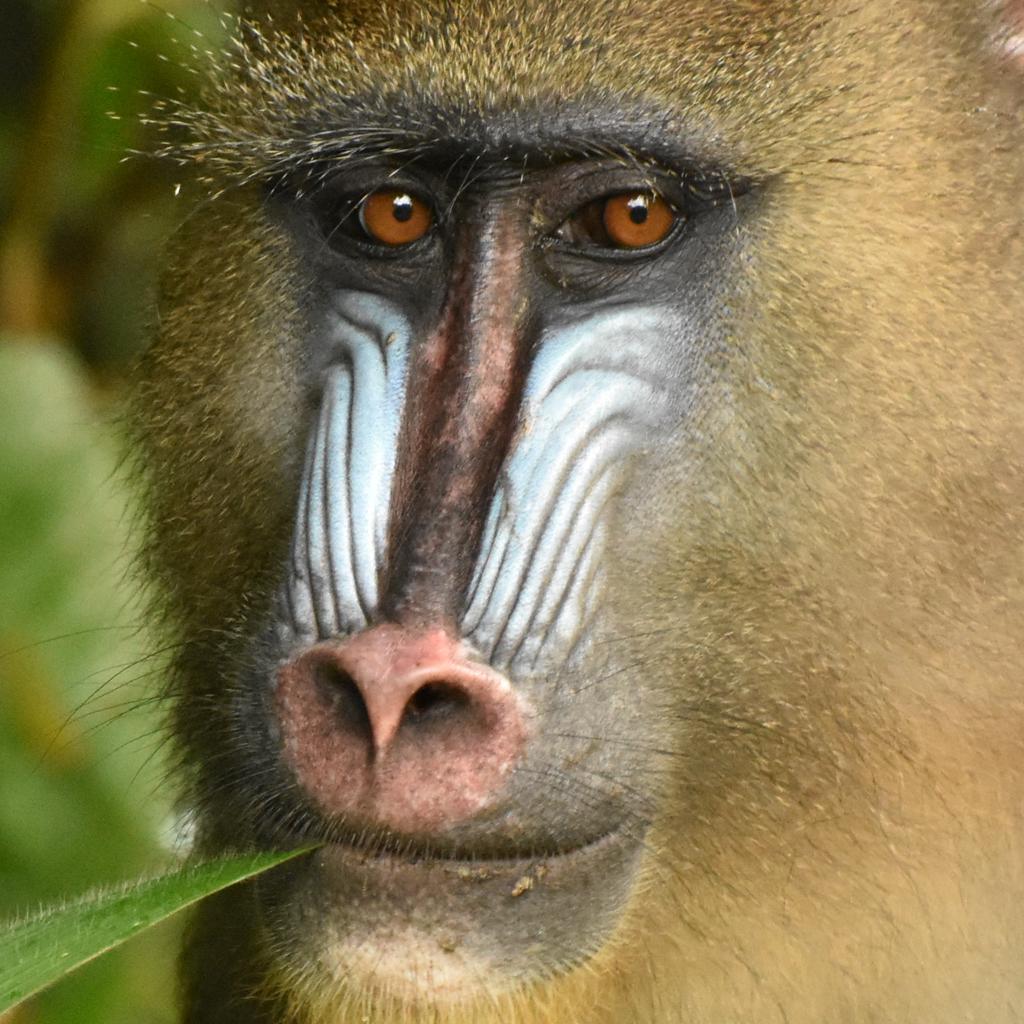} & 
    \includegraphics[scale=0.08]{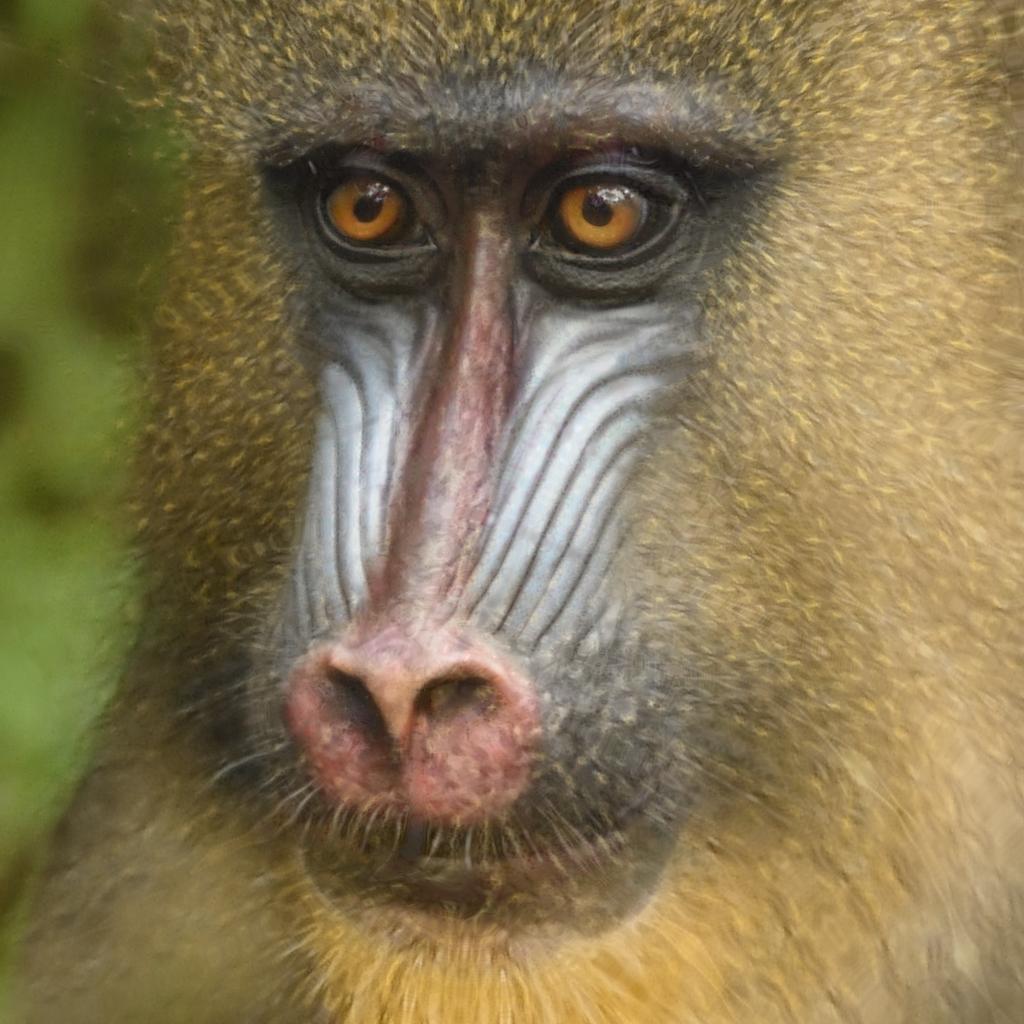} &
    \includegraphics[scale=0.08]{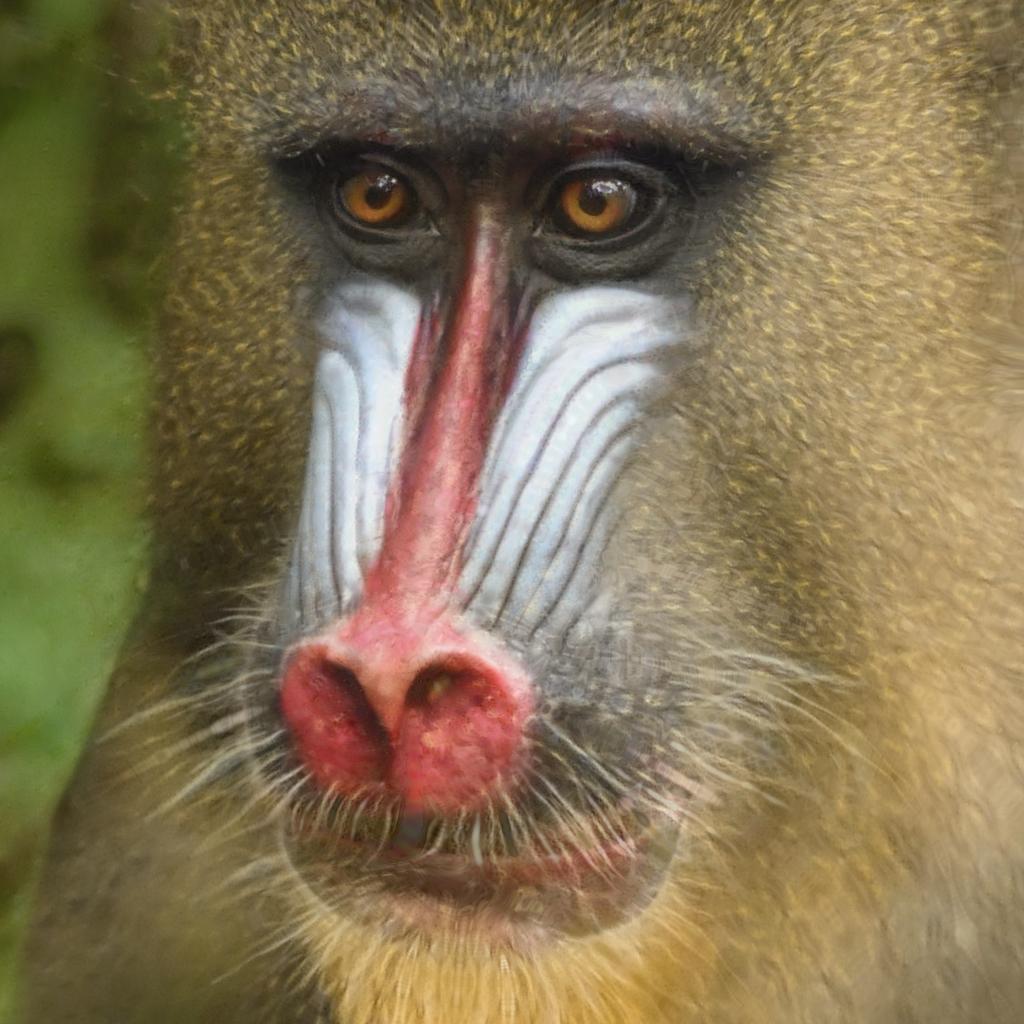} & 
    \includegraphics[scale=0.08]{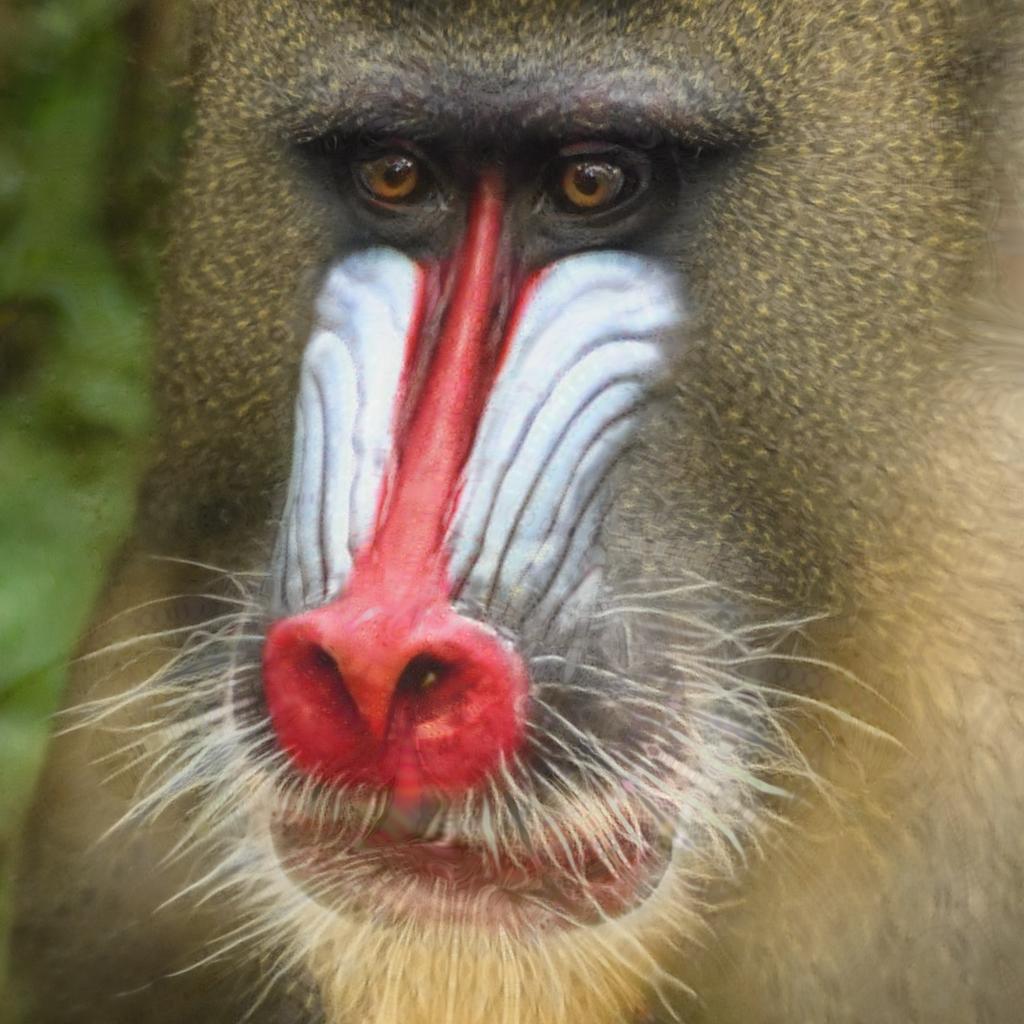}\\
    \scriptsize{\#20210130\_id62\_malsub(4)} & $S_o = -3.84$ & $S_r = -1.77$ (+1.5$\sigma$) & $S_r = 0.31$ (+3$\sigma$) \\ 
     &            & $S_e = -1.84$, $i=3$ & $S_e = 0.39$, $i=9$ \\
    \includegraphics[scale=0.08]{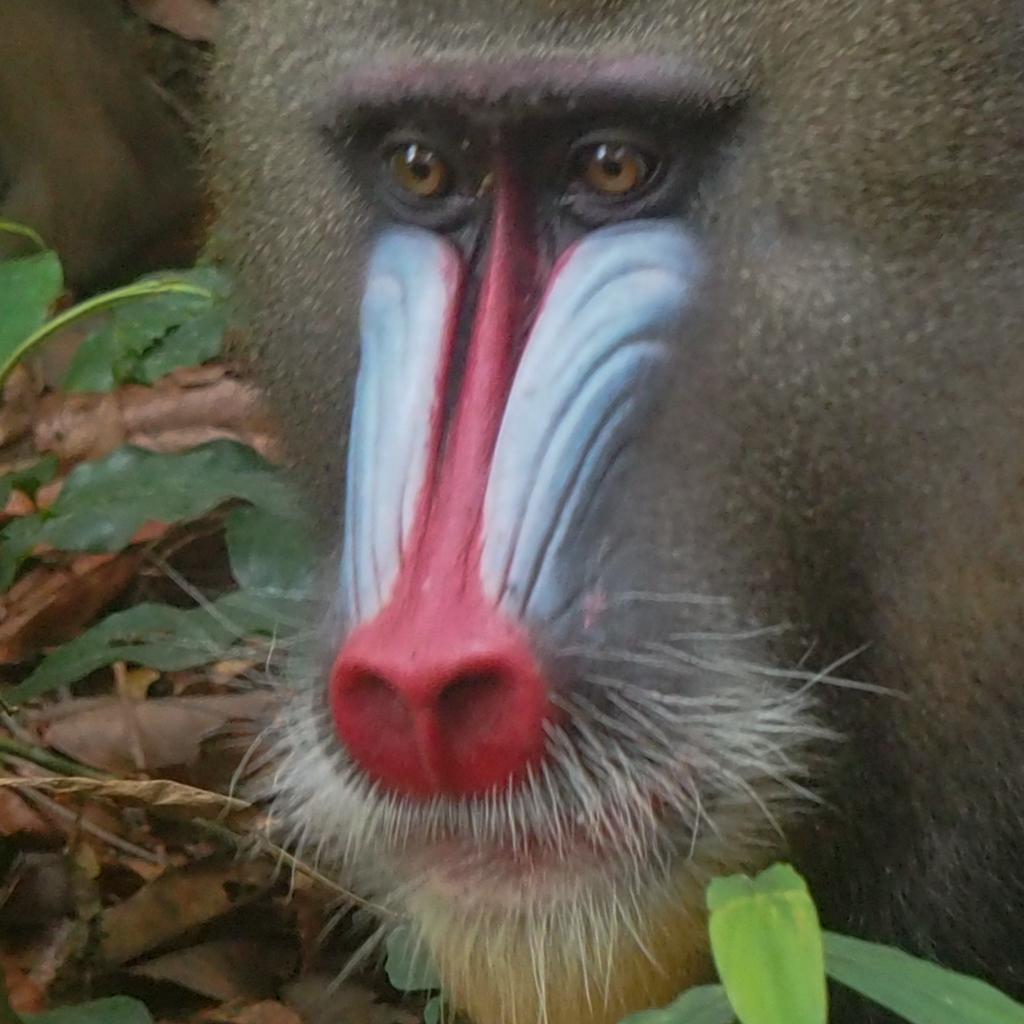} & 
    \includegraphics[scale=0.085]{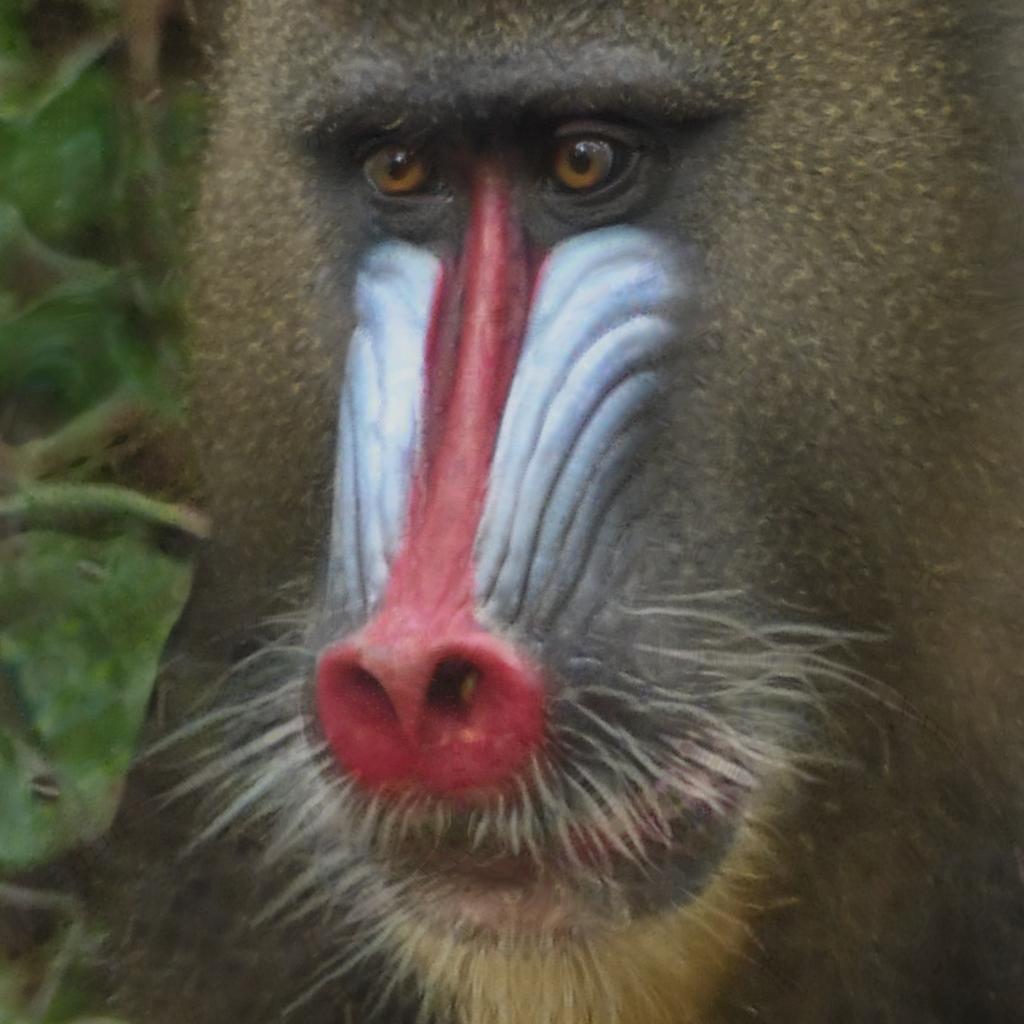} &
    \includegraphics[scale=0.085]{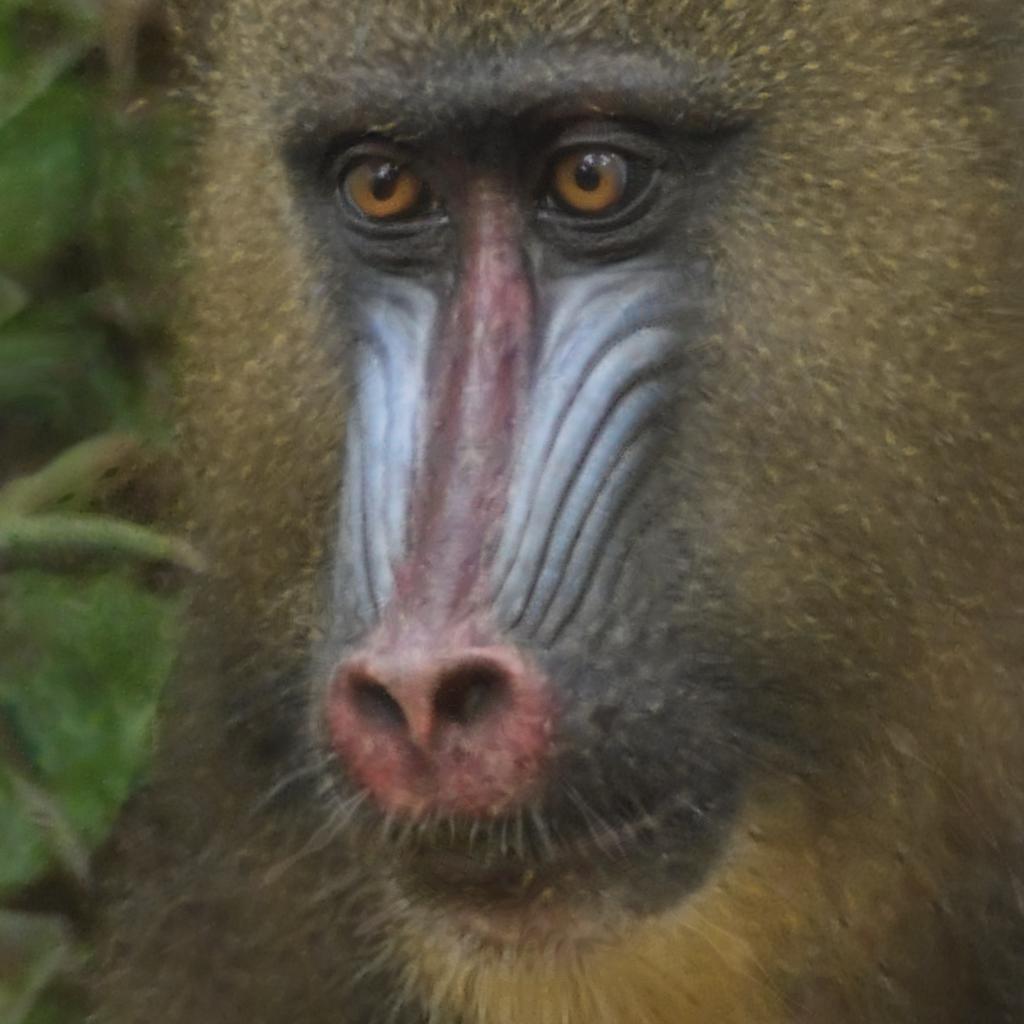} & 
    \includegraphics[scale=0.085]{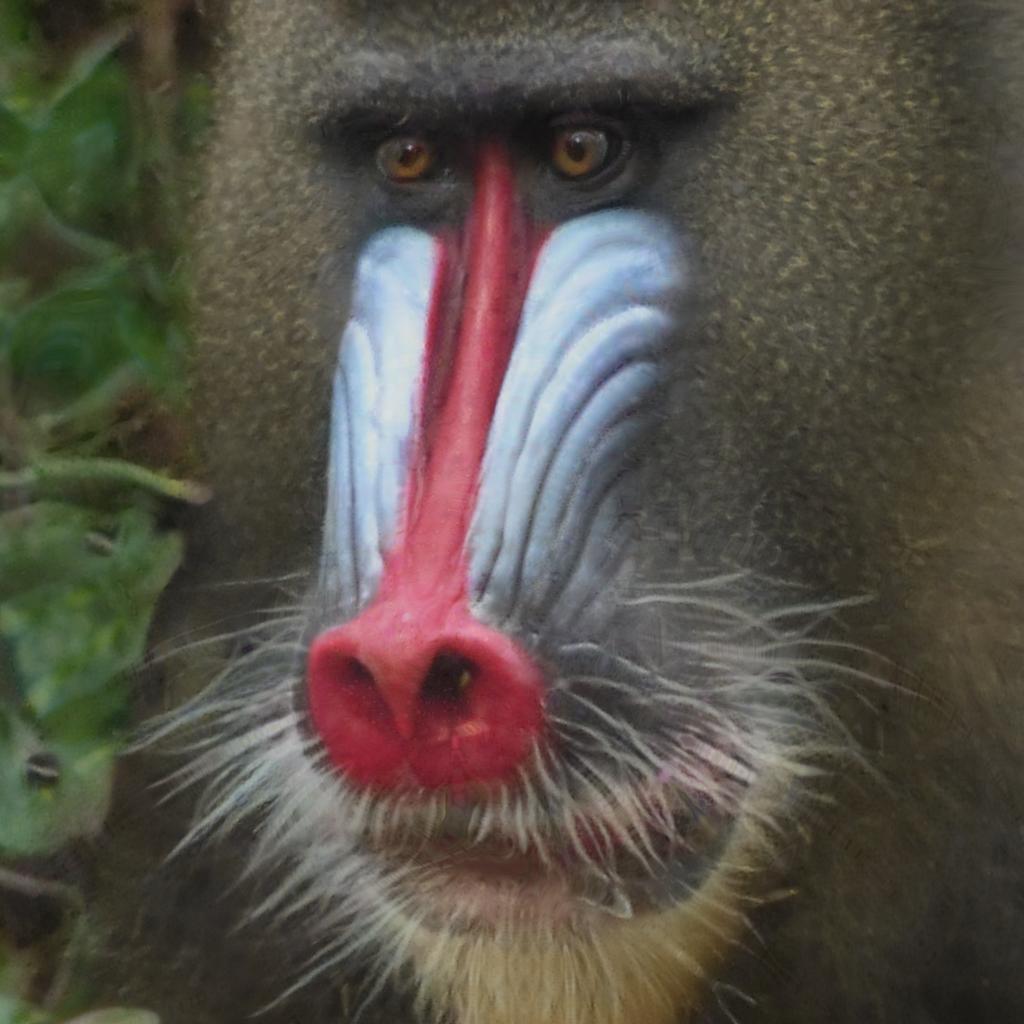}\\
    \scriptsize{\#20220717\_id167\_(2)} & $S_o = 0.21$ & $S_r = -3.25$ (-2.5$\sigma$) & $S_r = 1.59$ (+1$\sigma$) \\ 
     &            & $S_e = -3.35$, $i=8$ & $S_e = 1.51$, $i=1$ \\
\end{tabular}    
\caption{Examples of encoded and edited male mandrill face images: The first column is the original image from the MFD database; The second column is the image encoded with pSp-mandrill;  The third and fourth columns are edited images with different editing values, the first being the least masculine and the second the most masculine.}
\label{fig:maleExamples}
\end{figure}

\begin{figure}[hbtp!]
\center
\begin{tabular}{cccc} 
    \textbf{Real image} & \textbf{Encoded image} & \textbf{Edited image 1} & \textbf{Edited image 2} \\
    \includegraphics[scale=0.08]{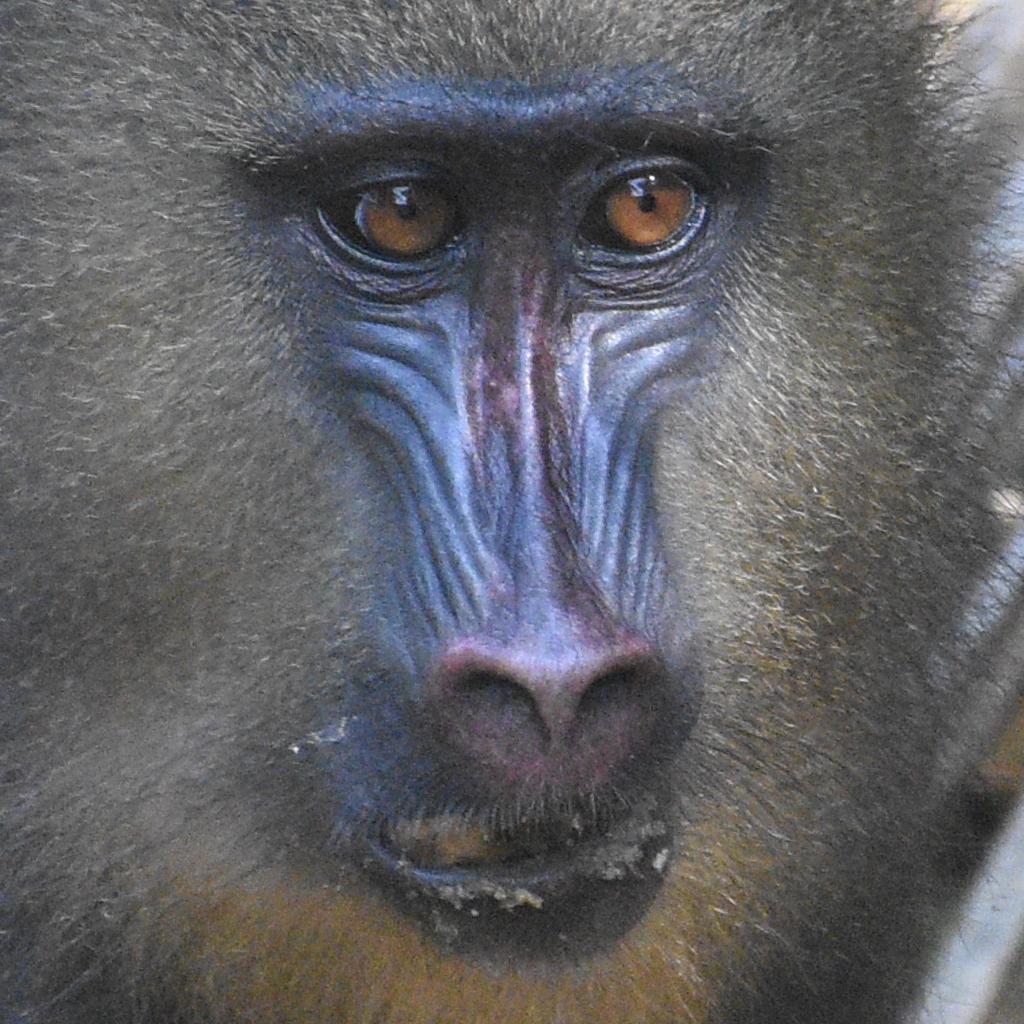} & 
    \includegraphics[scale=0.08]{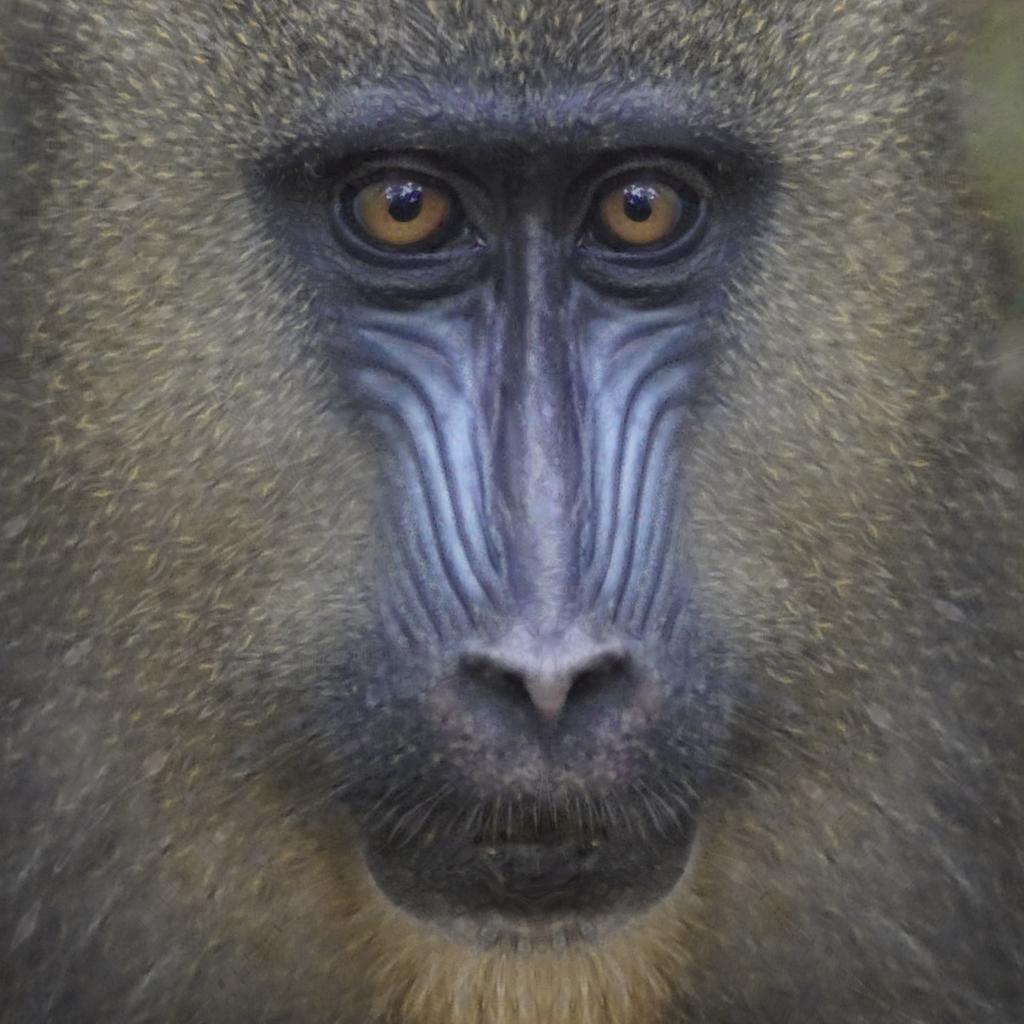} &
    \includegraphics[scale=0.08]{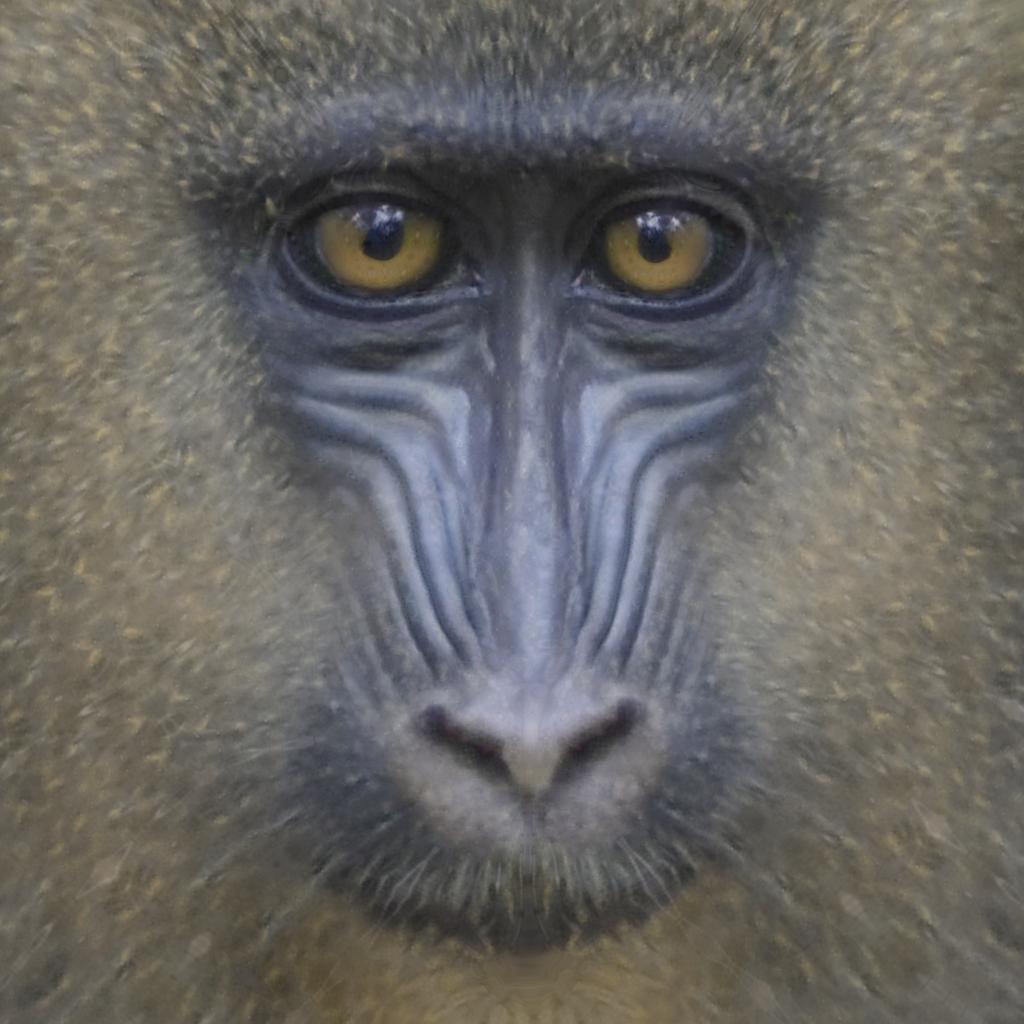} & 
    \includegraphics[scale=0.08]{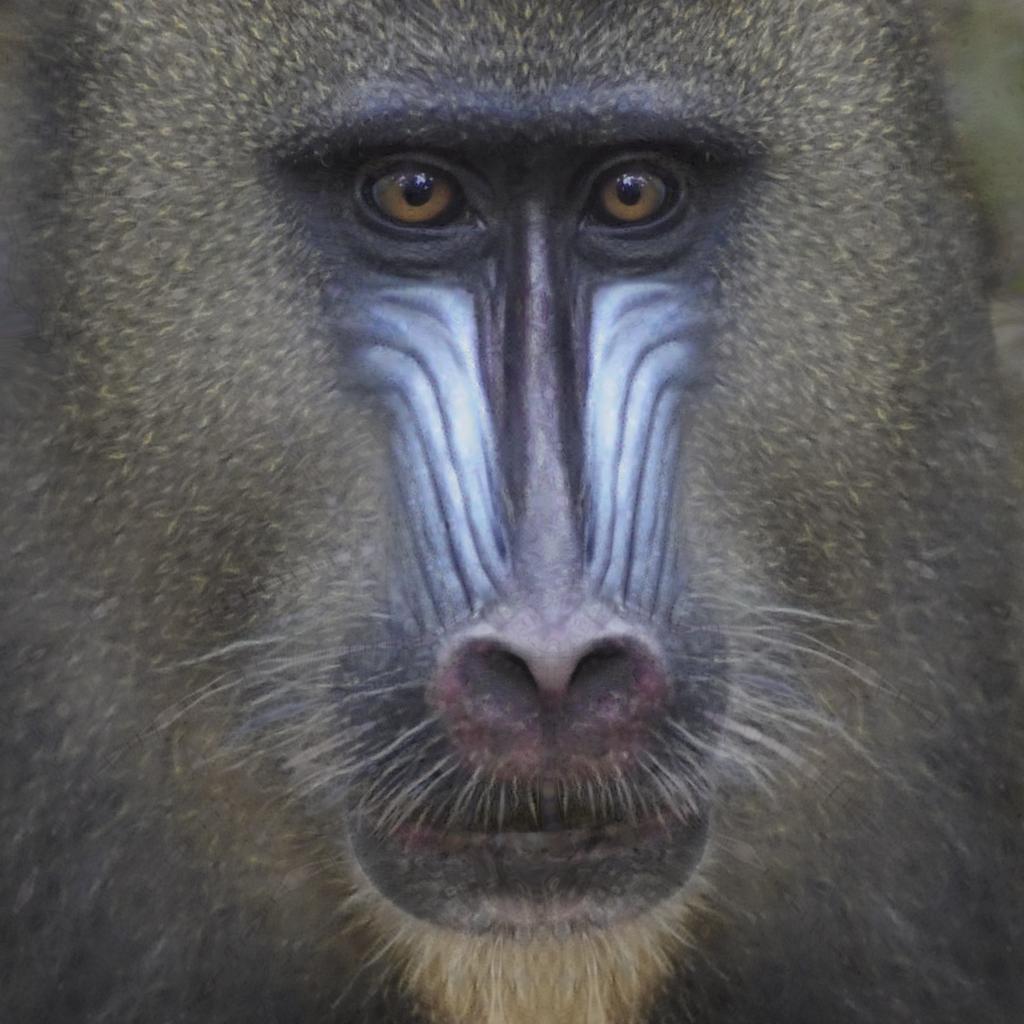}\\
    \scriptsize{\#20211219\_id174\_femadu} & $S_o = -5.02$ & $S_r = -7.72$ (-2$\sigma$) & $S_r = -2.32$ (+2$\sigma$) \\ 
     &            & $S_e = -7.66$, $i=10$ & $S_e = -2.23$, $i=4$ \\
    \includegraphics[scale=0.08]{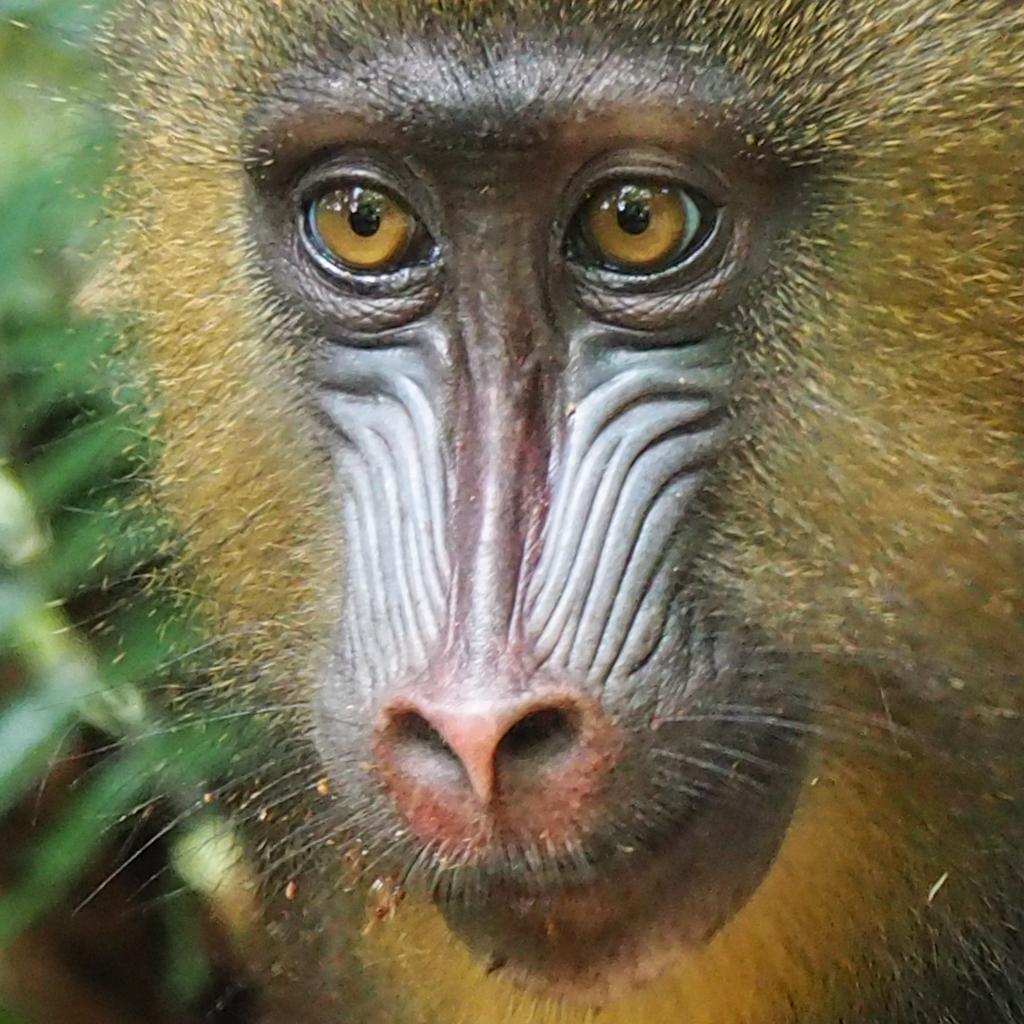} & 
    \includegraphics[scale=0.08]{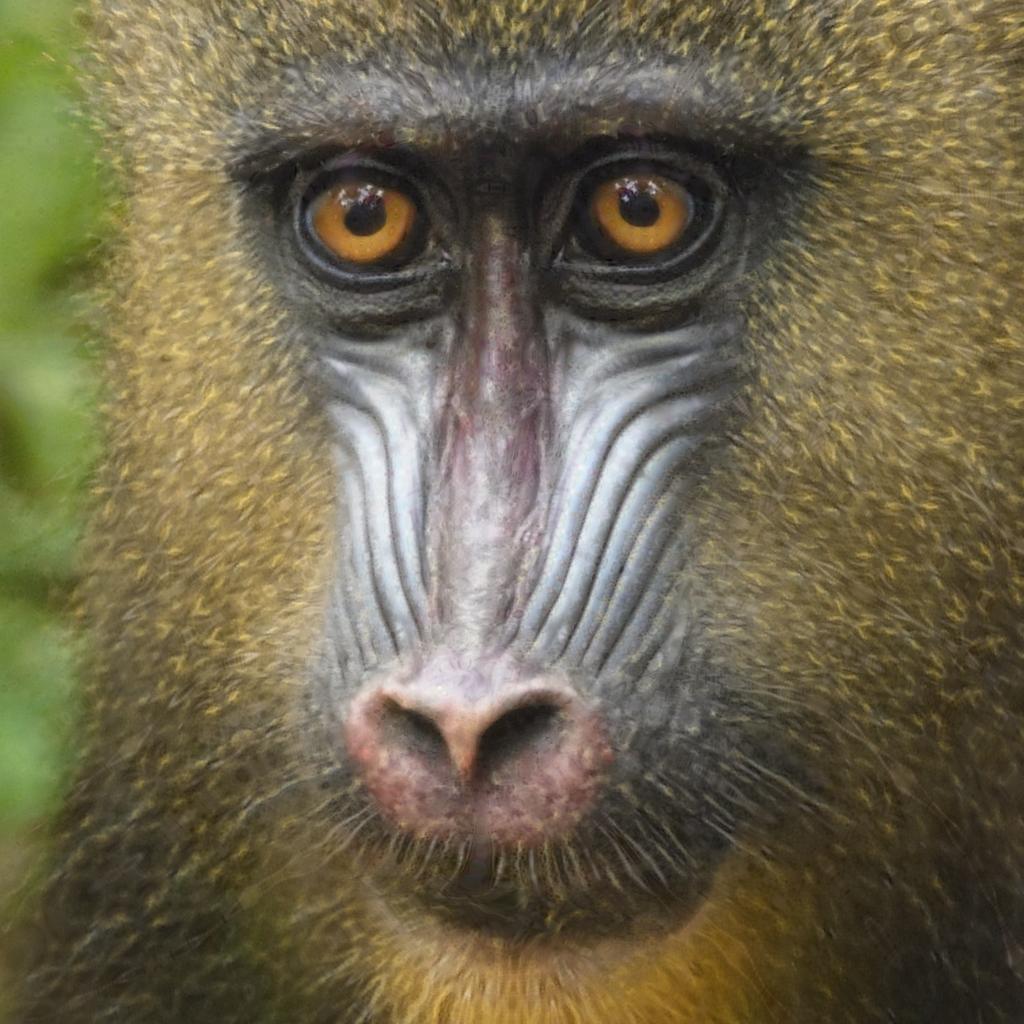} &
    \includegraphics[scale=0.08]{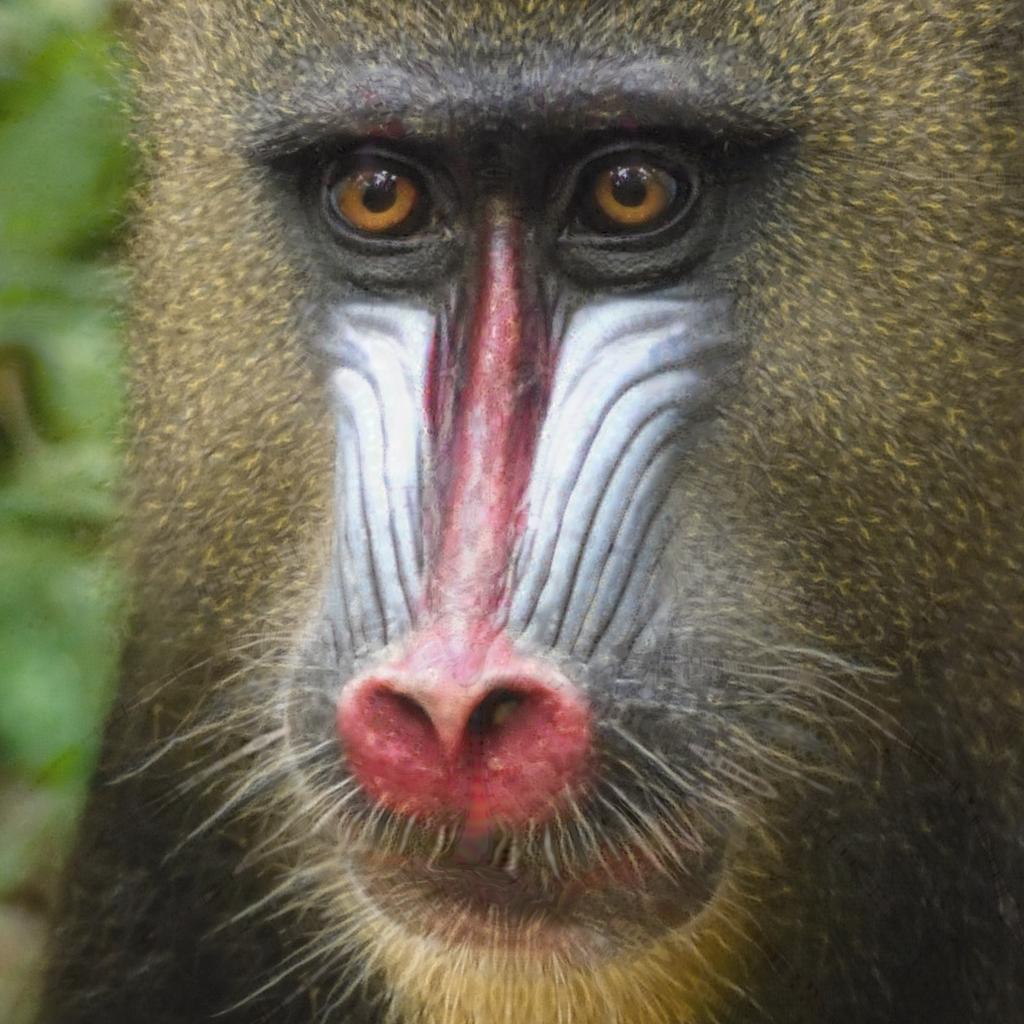} & 
    \includegraphics[scale=0.08]{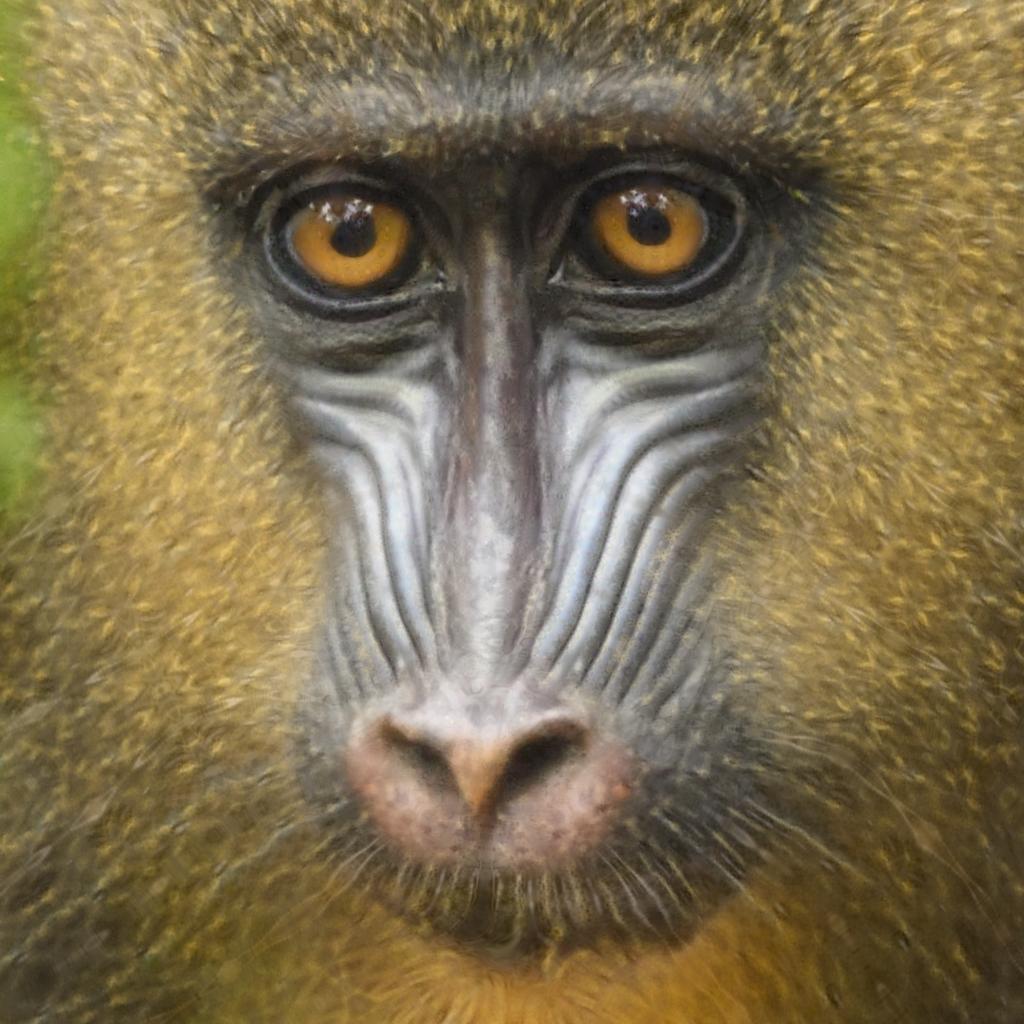}\\
    \scriptsize{\#20220426\_id228\_femadu(1)} & $S_o = -6.86$ & $S_r = -2.81$ (+3$\sigma$) & $S_r = -8.21$ (-1$\sigma$) \\ 
     &            & $S_e = -2.91$, $i=10$ & $S_e = -8.16$, $i=11$ \\
    \includegraphics[scale=0.08]{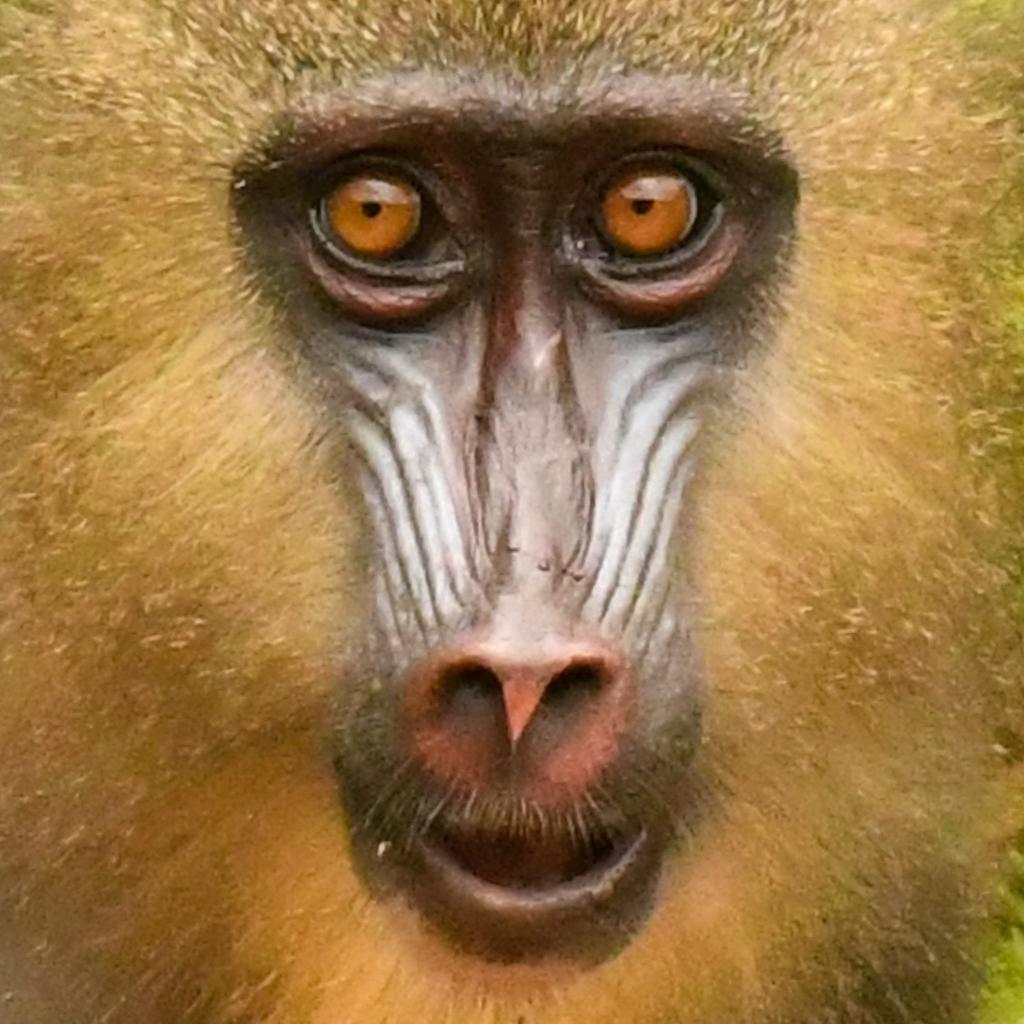} & 
    \includegraphics[scale=0.08]{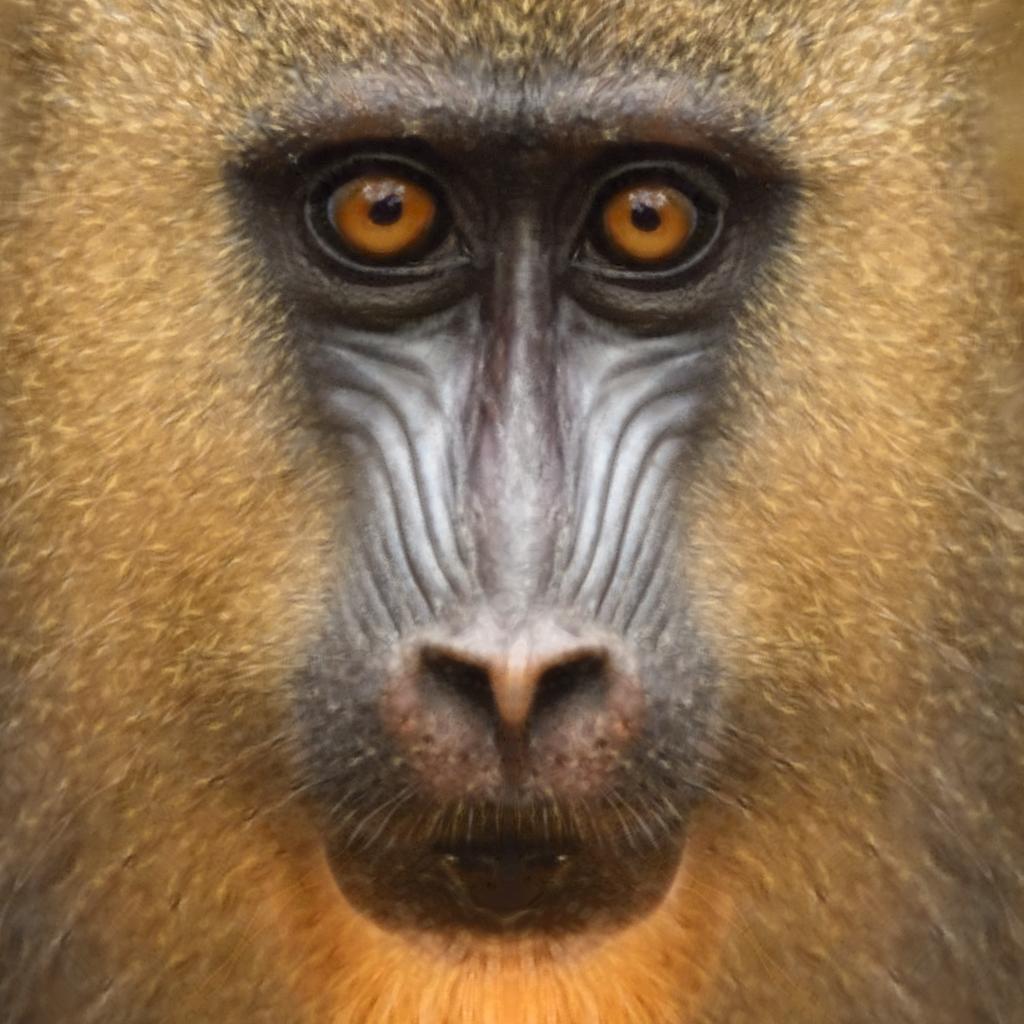} &
    \includegraphics[scale=0.08]{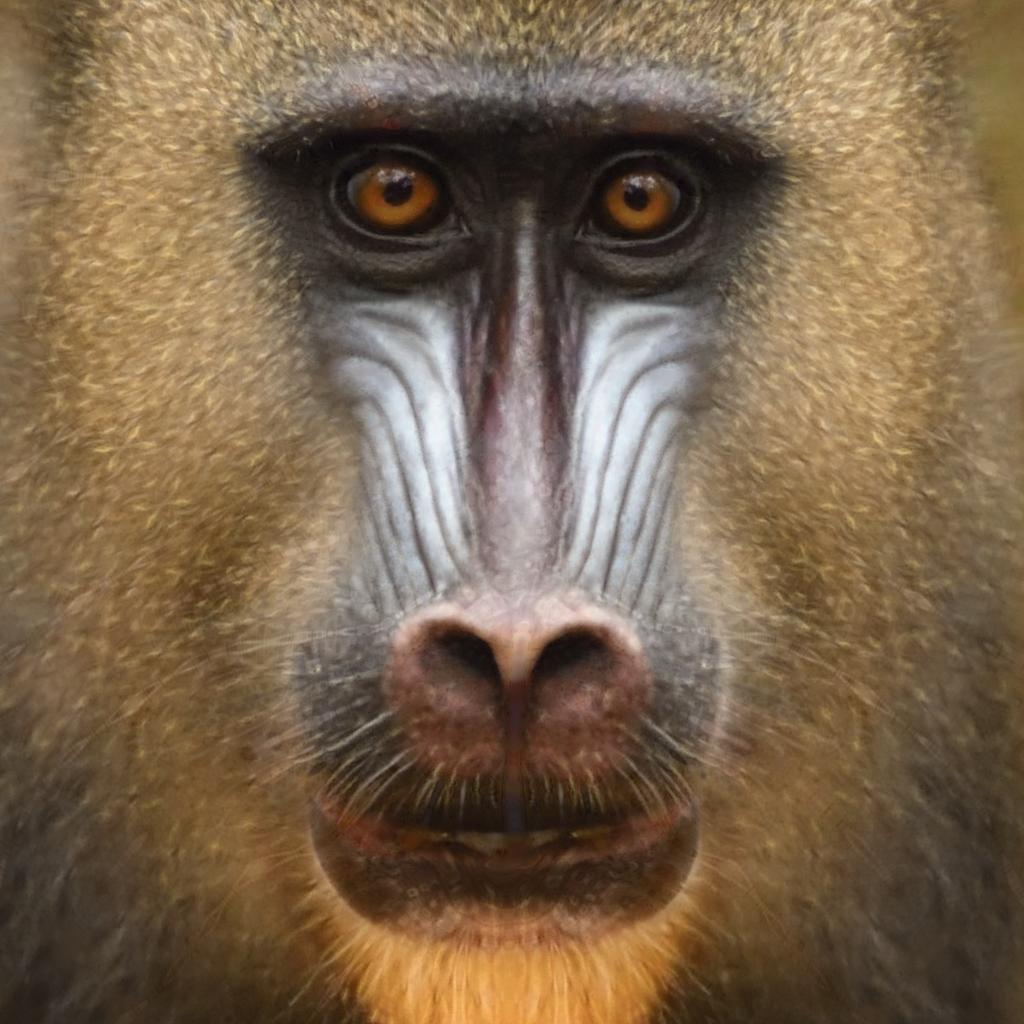} & 
    \includegraphics[scale=0.08]{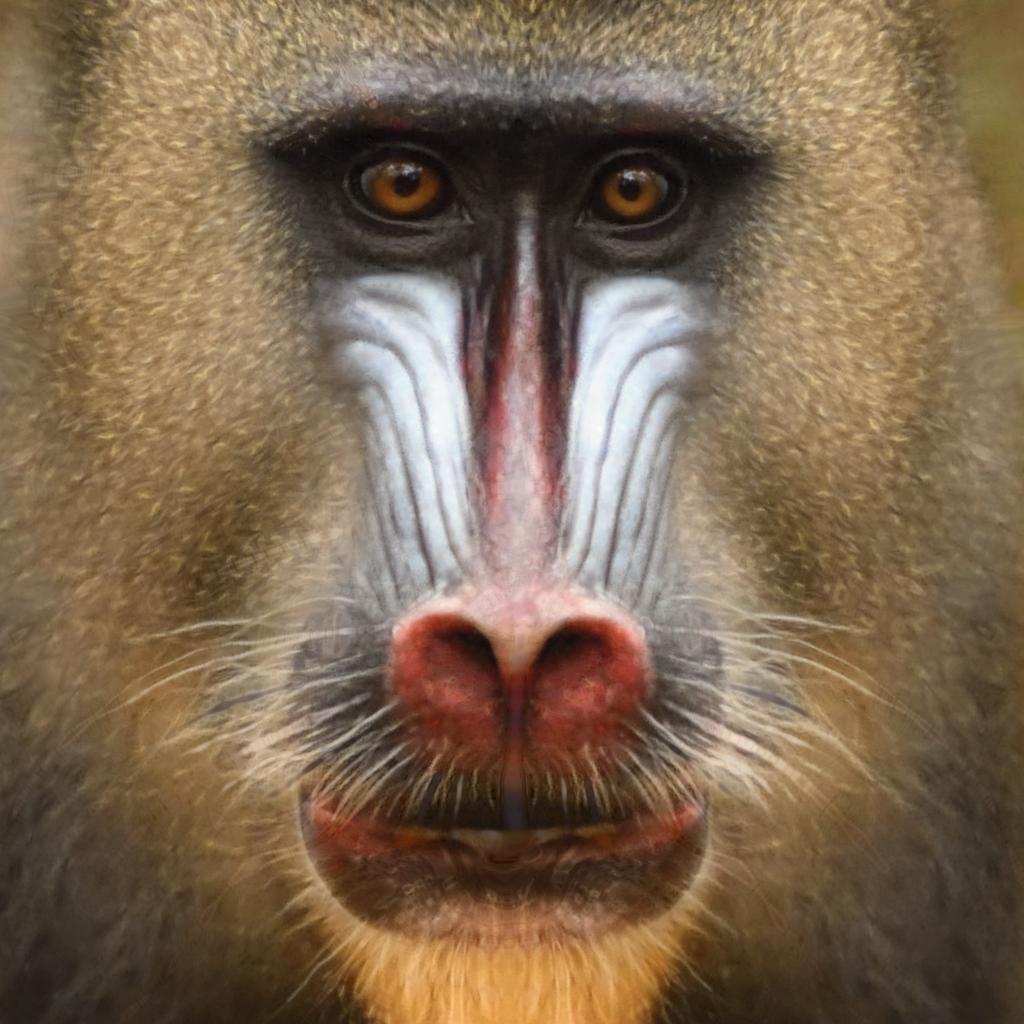}\\
    \scriptsize{\#20220701\_id199\_femadu(3)} & $S_o = -7.92$ & $S_r = -5.22$ (+2$\sigma$) & $S_r = -2.52$ (+4$\sigma$) \\ 
     &            & $S_e = -5.10$, $i=7$ & $S_e = -2.62$, $i=3$ \\
    \includegraphics[scale=0.08]{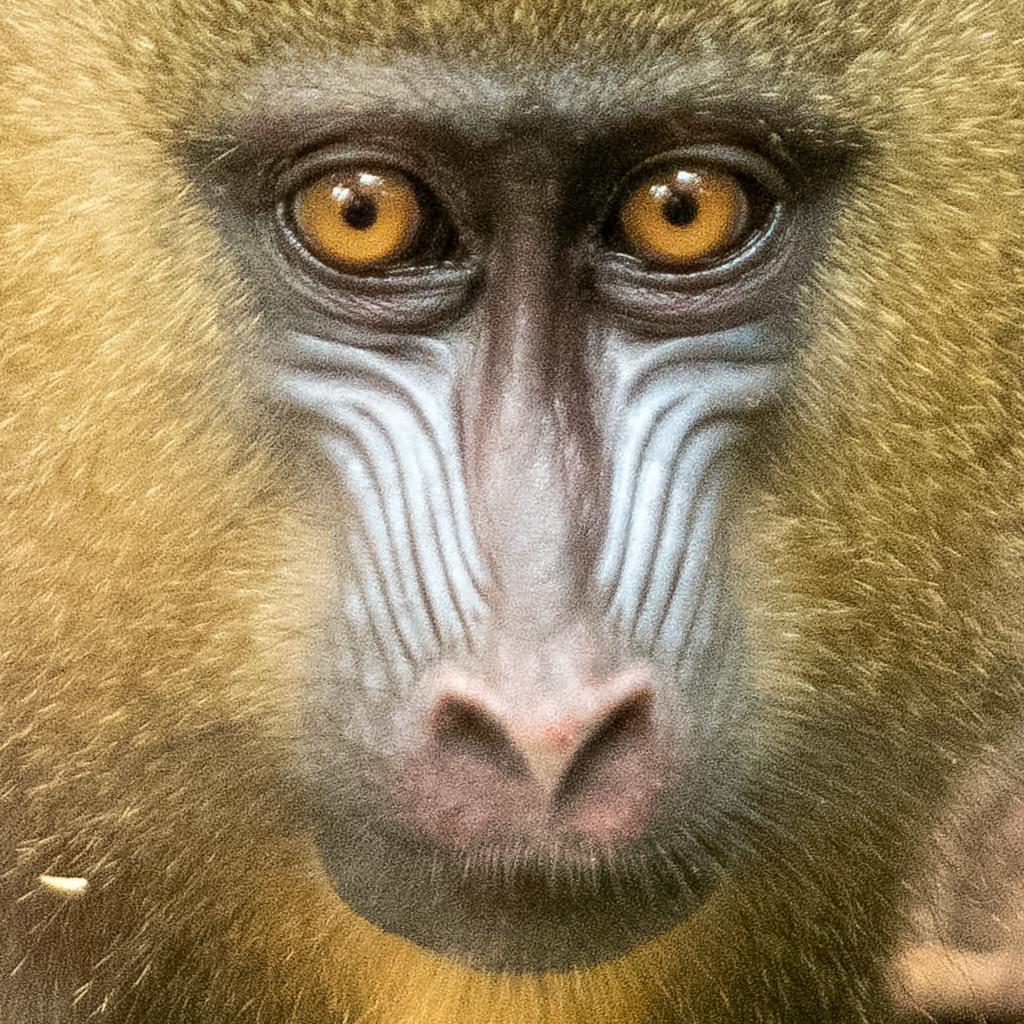} & 
    \includegraphics[scale=0.08]{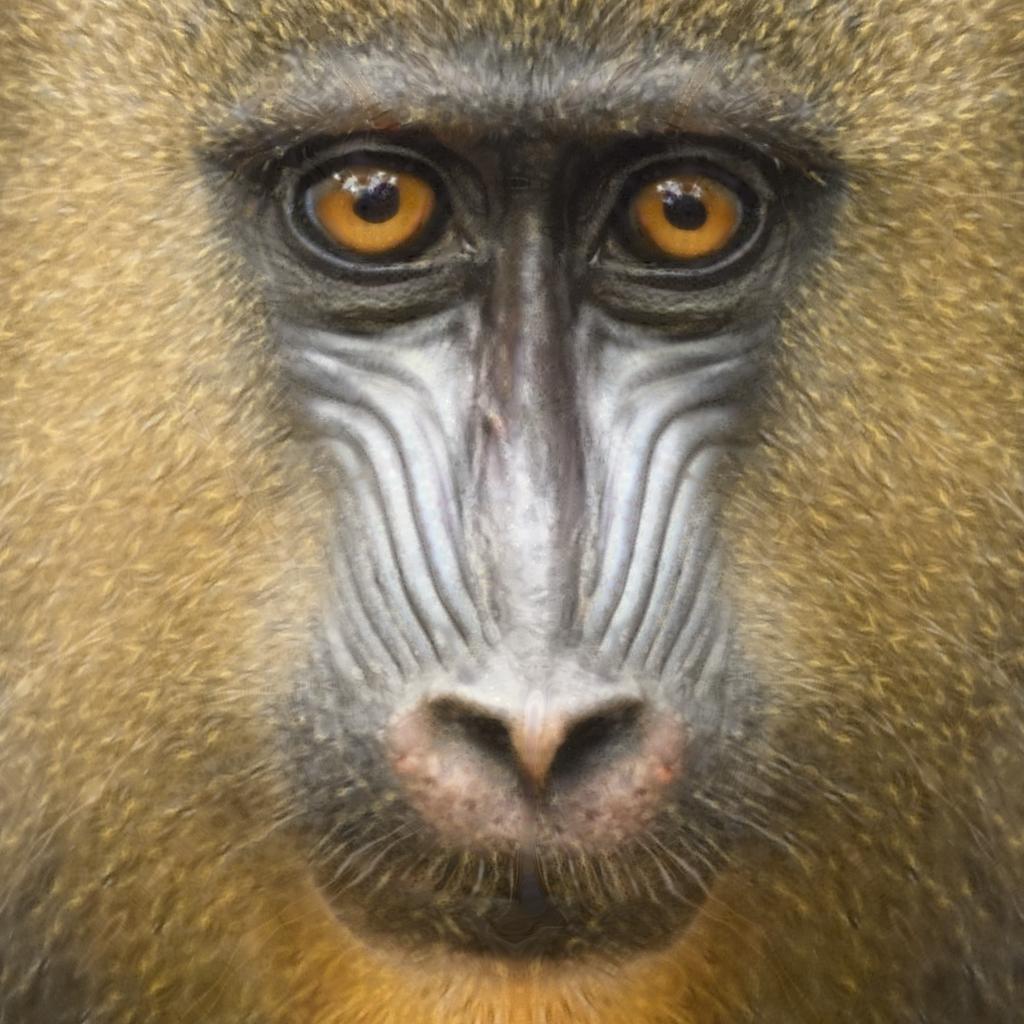} &
    \includegraphics[scale=0.08]{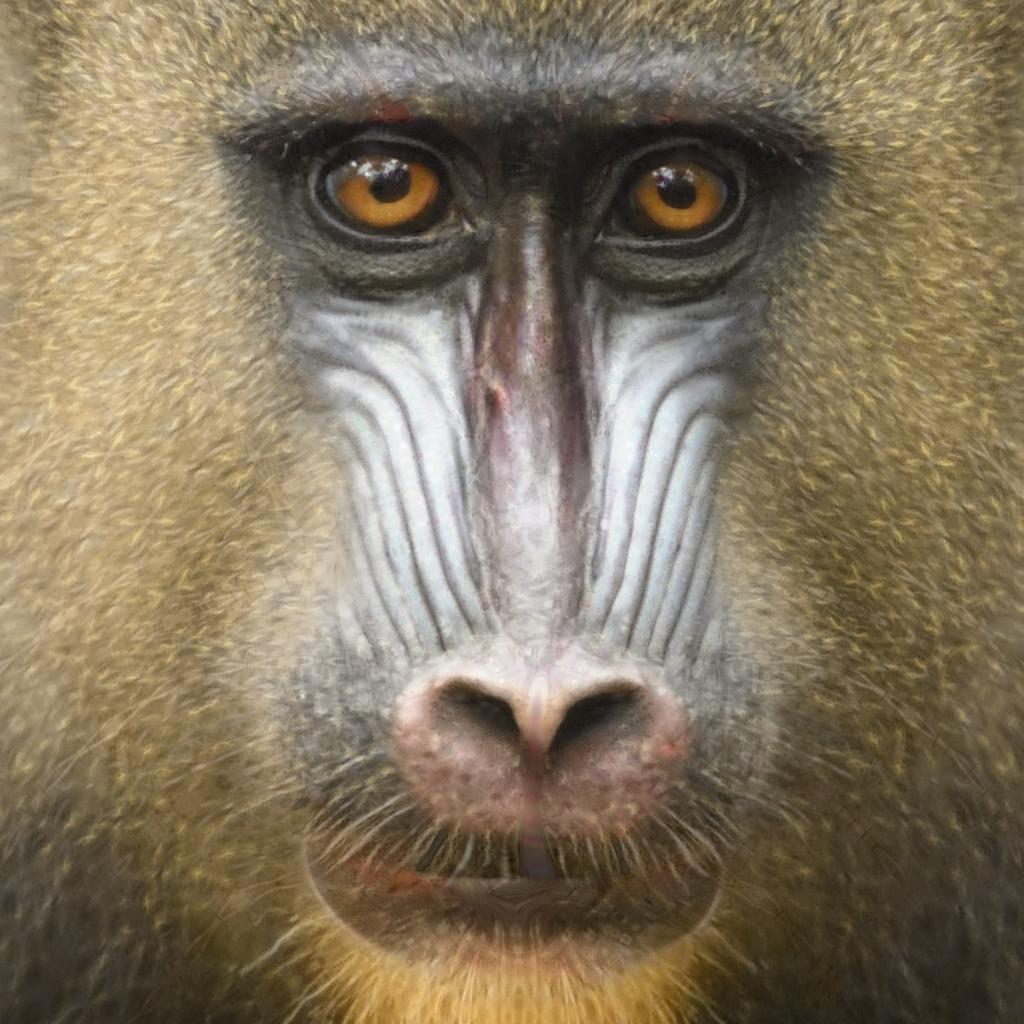} & 
    \includegraphics[scale=0.08]{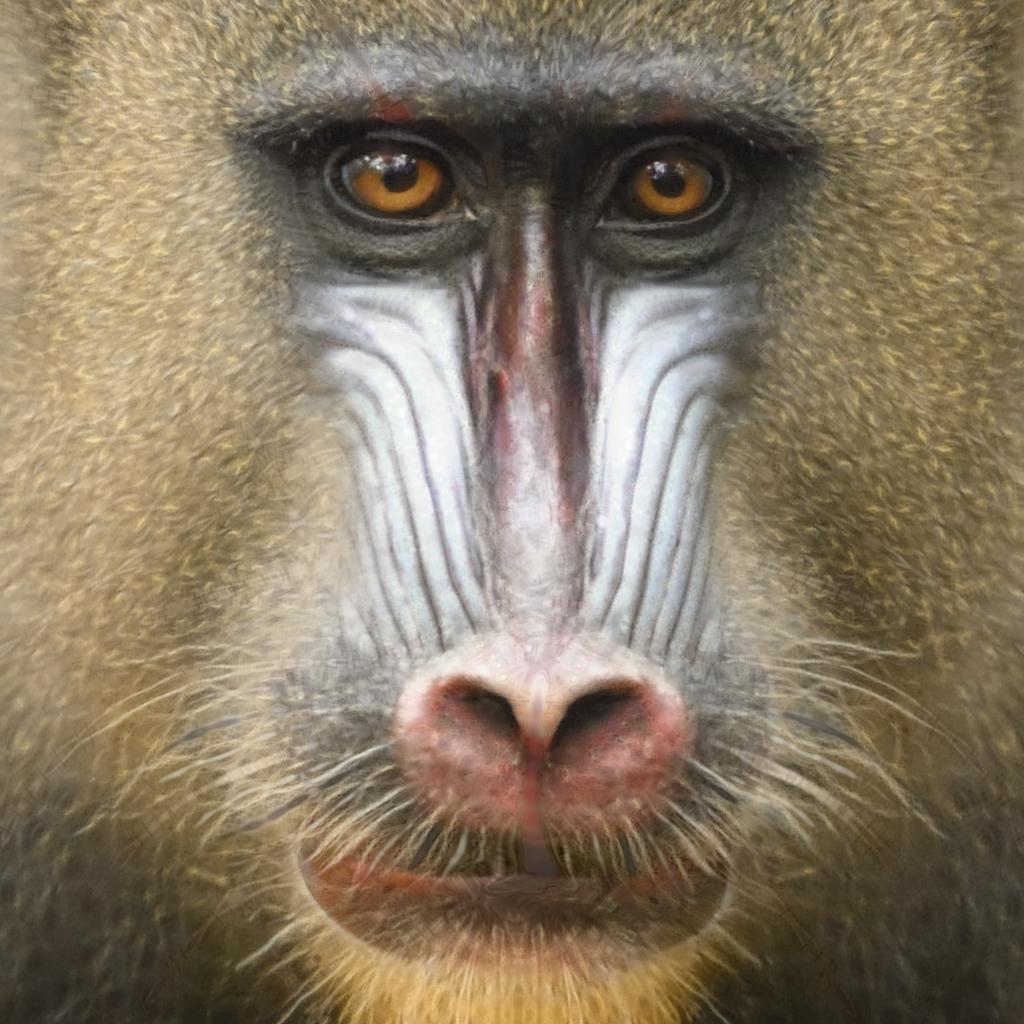}\\
    \scriptsize{\#20220701\_id220\_femadu} & $S_o = -8.28$ & $S_r = -5.58$ (+2$\sigma$) & $S_r = -3.55$ (+3.5$\sigma$) \\ 
     &            & $S_e = -5.69$, $i=7$ & $S_e = -3.64$, $i=4$ \\
\end{tabular}    
\caption{
Examples of encoded and edited female mandrill face images: The first column is the original image from the MFD database; The second column is the image encoded with pSp-mandrill;  The third and fourth columns are edited images with different editing values, the first being the most feminine and the second the least feminine.}
\label{fig:femaleExamples} 
\end{figure}

The results presented in Section~\ref{subsec:étapes} can be applied to any mandrill face image. In this section, we present the pSp-mandrill encoding and editing obtained for 4 images of male mandrill faces, as shown in Fig.~\ref{fig:maleExamples}, and 4 images of female mandrill faces, as shown in Fig.~\ref{fig:femaleExamples}, illustrating several examples of positive and negative deviation step size, within the bounds $\mu \pm 2 \sigma$. Note that an original image can still be edited if its sex level is outside the bounds, as in the case of the image \#20210130\_id62\_malsub(4), Fig.~\ref{fig:maleExamples}, whose original sex level of $-3.84$ is below the minimum male sex level limit of $\mu - 2 \sigma = -3.72$. In this case, editing can only be done by increasing the sex level. For the image \#20210130\_id62\_malsub(4), Fig.~\ref{fig:maleExamples}, the edited image $1$ is obtained by increasing the sex level of $1.5 \sigma$ and the edited image $2$ of $3 \sigma$. It can be seen that for editions close to the upper limit of +2$\sigma$ for females and lower limit of -2$\sigma$ for males, the sex of the mandrills is ambiguous, which is expected as this corresponds to the area of overlap between the two distributions on the sex axis.

\begin{figure}[hbtp!]
%\center
    \includegraphics[width= 7cm]{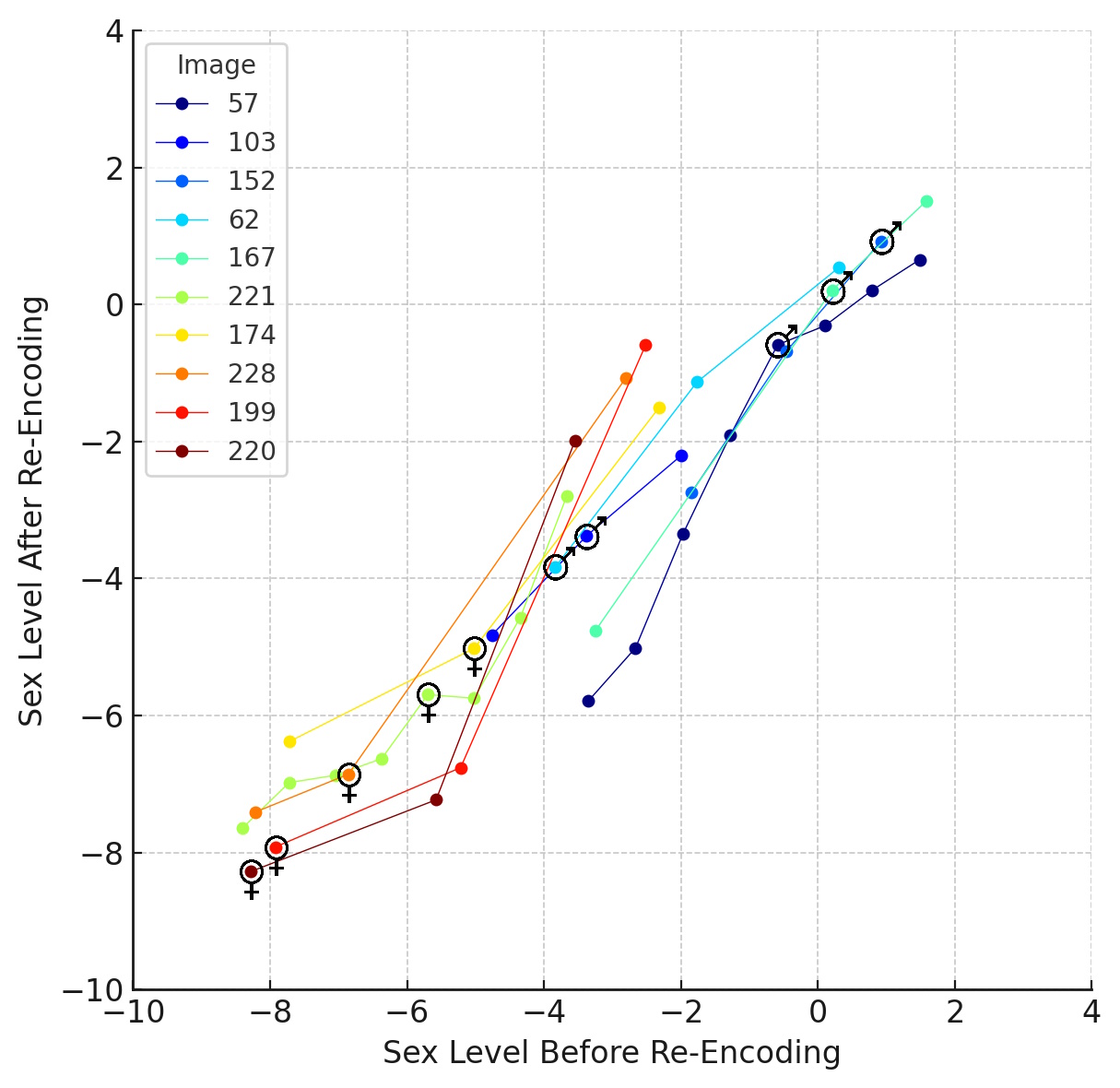} 
\caption{Sex level editing as a function of the obtained sex levels after decoding and re-encoding for all the editing performed and illustrated Fig.~\ref{fig:EditedMaleMandrillFace8}, Fig.~\ref{fig:EditedFemaleMandrillFace8}, Fig.~\ref{fig:maleExamples} and Fig.~\ref{fig:femaleExamples}.}
\label{fig:plot_reencoding} 
\end{figure}

Fig.~\ref{fig:plot_reencoding} shows the relationships of the sex levels before the decoding-re-encoding step as a function of sex levels obtained after the decoding-re-encoding step, for all the editing performed and illustrated in Fig.~\ref{fig:EditedMaleMandrillFace8}, Fig.~\ref{fig:EditedFemaleMandrillFace8}, Fig.~\ref{fig:maleExamples} and Fig.~\ref{fig:femaleExamples}. We observe an almost monotonic increasing relationship between the two axes, although this is not strictly the case everywhere, which justifies the conditions $C$ and $F$ of Algorithm~\ref{algo:alg2}, where the $\Delta_e$ optimization can be decreasing. However, there is a visual tendency towards a linear or logit relationship between the two axes. Thus, estimating the first value of $\Delta_e$ with a predictive model adapted to these points is an area for improvement to reduce the number of iterations of Algorithm~\ref{algo:alg2}. However, note that because convergence was reached very quickly (in a few iterations) with our approach, it is not sure that such an analytic approach is worthwhile. It can also be seen that females have predominantly smaller sex level values than males, which corresponds well to the distribution of real images projected on the axis.

%%%%%%%%%%%%%%%%%%%%%%%%%%%%%%%%%%%%%%%%%%%%%%%%%%%%%%%
\subsection{Discussion}
\label{subsec:discussion}

%Some of the results present distinctive aspects that need to be discussed. 
In this section, we propose to discuss some of the results obtained, which present some particular aspects. First of all, we have chosen to limit ourselves to bounds of +2$\sigma$ and -2$\sigma$ around the sex level averages of each sex when editing the images. Indeed, generative AI models are generally good at interpolating between images in the training database, but poor at extrapolating beyond this data. For example as shown in {Fig.~\ref{fig:artefacts}, with an editing reaching +4$\sigma$ from the male mean, the image of the edited mandrill face shows artifacts that make it unrealistic. However, experts accustomed to observing mandrills have confirmed the plausibility of these types of editing, when they remain between bounds.

\begin{figure}[hbtp!]
\center
    \includegraphics[width=3.4cm]{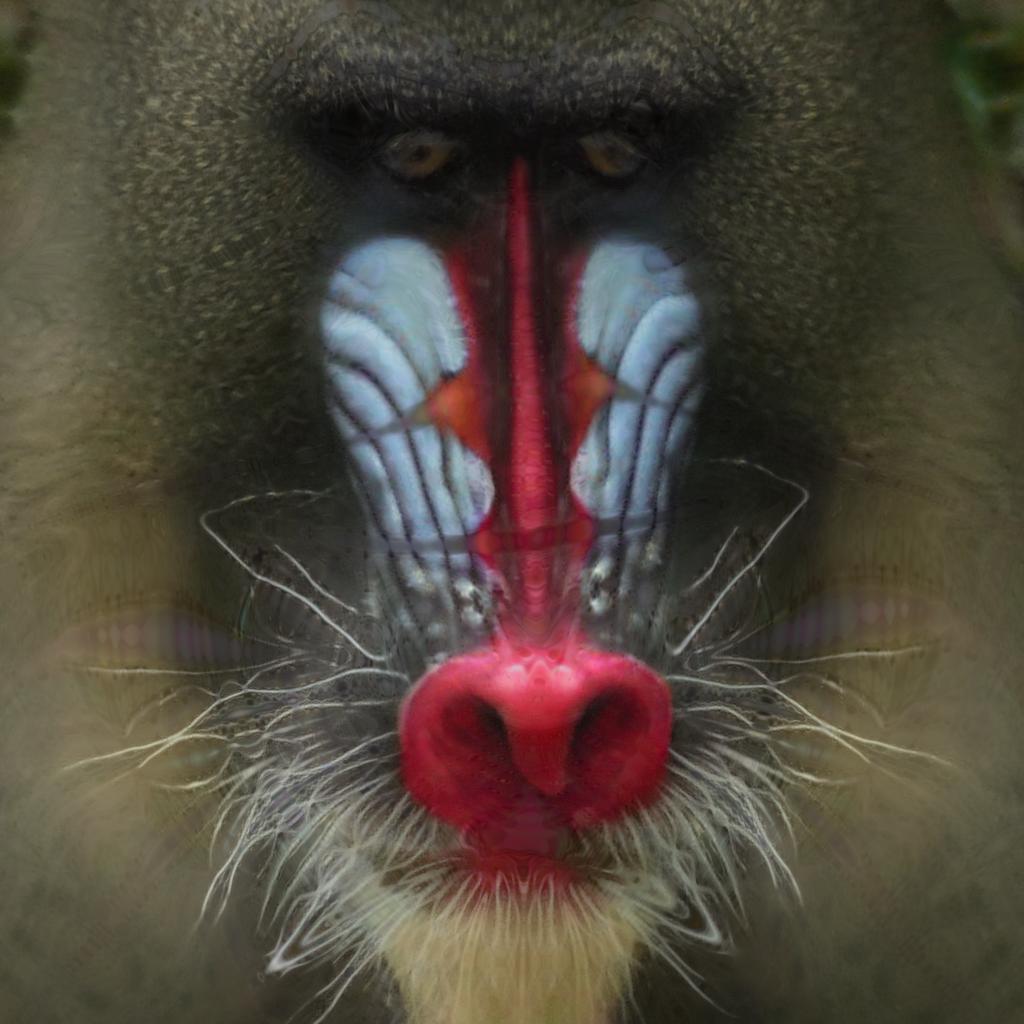}   
\caption{Editing out of bounds of the face of the mandrill \#20150517\_id57\_maladu\_(13), showing some artefacts.}
\label{fig:artefacts} 
\end{figure}

In addition, behavioral and non-visual features seem to emerge from the editing. In some cases, the more masculine an image is edited, the more the mandrill's mouth is open, as shown in the last column of Fig.~\ref{fig:maleExamples}, in particular for the second and the fourth examples. This may be explained by the fact that male mandrills in the MFD database are more aggressive and open their mouths to scream. In other cases, the more feminine an image is edited, the more the mandrill turns its head away. Indeed, female mandrills are more fearful and turn their heads when a photo is taken to run away.

It is also noticeable that the background of the images behind the mandrills is slightly modified by the editing process. This shows that the StyleGAN3-mandrill's space $W+$ does not allow perfect disentanglement of the gender variable. Some work ~\cite{alaluf_third_2022} suggests that editing in a different GAN space, called S-space, could improve this, which is a direction for future improvement.

Beyond results, the method itself presents some points to discuss. While the sex axis is calculated in the space $W$, editing is performed in the space $W+$, duplicating the axis 18 times. This can lead to some inaccuracy, with editing in certain dimensions that do not correspond precisely to the sex, but more to the age, for example. This probably explains the need to optimize the editing level with Algorithm~\ref{algo:alg2}. One way of improving this would be to train the SVM directly in the space $W+$, by re-encoding in the space $W+$ the images generated from $W$ and use them for training.

\begin{figure}[hbtp!]
    \centering
        \includegraphics[width=10.cm]{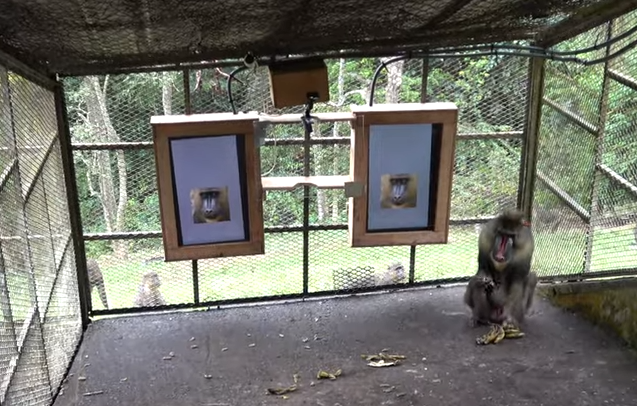}      
        \caption{Illustration showing the use of the proposed methods to conduct behavioral choice experiments with semi-captive mandrills in Gabon. The aim is to determine whether sex level has an influence on mate choice.}
    \label{fig:experiment}
\end{figure}

The methods proposed in this paper aim to conduct behavioral choice experiments with semi-captive mandrills in Gabon. The aim is to determine whether sex level has an influence on mate choice. To improve the attractiveness of synthetic mandrill images for real mandrills, we animated the images by editing them according to head orientation, thus creating a video of a mandrill moving its head from left to right. As shown in Fig.~\ref{fig:experiment}, two images of mandrill faces of the same individual are presented to a mandrill of the opposite sex, on two juxtaposed screens. Using various sensors, including cameras, we then measure which screen attracts the observed mandrill the most. These approaches could be applied in many different ways. We could go even further by exploiting the potential of GAN to improve the quality of the images generated~\cite{10093128}.

%%%%%%%%%%%%%%%%%%%%%%%%%%%%%%%%%%%%%%%%%%%%%%%%%%%%%%%
%                     Conclusion                      %
%%%%%%%%%%%%%%%%%%%%%%%%%%%%%%%%%%%%%%%%%%%%%%%%%%%%%%%
\section{Conclusion}
\label{sec:conclusion}
In this paper, we presented a GenAI-based framework for generating, editing and assessing images of mandrill faces based on their sex. More precisely, we used a StyleGAN3 trained on a database of thousands of real mandrill images, StyleGAN3-mandrill, to generate synthetic images of these primates, coupled with an encoder trained on this same database,  pSp-mandrill, to encode the real images in a latent space close to that of the trained GAN. We succeeded in finding a sex axis on these two spaces ($W$ and $W+$) to edit the images in order to change their sex level and thus, make them more or less feminine or masculine. In addition, we proposed a method for assessing this editing through statistics of the distribution of real images in the database on the sex axis. Our approach not only shows the potential of the StyleGAN3-based framework in animal image manipulation, but also helps scientists working with visual stimuli to design stimuli with controlled variation for psychological and behavioral experiments with both humans and animals.

This work has also revealed several limitations of the GenAI-based approach, such as an imperfect disentanglement of the features of interest, which will be the focus of features research. This work would also be enhanced if applied to other animal species, included humans, which would nevertheless require a database on the same scale as that for mandrills.
\begin{acks}
This work was supported in part by the Agence Nationale de la Recherche (ANR-20-CE02-0005-01), in part by the CNRS through the MITI interdisciplinary programs (Programme Interne Blanc MITI 2023.1 - Projet: DEEPCOM- L'intelligence artificielle pour étudier la communication).
\end{acks}

\bibliographystyle{ACM-Reference-Format}
\bibliography{biblio}

\end{document}